%% file: main.tex
\documentclass[10pt]{article} 
\usepackage[preprint]{tmlr}

\input{math_commands.tex}

\usepackage{hyperref}
\usepackage{url}

\usepackage[utf8]{inputenc} 
\usepackage[T1]{fontenc}    
\usepackage{booktabs}       
\usepackage{amsfonts}       
\usepackage{nicefrac}       
\usepackage{microtype}      
\usepackage{xcolor}         
\usepackage{subcaption}

\usepackage{amsmath}
\usepackage{amssymb}
\usepackage{mathtools}
\usepackage{amsthm}

\usepackage{tikz}
\usetikzlibrary{automata,chains,arrows,arrows.meta,shadows,patterns,calc,backgrounds,positioning}
\usepackage[normalem]{ulem}

\tikzset{
->, 
>=stealth, 
node distance=2cm, 
every state/.style={thick, fill=gray!10}, 
initial text=$ $, 
every axis/.style={-},
}

\usepackage[capitalize,noabbrev]{cleveref}

\title{Length Generalization with Log-Depth Recurrent Units}

\author{\name Charles Pert \email charles.pert@imperial.ac.uk \\
      \addr Department of Computing\\
      Imperial College London
      \AND
      \name Dalal Alrajeh \email dalal.alrajeh@imperial.ac.uk \\
      \addr Department of Computing\\
      Imperial College London
      \AND
      \name Alessandra Russo \email a.russo@imperial.ac.uk \\
      \addr Department of Computing\\
      Imperial College London}

\begin{document}

\maketitle

\begin{abstract}
    Length generalization remains a persistent challenge for neural networks: recurrent models tend to suffer from positional biases, while transformers are constrained by fixed computational depth. Regular languages provide a frequently used testbed for evaluating length generalization, as label prediction can be checked for any sequence length. We propose MLP-LDRU, a type of Log-Depth Recurrent Unit, which captures a class of associativity-biased operators designed to approximate recurrence through parallel reduction. We evaluate MLP-LDRU on 21 regular-language tasks, consisting of standard benchmarks and new prefix languages, where it achieves 100\% out-of-distribution accuracy on 18 tasks and at least 99.9\% on the remaining 3 when increasing max training length, outperforming comparable recurrent and attention-based models. We further evaluate MLP-LDRU beyond regular languages on ListOps and NLP classification benchmarks, where it performs competitively.
\end{abstract}

\section{Introduction}

Generalization to out-of-distribution (OOD) sequence lengths remains a central open challenge in neural modeling. As defining OOD sequences is difficult in natural language contexts, regular languages have become a popular testbed for studying this problem. Regular languages are diverse and enable precise mathematical formulation, directly connecting to automata theory, enabling exact validation of correctness at any sequence length. Evaluating models on these tasks enables precise measurement of whether a model has learned the rules underpinning the task, enabling a characterization of OOD generalization~\citep{strobl2024}. %

Prior works~\citep{deletang2023neural,Chi2023,butoi2025training} highlight a trade-off: RNNs~\citep{elman1990finding} offer length-adaptive computation, but earlier tokens undergo more recurrent steps than later ones, which can lead to long-range memory issues~\citep{bengio1994learning}. Transformers~\citep{vaswani2017attention}, on the other hand, apply a fixed stack of layers independent of sequence length, and tend to generalize less reliably to OOD lengths. We are interested in enabling recurrence using parallel reduction. It has been shown~\citep{ladner1980parallel,BlellochTR90} that with an associative operator, the parallel reduction algorithm can give a log-depth implementation of a recurrence. But these operator-level constraints have not been sufficiently explored. We hypothesize that learning such operators via an explicit associativity (and identity) bias can improve length extrapolation on regular language tasks. We start from regular tasks as a rigorous first step toward exploring length-extrapolating sequence models in more complex settings.

In this paper, we explore log-depth sequence processing by learning reduction operators, in particular, those designed to approximate associativity, and name this class \emph{Log-Depth Recurrent Units} (LDRUs), which compose token embeddings via a binary operator applied in a balanced tree. Applying this operator in a reduction yields uniform $\mathcal{O}(\log n)$ computational depth across tokens, mitigating the positional bias of standard RNNs while enabling parallel computation to behave as a recurrence along the sequence. In this paper, we experiment using \emph{MLP-LDRU}\footnote{Code will be released upon publication.}, an LDRU given in Section~\ref{sec:method}. LDRUs are motivated by the monoid view of regular languages (see Appendix~\ref{app:connection_theory} for details), where concatenation corresponds to an associative product; this suggests that if a learned operator approximates associativity, then using it as a reduction operator will likewise approximate a sequential recurrence, a property that supports length generalization on regular tasks~\citep{deletang2023neural}.

LDRU-based models form a subclass of balanced binary-tree recursive neural networks~\citep[BBT-RvNNs;][]{munkhdalai-yu-2017-neural,NEURIPS2019_d8e1344e,chowdhury2023recursion}. We use the same log-depth reduction, but with a different learning target: instead of learning an unconstrained tree cell, we intentionally bias the operator toward approximate associativity and enforce identity behavior (padding acts as a neutral element). These biases ideally enable the learning of an approximation of a sequential recurrence.

We evaluate length generalization on a suite of regular language transduction tasks, including \emph{prefix languages} that isolate long-range dependency handling under length extrapolation. MLP-LDRU achieves 100\% OOD accuracy on 18 of 21 tasks when trained on sequences up to length 40, and near-perfect performance on the remaining 3. Further analysis of the underlying monoids of the tasks suggests that remaining failures stem from insufficient coverage of monoid compositions in training data rather than an architectural limitation. We additionally study ListOps~\citep{nangia2018listops} to test compositional extrapolation and report results on several natural language classification benchmarks to probe behavior beyond regular languages.

This paper makes the following contributions:
\begin{itemize}
    \item We propose \emph{MLP-LDRU}, a specific Log-Depth Recurrent Unit, a binary operator designed to induce approximately associative behavior with identity handling. 
    \item We empirically evaluate MLP-LDRU on 21 regular tasks, with transfer probe experiments on ListOps extrapolation and NLP classification benchmarks.
    \item We introduce \emph{prefix languages}, a new class of regular tasks designed to test long-range dependency handling under length extrapolation.
    \item We analyze training data requirements for length generalization, relating failures to insufficient coverage of equivalence class compositions in short sequences.
\end{itemize}

The rest of this paper is organized as follows. Section~\ref{sec:prefix_languages} introduces the prefix languages. Section~\ref{sec:method} details MLP-LDRU. Section~\ref{sec:experiments} explains our experimental set-up, and we provide the results in Section~\ref{sec:results}. Related work is discussed in Section~\ref{sec:related_work}. We discuss the implications of our findings in Section~\ref{sec:discussion}. Finally, we conclude in Section~\ref{sec:conclusion}.

\section{Prefix Languages}
\label{sec:prefix_languages}

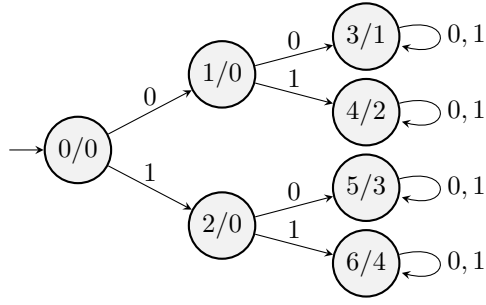
\begin{figure}[!ht]
    \centering
    \begin{tikzpicture}[every node/.style={inner sep=0pt}]
    \node[state, initial] (q0) {$0/0$};
    
    \node[state, right=1cm of q0, yshift=1cm] (q1) {$1/0$};
    
    \node[state, right=1cm of q0, yshift=-1cm] (q2) {$2/0$};

    \node[state, right=1cm of q1, yshift=0.5cm] (q3) {$3/1$};
    
    \node[state, right=1cm of q1, yshift=-0.5cm] (q4) {$4/2$};

    \node[state, right=1cm of q2, yshift=0.5cm] (q5) {$5/3$};
    
    \node[state, right=1cm of q2, yshift=-0.5cm] (q6) {$6/4$};

    \draw (q0) edge[] node[yshift=0.2cm] (int00) {$0$} (q1);
    \draw (q0) edge[] node[yshift=0.2cm] (int01) {$1$} (q2);

    \draw (q1) edge[] node[yshift=0.2cm] (int10) {$0$} (q3);
    \draw (q1) edge[] node[yshift=0.2cm] (int11) {$1$} (q4);
    \draw (q2) edge[] node[yshift=0.2cm] (int12) {$0$} (q5);
    \draw (q2) edge[] node[yshift=0.2cm] (int13) {$1$} (q6);
    
    \draw (q3) edge[loop right] node[xshift=0.1cm] (acc0) {$0,1$} (q3);
    \draw (q4) edge[loop right] node[xshift=0.1cm] (acc1) {$0,1$} (q4);
    \draw (q5) edge[loop right] node[xshift=0.1cm] (acc2) {$0,1$} (q5);
    \draw (q6) edge[loop right] node[xshift=0.1cm] (acc3) {$0,1$} (q6);
    \end{tikzpicture}
    \caption{The prefix language $P_{2,2}$. Nodes are states, and edges are transitions. Labels within nodes $X/Y$ indicate the state $X$ and output symbol $Y$ if a sequence's run ends in $X$.}
    \label{fig:prefix_language_machine}
\end{figure}

We introduce prefix languages, denoted $P_{p,q}$, regular tasks for testing long-range dependency handling through a test of persistent memory of early-state information under distractors. The first $p$ symbols determine the machine's final state, requiring a model to retain memory of its current state while processing the remainder of the sequence. As each $p$-length prefix leads to a different final state, these tasks also require a model to distinguish between different orderings of the same information (i.e., consider positional information). We also parameterize $q$, the number of symbols in the alphabet. See Fig.~\ref{fig:prefix_language_machine} for a graphical representation of $P_{2,2}$. We give a mathematical definition of prefix languages in Appendix~\ref{app:prefix_languages_definition}.

\section{Method}
\label{sec:method}

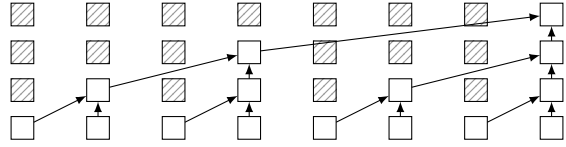
\begin{figure}[bhtp]
    \centering
    \begin{tikzpicture}[scale=1,
            square/.style={
                rectangle,
                draw=black,
                fill=white,
                minimum size=0.3cm,
                inner sep=0pt,
                outer sep=0pt,
                anchor=center
            },
            inactive/.style={
                rectangle,
                draw=black,
                pattern=north east lines,
                pattern color=black!40,
                minimum size=0.3cm,
                inner sep=0pt,
                outer sep=0pt,
                anchor=center
            },
            arrow/.style={->,>=latex}
        ]
        \foreach \x in {0,...,7} {
            \node[square] (L1\x) at (\x,0) {};
        }
        
        \foreach \x in {1,3,5,7} {
            \node[square] (L2\x) at (\x,0.5) {};
        }
        \foreach \x in {0,2,4,6} {
            \node[inactive] at (\x,0.5) {};
        }
        
        \foreach \x in {3,7} {
            \node[square] (L3\x) at (\x,1) {};
        }
        \foreach \x in {0,1,2,4,5,6} {
            \node[inactive] at (\x,1) {};
        }
        
        \node[square] (L4) at (7,1.5) {};
        \foreach \x in {0,...,6} {
            \node[inactive] at (\x,1.5) {};
        }
        
        \foreach \x in {0,1} {
            \draw[arrow] (L1\x) -- (L21);
        }
        \foreach \x in {2,3} {
            \draw[arrow] (L1\x) -- (L23);
        }
        \foreach \x in {4,5} {
            \draw[arrow] (L1\x) -- (L25);
        }
        \foreach \x in {6,7} {
            \draw[arrow] (L1\x) -- (L27);
        }
        
        \draw[arrow] (L21) -- (L33);
        \draw[arrow] (L23) -- (L33);
        \draw[arrow] (L25) -- (L37);
        \draw[arrow] (L27) -- (L37);
        
        \draw[arrow] (L33) -- (L4);
        \draw[arrow] (L37) -- (L4);
        
    \end{tikzpicture}
    \caption{Log-depth reduction on a length-8 sequence: a pairwise composition operator is applied in parallel to reduce the sequence to a single embedding in $\log_2(8)=3$ steps, illustrating the $O(\log n)$ depth.}
    \label{fig:info-flow}
\end{figure}

In this section, we describe the MLP-based operator (used in MLP-LDRU, denoted as $\odot_\theta$) used in our experiments and describe how it produces a sequence embedding. See Fig.~\ref{fig:info-flow} for an illustration of the parallel reduction on a length-8 sequence. Given a sequence of length $n$ token embeddings, the model repeatedly applies $\odot_\theta$ in parallel to adjacent pairs for $\lceil \log_2 n \rceil$ steps. Each step consists of applying the operator on all pairs, followed by a residual feedforward network, then layer normalization. The weights of these components are shared across steps. After each reduction step, we apply dropout, empirically observed to encourage more robust generalization (see Appendix~\ref{app:optimizer_ablation}). At any step, if the length of the sequence is odd, we pad it with a zero embedding, $\mathbf{0}$. We treat $\mathbf{0}$ as a neutral placeholder in $\odot_\theta$: if either input is $\mathbf{0}$, then the operator output is the other input (i.e., composition is skipped). The residual connection and layer normalization proceed as standard. This pass-through prevents padding-induced artifacts and avoids requiring $\odot_\theta$ to learn an identity map. 

The core of $\odot_\theta$ is an element-wise sum (a natural associative operation) with learned gating weights to control information flow. We produce the gating weights using an MLP. The operator's gating design draws from the success of gating information flow in recurrent architectures~\citep{hochreiter1997long,chung2014empirical,jozefowicz2015an,NEURIPS2019_d8e1344e}. We define $\odot_\theta : \mathbb{R}^d \times \mathbb{R}^d \rightarrow \mathbb{R}^d$ on inputs $\mathbf{h}_i, \mathbf{h}_j \in~\mathbb{R}^d$ as follows:
\begin{gather}
[\mathbf{g}_i; \mathbf{g}_j] = \text{MLP}([\mathbf{h}_i; \mathbf{h}_j]) \text{ where } \mathbf{g}_i,\mathbf{g}_j \in \mathbb{R}^{d}\label{eq:mlp} \\
\mathbf{f}_i = \mathbf{V}_i(\mathbf{g}_i \circ \mathbf{h}_i) + \mathbf{b}_i, \mathbf{f}_j = \mathbf{V}_j(\mathbf{g}_j \circ \mathbf{h}_j) + \mathbf{b}_j \\
\odot_\theta(\mathbf{h}_i, \mathbf{h}_j) = \mathbf{W}_{\text{out}}(\mathbf{f}_i + \mathbf{f}_j) + \mathbf{b}_{\text{out}} \in \mathbb{R}^d,\label{eq:odot_theta}
\end{gather}
where $[\cdot; \cdot]$ denotes concatenation, $\circ$ is element-wise multiplication, $\mathbf{V}_i,\mathbf{V}_j,\mathbf{W}_{\text{out}} \in \mathbb{R}^{d \times d}$ are learned projection matrices, and $\mathbf{b}_i,\mathbf{b}_j,\mathbf{b}_{\text{out}} \in \mathbb{R}^d$ are learned bias vectors. We use a 3-layered MLP with expansion factors $(1, 2, 1)$. $\mathbf{V}_i$, $\mathbf{V}_j$, and $\mathbf{W}_{out}$ are all initialized with the identity, and the corresponding biases are initialized to $\mathbf{0}$. The element-wise sum combining token embeddings and the combination of these initializations biases $\odot_\theta$ toward approximately associative behavior.

We briefly describe how a sequence embedding is produced from beginning to end. An embedded sequence is reduced to a single embedding by iteratively applying $\odot_\theta$ steps. The single embedding, $\mathbf{h}_n$, is passed to a linear classifier to produce class logits. Complete model diagrams and initialization schemes are provided in Appendix~\ref{app:ldru_operator_details}.

\subsection{Complexity Analysis}
\label{sec:complexity}

MLP-LDRU achieves $O(n d^2)$ work complexity with $O(\log n)$ computational depth (referred to as depth), combining RNN-like linear scaling (of sequence length) with logarithmic-time parallelization. This contrasts with RNNs' $O(n)$ sequential depth and transformers' $O(n^2 d + n d^2)$ quadratic sequence length complexity. Further discussion on complexity is provided in Appendix~\ref{app:computational_complexity}.

\begin{table}[h]
\centering
\small
\caption{Computational complexity comparison. The LDRU achieves $\log n$ depth and avoids quadratic length scaling.}
\begin{tabular}{lcc}
\toprule
Architecture & Work Complexity & Depth \\
\midrule
RNN & $O(nd^2)$ & $O(n)$ \\
Transformer & $O(n^2 d + n d^2)$ & $O(1)$ \\
\textbf{MLP-LDRU (Ours)} & $\mathbf{O(nd^2)}$ & $\mathbf{O(\log n)}$ \\
\bottomrule
\end{tabular}
\end{table}

\begin{figure}[htbp]
    \centering
    \includegraphics[]{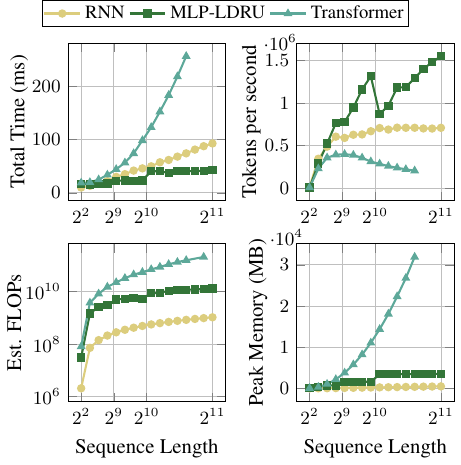}
    \caption{Empirical runtime analysis of MLP-LDRU, RNN, and transformer across small (160k) parameter regime used for our regular task study. We note that MLP-LDRU's throughput reaches maxima at $2^n$ as these are the optimally efficient lengths for reduction.}
    \label{fig:scaling_plots}
\end{figure}

We present the practical trade-offs between MLP-LDRU, RNN, and transformer in Fig.~\ref{fig:scaling_plots}. We profile the forward and backward passes of the three architectures across sequence length with a batch size of 32. Although MLP-LDRU incurs higher FLOPs and peak memory usage than the RNN, its greater GPU parallelism yields higher throughput. Transformer is significantly slower and costs more FLOPs due to its quadratic length complexity.

\section{Experiments}
\label{sec:experiments}

We evaluate MLP-LDRU using sequence classification tasks. For regular tasks, this corresponds to predicting the single output symbol produced by the task's corresponding Moore machine. We illustrate these machines in Appendix~\ref{app:experimental_details}. The regular tasks we evaluate are a combination of those previously studied by~\citet{deletang2023neural} and \citet{Bhattamishra2020OnTA}, and 6 parameterizations of prefix languages. We also evaluate ListOps~\citep{nangia2018listops} to assess MLP-LDRU on hierarchical tasks beyond regular languages. We aim to answer the following research questions:
\begin{description}
    \item\textbf{RQ1} Does MLP-LDRU's generalization ability outperform established state-of-the-art architectures on: (a) regular tasks and (b) ListOps?
    \item\textbf{RQ2} Does increasing the maximum training sequence length improve OOD performance?
    \item \textbf{RQ3} Does the choice of reduction operator impact OOD performance?
\end{description}

All models were trained under identical conditions (same training lengths, optimizer, regularization, and number of steps), with full hyperparameters detailed in Appendix~\ref{app:experimental_details}. We train each model by minimizing the cross-entropy loss between the class logits and the ground-truth labels. We use the AMSGrad~\citep{reddi2018on} optimizer for regular tasks, as we empirically observed improved generalization (see Appendix~\ref{app:optimizer_ablation} for details). We use the Adam~\citep{kingma2015adam} optimizer for non-regular tasks. We do not use early stopping. We use 3 seeds for regular tasks, and 5 seeds for non-regular tasks. We now describe the experimental setup for each research question.

\textbf{RQ1a} We train MLP-LDRUs on 21 regular tasks on sequences of variable length, from 1 to 40. We evaluate generalization using OOD accuracy, i.e., the mean accuracy over 512 sampled sequences for each length from 41 to 500. This setup is standard in length generalization studies on regular tasks~\citep{deletang2023neural,liu2023transformers,ruoss2023randomized,Chi2023}. We compare MLP-LDRU to the classic RNN, LSTM, Gated DeltaNet~\citep{yang2025gated}, transformer (TF), BBT-GRC~\citep{shi-etal-2018-tree}, a BBT-RvNN equipped with a Gated Recursive Cell~\citep[GRC;][]{NEURIPS2019_d8e1344e}, and RegularGPT~\citep{Chi2023} models. We train for 100k or 1M steps, depending on the task. We use 3 positional encodings (PEs) for the transformer: NoPE~\citep{kazemnejad2023the}, ALiBi~\citep{press2022train}, and randomized RoPE~\citep{ruoss2023randomized}, denoted as $\sim$RoPE. We assess the statistical significance of performance differences between MLP-LDRU and baseline architectures across tasks using Student's t-tests.

\textbf{RQ1b} We train MLP-LDRUs on ListOps sequences from length 5 to 40 with a maximum nesting of 3 and at most 9 arguments per list. We train models using 3 dataset sizes: 100k, 500k, and 1M sequences. For testing, we consider length buckets: $[5, 20]$, $[21, 40]$, $[41, 60]$, $[61, 80]$, $[81, 100]$, $[101, 200]$; max depths: 3, 5; max arguments: 9, 14. We test 10k samples per combination, except for lengths $[81,100]$, $[101,200]$ with max depth 3 and max arguments 9. We compare MLP-LDRU to LSTM, transformer with NoPE, ALiBi, and Sinusoidal PEs, BBT-GRC, and RIR-GRC~\citep{chowdhury2023recursion}. Further details can be found in Appendix~\ref{app:listops_details}.

\textbf{RQ2} We train MLP-LDRUs and RNNs on $D_4$, $D_6$, $D_8$, and $D_{12}$ tasks (where $D_n$ is the recognition of the Dyck-1 language up to $n$ stack depth), with varied maximum training lengths: $40$, $60$, $100$, and $150$. We use these tasks because neither the RNN nor MLP-LDRU generalizes with 100\% accuracy to these tasks when trained with a maximum training length of $40$ (except for MLP-LDRU on $D_4$). We report the OOD accuracy, e.g., for a maximum training length of $60$, the evaluation is on sequences of length $61$ to $500$.

\textbf{RQ3} We compare five choices of reduction operator evaluated on the regular tasks from \citet{deletang2023neural}. All settings are otherwise the same as for \textbf{RQ1}. We test: (1) element-wise sum, (2) a linear projection of the concatenation of the two embeddings without activation, (3) a simple gated combination of the two embeddings, (4) GRC, and (5) the MLP-based operator used in MLP-LDRU. We report mean OOD accuracy for each operator-task combination.

\section{Results}
\label{sec:results}

We present the results of the \textbf{RQ1a} regular experiments in Table~\ref{tab:central_result}. MLP-LDRU demonstrates stronger or comparable length generalization across these tasks, achieving 100\% OOD accuracy on 18 tasks and near-perfect performance on the remaining 3 tasks. MLP-LDRU outperforms all 3 transformer variants except on $D_8$ and $D_{12}$ for $\sim$RoPE; moreover, $\sim$RoPE achieves the highest OOD accuracy of all baselines on $D_{12}$, indicating that while it is not as consistent as MLP-LDRU, it is certainly more powerful than NoPE and ALiBi on $D_n$. However, $\sim$RoPE does not provide consistent gains across regular tasks as its performance on Modular Arithmetic is much worse (31.4\%) compared to the other TF PEs considered (76.0\%, 54.4\% for NoPE and ALiBi, respectively). Gated DeltaNet (and two additional SSMs that we include in the appendix, S5~\citep{smith2023simplified} and SD-SSM~\citep{terzic2025sdssm}) never exceeds MLP-LDRU performance on any task. Likewise, MLP-LDRU outperforms RegularGPT and matches or exceeds the capabilities of RNNs (except on $D_8$), LSTMs, and BBT-GRC. We compared MLP-LDRU and RNN performance on $D_8$ and determined that the performance gap is not significant. Full results tables are given in Appendix~\ref{app:full_results}.

\input{tables/central_result.tex}

\input{tables/listops_summary.tex}

\cref{tab:listops_summary} presents the results of the \textbf{RQ1b} ListOps experiments on the 1M-size dataset (additional results provided in Appendix~\ref{app:full_results}). RIR-GRC and BBT-GRC achieve the highest accuracy across test sequences for all dataset sizes, followed by MLP-LDRU, then transformer (ALiBi) and LSTM, then the other transformers. Notably, under our ListOps experimental setup (train lengths 5-40, Adam optimizer), all models' accuracy drops on sequences longer than those seen during training. A small hyperparameter sweep on MLP-LDRU revealed that its performance improves when associativity regularization is applied to $\odot_\theta$, i.e., minimizing the cosine distance between $(\mathbf{h}_a \odot_\theta \mathbf{h}_b)\odot_\theta \mathbf{h}_c$ and $\mathbf{h}_a \odot_\theta(\mathbf{h}_b \odot_\theta \mathbf{h}_c)$, see Appendix~\ref{app:listops_details} for performance varying the strength of this regularization term, $\lambda_{assoc}$. This observation indicates that, despite not achieving the highest performance, additional optimizer-level bias toward associativity improves performance. We detail this loss term in Appendix~\ref{app:associativity_regularization}.

\begin{figure}[thbp]
    \centering
    \includegraphics[]{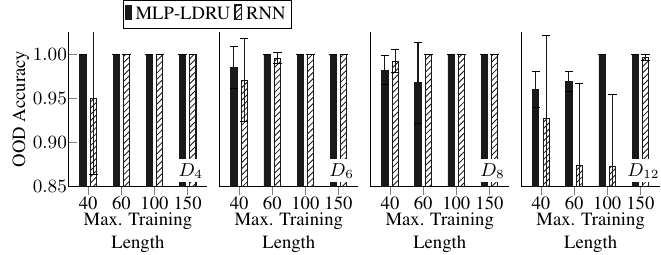}
    \caption{Trained MLP-LDRUs and RNNs tend to achieve higher OOD accuracy on the $D_n$ tasks as maximum training length increases.}
    \label{fig:sequence_len_exps}
\end{figure}

We present the results for \textbf{RQ2} in Fig.~\ref{fig:sequence_len_exps}. The OOD accuracy of both models tends to improve as the maximum training length increases. %
MLP-LDRU's performance on $D_{12}$ is consistent with its performance on the other $D_n$ tasks, whereas the RNN's performance on $D_{12}$ indicates that it is highly sensitive to weight initialization. This observation suggests that it may be easier to induce a monoid-like structure rather than a DFA-like structure for complex languages.

We present the results of \textbf{RQ3} in Table~\ref{tab:operator_performance}. MLP-LDRU achieves 100\% OOD accuracy on the four tasks. Of the tasks, Modular~Arithmetic appears to have the highest sensitivity to operator choice: with element-wise sum achieving 32.6\%, linear operator achieving 60.8\%, gated sum achieving 67.1\%, and GRC achieving 82.3\% OOD accuracy. In contrast, Parity~Check and Cycle~Navigation maintain 100.0\% OOD accuracy across all operators. Even~Pairs appears to be a task with intermediate sensitivity, with element-wise sum dropping to 51.9\% while the other choices maintain 100.0\% OOD accuracy. These results highlight that OOD generalization under the reduction depends considerably on the choice of operator.

\begin{table}[bhtp]
\centering
\small
\caption{OOD accuracies across different choices of operator.}
\label{tab:operator_performance}
\begin{tabular}{lcccc}
\toprule
\textbf{Operator} & \textbf{\begin{tabular}[c]{@{}c@{}}Even\\ Pairs\end{tabular}} & \textbf{\begin{tabular}[c]{@{}c@{}}Modular\\ Arithmetic\end{tabular}} & \textbf{\begin{tabular}[c]{@{}c@{}}Parity\\ Check\end{tabular}} & \textbf{\begin{tabular}[c]{@{}c@{}}Cycle\\ Navigation\end{tabular}} \\ \cmidrule(l){1-5}
E. Sum & $51.9\pm0.1$ & $32.6\pm0.3$ & $100.0\pm0.0$ & $100.0\pm0.0$ \\
Linear & $100.0\pm0.0$ & $60.8\pm1.7$ & $100.0\pm0.0$ & $100.0\pm0.0$ \\
G. Sum & $100.0\pm0.0$ & $67.1\pm2.9$ & $100.0\pm0.0$ & $100.0\pm0.0$ \\
GRC & $100.0\pm0.1$ & $82.3\pm 15.3$ & $100.0\pm0.0$ & $100.0\pm0.0$ \\
MLP (ours) & $100.0\pm0.0$ & $100.0\pm0.0$ & $100.0\pm0.0$ & $100.0\pm0.0$ \\ \bottomrule
\end{tabular}
\end{table}

\section{Discussion}
\label{sec:discussion}
Maximum training length is the primary driver of OOD performance for regular tasks. As the maximum training length increases, MLP-LDRU reaches 100\% OOD accuracy on $D_6$ by length 60 and 99.9--100.0\% on $D_8$ and $D_{12}$ by length 100 (Fig.~\ref{fig:sequence_len_exps}). The RNN shows a similar trend but with higher variability on $D_{12}$ (standard deviations $8.2$--$9.5\%$ up to length 100). These observations are consistent with the positional bias of recurrent networks, where earlier tokens must pass through longer computational paths. Learning $D_{12}$ requires the model to handle a greater number of possible state runs compared to the other $D_n$ tasks, which may increasingly challenge its capacity for long-term dependencies. We do not claim a definitive causal explanation here, but emphasize MLP-LDRU's more stable behavior under increased training lengths.

\begin{figure*}[!htbp]
    \centering
    \includegraphics[width=\textwidth]{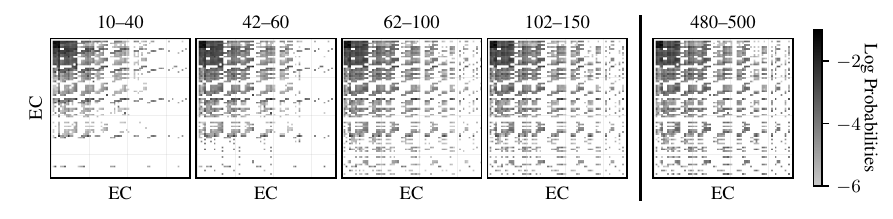}
    \caption{EC composition analysis for the $D_6$ syntactic monoid across different sequence length ranges. Each heatmap shows the log probability of compositions between ECs in the monoid. The sparse composition patterns in shorter sequences (lengths 10--40) explain why training with insufficient length leads to imperfect generalization, as it does not cover all compositions in the longer sequences (lengths 480--500).}
    \label{fig:monoid_compositions}
\end{figure*}

Fig.~\ref{fig:monoid_compositions} provides insight into why MLP-LDRU does not reach 100.0\% OOD accuracy on the $D_n$ languages when trained with insufficient sequence lengths. The heatmaps show the empirical probability of witnessing equivalence class (EC) compositions in the underlying monoid for even-length sequences from $D_6$. This task provides a useful compromise: it is among the harder cases where models do not immediately generalize, while remaining tractable to visualize. The leftmost heatmap reflects training sequences of length 10--40 and shows that many complex compositions are rarely observed, whereas the rightmost heatmap, taken from the longest OOD sequences we evaluate, displays a denser pattern. This sparsity gap is consistent with the empirical observations provided in Fig.~\ref{fig:sequence_len_exps}, where $D_6$ achieves $98.5\%$ OOD accuracy when trained on sequences up to length 40 but reaches $100\%$ once trained on sequences of length 60 and above. These findings suggest that limited training lengths mean the model is not sufficiently exposed to rare EC compositions, leading to imperfect generalization. We provide further results on the interpretability of sequence embeddings from trained MLP-LDRUs in Appendix~\ref{app:interpretability_analysis}. These results indicate that MLP-LDRU learns to cluster sequence embeddings according to their corresponding ECs in the syntactic monoid, further supporting the hypothesis that MLP-LDRU induces monoid-like structures to achieve length generalization.

On ListOps, the tree-based baselines are superior: BBT-GRC achieves the strongest overall extrapolation across the majority of length/depth/arity buckets. MLP-LDRU nevertheless remains competitive, surpassing LSTM and transformer baselines at 500k and 1M training examples; at 100k, transformer (ALiBi) is slightly stronger. These results suggest that (i) explicit hierarchical inductive bias is valuable for ListOps, and (ii) encouraging approximate associativity remains beneficial outside regular languages.

The LDRU's design reflects two key principles for length generalization. First, the reduction's structure gives all tokens equal computational depth, mitigating depth-related positional bias. Second, because we reuse an operator biased toward approximate associativity, length generalization reduces to learning the correct composition of pairs of subsequence embeddings.

Empirically, however, operator choice is critical for harder tasks. Only MLP-LDRU achieves 100\% OOD accuracy on Modular~Arithmetic, whereas even strictly associative elementwise-sum fails, suggesting that associativity alone is not sufficient and that there must be some additional consideration to enable consistent generalization. Furthermore, despite the similarity between the GRC and the MLP operator, the GRC is inconsistent across seeds on Modular~Arithmetic, indicating that it does not reliably learn approximately associative composition in this setting.

Beyond formal languages, Appendix~\ref{app:nlp_details} reports trained-from-scratch NLP classification experiments on GLUE~\citep{wang2018glue}, AG's~News, and DBPedia~\citep{Zhang2015CharacterlevelCN}. These results show that MLP-LDRU remains competitive with similarly trained Transformer baselines, and we leave pretrained and large-scale NLP evaluation to future work.

\section{Related Work}
\label{sec:related_work}
We position the LDRU within the context of related work.

\textbf{Tree-Based Recursive Neural Networks} BBT-RvNNs~\citep{munkhdalai-yu-2017-neural,chowdhury2023recursion} form a superset of LDRUs, but they differ significantly in their purpose. Typically, RvNNs~\citep{socher-etal-2013-recursive,tai-etal-2015-improved,yu-liu-2018-sliced,shi-etal-2018-tree,NEURIPS2019_d8e1344e} use a recursive cell to process sequences according to a binary tree that is either fixed, heuristically chosen, or induced from a linguistic prior. In contrast, an LDRU is explicitly designed to approximate a compositional algebra. This inductive bias is absent in standard recursive cells, which makes them less suitable for learning stable or repeatable composition rules.

More recent works, such as Recursion-in-Recursion (RIR)~\citep{chowdhury2023recursion}, also leverage log-depth recursion to achieve strong length generalization on complex algorithmic tasks, including ListOps. RIR is a framework for trading off speed with RvNN expressivity, via a two-level recursion using an RvNN (inner recursion) within an $k$-ary tree structure (outer recursion). The LDRU is distinct from these methods because the operators are biased toward learning associative structures. Approximate associativity allows the reduction to behave like a recurrence over the input sequence. This distinction matters for generalization: without an associativity bias, tree cells can learn composition rules that are not recurrence-like, whereas LDRUs are designed so that composing arbitrary subsequences under reduction is as consistent as possible. These algorithmic constraints align with our equivalence class analysis, explaining why MLP-LDRU generalized to $D_8$ and $D_{12}$ better as the maximum training length was increased.

\textbf{Length Generalization} Formal languages are a key test of systematic reasoning and generalization in sequence models~\citep{deletang2023neural,butoi2025training}. Classic RNNs and LSTMs tend to show alignment with regular languages~\citep{Merrill2020}, but using a state-based inductive bias limits efficient parallelization and introduces challenges with long-range dependencies~\citep{bengio1993the,bengio1994learning}. Our evaluation of the prefix languages highlights the RNN's difficulty in modeling long-range dependencies. transformers have well-documented length generalization problems~\citep{anil2022exploring,liu2023transformers,hahn2024sensitive,zhou2024what,zhou2024transformers,huang2025a,huang2025how}, and state space models~\citep{gu2022efficiently} exhibit similar limitations~\citep{fan2024advancing,sarrof2024the,terzic2025sdssm,grazzi2025unlocking}. Furthermore, large-scale empirical studies on state-of-the-art SSMs~\citep{terzic_2025_pdssm,walker2025structured} support that SSMs tend not to generalize to regular tasks. Our empirical evaluation also confirms this limitation, with the evaluated SSMs underperforming MLP-LDRU on all 21 regular tasks. Theoretical work has shown that transformers can solve specific regular tasks like Parity~Check~\citep{chiang2022overcoming}, but they are generally not suited to the structure of finite-state automata~\citep{Hahn2020}. However, recurrent transformers trade parallelism for improved length generalization~\citep{soulos2024recurrent,fan2025looped}. Of these architectures, RegularGPT~\citep{Chi2023} represents the closest approach to our work through adaptive weight sharing and sliding-window-dilated attention, which together implement the scan. However, layer norm parameters are not shared between adaptive layers, and empirically, RegularGPT's performance underperforms MLP-LDRU.

\textbf{Architecture Modifications} One line of research toward length generalization explores augmentations to existing architectures for improving length generalization. Approaches for improving generalization in transformers include scratchpad methods~\citep{nye2022show,kazemnejad2023the} that use intermediate reasoning steps, position coupling~\citep{cho2024position,mcleish2024transformers,cho2025arithmetic} that assign structure to positional encodings, and modifying positional encodings~\citep{press2022train,ruoss2023randomized}. In this work, our experiments focus on sequence classification or single-token prediction, where scratchpad methods and position coupling are not directly applicable, as they require autoregressive decoding. Prior evidence on randomized positional encodings shows that such modifications can yield task improvements, but they do not enable reliable generalization across regular tasks~\citep{ruoss2023randomized}, consistent with our evaluation on $\sim$RoPE. Other work has considered additional loss terms to encourage length generalization~\citep{butoi2025training}, observing that, again, its improvements are not robust across all tasks. This contrasts with MLP-LDRU, which achieves reliable generalization across all 21 regular tasks we evaluate.

\textbf{Reduction Applications} The reduction (and its more general algorithm, the scan~\citep{BlellochTR90}) has been used to accelerate existing machine learning methods, including RNNs~\citep{Martin2018} and backpropagation~\citep{Wang2020}. More recently, it has been used to more efficiently compute linear state updates in SSMs like S4~\citep{gu2022efficiently}, Mamba~\citep{gu2023mamba}, and S5~\citep{smith2023simplified}. These approaches have enabled fast, state-of-the-art performance in reinforcement learning contexts~\citep{lu2023structured}. Parallel DeltaNet~\citep{yang2024parallelizing} likewise accelerates linear recurrent updates but stays within a linear state-space formulation, and log-linear attention~\citep{guo2025log} achieves logarithmic depth by hierarchically expanding a linear RNN's state. While BBT-RvNNs fall within the prefix-scannable characterization of \citet{yau2025sequential}, our contribution concerns the design and analysis of a specific type of operator within that framework. Learnable monoids have also been proposed as aggregation functions over nodes in graphs~\citep{ong2022learnable}.

\section{Conclusion}
\label{sec:conclusion}

In this work, we make four contributions toward understanding length generalization in neural networks. We introduce LDRUs, a family of binary operators biased toward associativity, so that reduction can approximate a sequential recurrence. On 21 regular tasks, MLP-LDRU achieves 100\% OOD accuracy on 18 tasks when trained on lengths 1--40, and reaches at least 99.9\% on the remaining 3 when increasing the maximum training length. We complement these results with ListOps compositional extrapolation experiments and NLP classification benchmarks to probe behavior beyond regular tasks. We also (i) introduce \emph{prefix languages} to isolate long-range dependency handling under length extrapolation, (ii) provide an empirical monoid-based analysis connecting MLP-LDRU's generalization to exposure to equivalence class compositions, and (iii) report initial GPU benchmarks that quantify MLP-LDRU's practical trade-offs compared to classic RNNs and transformers.

Several research directions emerge from our findings. On the theory side, our $D_n$ results suggest that success requires sufficient coverage of equivalence class compositions, but we lack bounds on the minimum training lengths or sample complexity needed to achieve this coverage and a characterization of which compositions matter for a given task family. On the modeling side, an important next step is to adapt associativity-biased reduction to standard autoregressive objectives and test whether the learned composition operator that enables reliable length extrapolation on regular tasks can transfer to next-token prediction. More broadly, while regular languages provide exact verification, it remains to determine how far these ideas extend to richer compositional problems under distribution shift; larger and more diverse evaluations would clarify the scope and limitations of associativity as an inductive bias. Finally, monoid extraction from MLP-LDRU, e.g., identifying equivalence classes from embeddings and estimating the composition table, would provide a useful interpretability test of whether the model has learned the correct monoid structure.

\subsubsection*{Acknowledgments}
This work was supported by the UK EPSRC grant 2760033. %
We acknowledge computational resources and support provided by the Imperial College Research Computing Service. %

\bibliography{references}
\bibliographystyle{tmlr}


\appendix
\input{appendix.tex}

\end{document}

%% file: math_commands.tex
\usepackage{amsmath,amsfonts,bm}

\def\eqref#1{equation~\ref{#1}}

\def\1{\bm{1}}

\DeclareMathAlphabet{\mathsfit}{\encodingdefault}{\sfdefault}{m}{sl}
\SetMathAlphabet{\mathsfit}{bold}{\encodingdefault}{\sfdefault}{bx}{n}



%% file: tables/central_result.tex
\begin{table*}[!t]
\centering
\setlength{\tabcolsep}{4pt}
\caption{OOD accuracy results across 21 regular tasks. All architectures are trained on sequences with lengths 1--40 and evaluated on lengths 41--500. Results show mean $\pm$ standard deviation across seeds. $^{\downarrow}$ indicates MLP-LDRU performance is statistically significantly better ($p < 0.05$) than the baseline, while $^{\uparrow}$ indicates the baseline is statistically significantly better than MLP-LDRU.}
\label{tab:central_result}
\resizebox{\textwidth}{!}{
\begin{tabular}{@{}lccccccc@{}}
\toprule
Task &
  RNN &
  LSTM &
  Gated DeltaNet &
  TF (ALiBi) &
  TF ($\sim$RoPE) &
  BBT-GRC &
  MLP-LDRU \\ \midrule
Even~Pairs &
  ${77.4 \pm 12.2}$ &
  $100.0 \pm 0.0$ &
  ${52.2 \pm 0.8}^{\downarrow}$ &
  ${61.1 \pm 5.1}^{\downarrow}$ &
  $91.5 \pm 7.9$ &
  $100.0 \pm 0.0$ &
  $\mathbf{100.0 \pm 0.0}$ \\
Modular~Arithmetic &
  $100.0 \pm 0.0$ &
  $100.0 \pm 0.0$ &
  ${76.5 \pm 3.1}^{\downarrow}$ &
  ${54.4 \pm 4.5}^{\downarrow}$ &
  ${31.4 \pm 5.5}^{\downarrow}$ &
  $94.0 \pm 3.2$ &
  $\mathbf{100.0 \pm 0.0}$ \\
Parity~Check &
  $100.0 \pm 0.0$ &
  $100.0 \pm 0.0$ &
  ${53.4 \pm 0.3}^{\downarrow}$ &
  ${49.9 \pm 0.0}^{\downarrow}$ &
  ${49.9 \pm 0.0}^{\downarrow}$ &
  $100.0 \pm 0.0$ &
  $\mathbf{100.0 \pm 0.0}$ \\
Cycle~Navigation &
  $100.0 \pm 0.0$ &
  ${60.3 \pm 2.4}^{\downarrow}$ &
  ${25.8 \pm 3.8}^{\downarrow}$ &
  ${21.6 \pm 0.7}^{\downarrow}$ &
  ${23.3 \pm 1.5}^{\downarrow}$ &
  $100.0 \pm 0.1$ &
  $\mathbf{100.0 \pm 0.0}$ \\ \midrule
$D_2$ &
  $100.0 \pm 0.0$ &
  $100.0 \pm 0.0$ &
  ${100.0 \pm 0.0}$ &
  $100.0 \pm 0.0$ &
  $98.2 \pm 1.7$ &
  $100.0 \pm 0.0$ &
  $\mathbf{100.0 \pm 0.0}$ \\
$D_3$ &
  $100.0 \pm 0.0$ &
  $97.7 \pm 4.0$ &
  ${91.6 \pm 14.3}$ &
  $98.3 \pm 2.6$ &
  $92.5 \pm 10.2$ &
  $98.5 \pm 2.3$ &
  $\mathbf{100.0 \pm 0.0}$ \\
$D_4$ &
  $95.0 \pm 8.7$ &
  $100.0 \pm 0.0$ &
  ${93.1 \pm 6.0}$ &
  $96.0 \pm 4.7$ &
  ${97.6 \pm 0.4}^{\downarrow}$ &
  ${84.6 \pm 4.6}^{\downarrow}$ &
  $\mathbf{100.0 \pm 0.0}$ \\
$D_6$ &
  $97.1 \pm 4.7$ &
  $98.4 \pm 1.6$ &
  ${75.6 \pm 0.6}^{\downarrow}$ &
  $92.3 \pm 6.8$ &
  $98.4 \pm 1.0$ &
  ${84.7 \pm 6.0}^{\downarrow}$ &
  $\mathbf{98.5 \pm 2.4}$ \\
$D_8$ &
  $\mathbf{99.2 \pm 1.3}$ &
  ${89.0 \pm 4.3}$ &
  ${76.0 \pm 1.3}^{\downarrow}$ &
  $89.7 \pm 7.7$ &
  $98.7 \pm 0.8$ &
  $84.3 \pm 8.0$ &
  $98.1 \pm 1.6$ \\
$D_{12}$ &
  $92.7 \pm 9.5$ &
  ${82.1 \pm 1.7}^{\downarrow}$ &
  ${74.6 \pm 0.5}^{\downarrow}$ &
  $84.4 \pm 9.6$ &
  $\mathbf{98.7 \pm 1.0}^{\uparrow}$ &
  $90.5 \pm 0.7$ &
  ${96.0 \pm 2.1}$ \\ \midrule
Tomita~3 &
  $100.0 \pm 0.0$ &
  $100.0 \pm 0.0$ &
  ${99.9 \pm 0.1}$ &
  ${100.0 \pm 0.0}^{\downarrow}$ &
  ${70.7 \pm 8.9}^{\downarrow}$ &
  $99.4 \pm 0.7$ &
  $\mathbf{100.0 \pm 0.0}$ \\
Tomita~4 &
  $100.0 \pm 0.0$ &
  $100.0 \pm 0.0$ &
  ${99.1 \pm 0.2}^{\downarrow}$ &
  $100.0 \pm 0.0$ &
  ${85.5 \pm 2.1}^{\downarrow}$ &
  $99.4 \pm 1.1$ &
  $\mathbf{100.0 \pm 0.0}$ \\
Tomita~5 &
  $100.0 \pm 0.0$ &
  $100.0 \pm 0.0$ &
  ${75.1 \pm 0.4}^{\downarrow}$ &
  ${74.3 \pm 0.0}^{\downarrow}$ &
  ${74.3 \pm 0.0}^{\downarrow}$ &
  $100.0 \pm 0.0$ &
  $\mathbf{100.0 \pm 0.0}$ \\
Tomita~6 &
  $100.0 \pm 0.0$ &
  $100.0 \pm 0.0$ &
  ${54.7 \pm 0.5}^{\downarrow}$ &
  ${50.0 \pm 0.0}^{\downarrow}$ &
  ${50.0 \pm 0.0}^{\downarrow}$ &
  $100.0 \pm 0.0$ &
  $\mathbf{100.0 \pm 0.0}$ \\
Tomita~7 &
  $100.0 \pm 0.0$ &
  $100.0 \pm 0.0$ &
  ${100.0 \pm 0.0}$ &
  $100.0 \pm 0.0$ &
  $100.0 \pm 0.0$ &
  $100.0 \pm 0.0$ &
  $\mathbf{100.0 \pm 0.0}$ \\ \midrule
$P_{1,2}$ &
  ${82.2 \pm 3.3}^{\downarrow}$ &
  $100.0 \pm 0.0$ &
  ${52.4 \pm 1.7}^{\downarrow}$ &
  ${56.0 \pm 2.3}^{\downarrow}$ &
  $89.2 \pm 12.7$ &
  $100.0 \pm 0.0$ &
  $\mathbf{100.0 \pm 0.0}$ \\
$P_{2,2}$ &
  ${65.4 \pm 24.0}$ &
  $100.0 \pm 0.0$ &
  ${22.0 \pm 11.9}^{\downarrow}$ &
  $93.0 \pm 5.6$ &
  ${86.4 \pm 5.0}^{\downarrow}$ &
  $99.9 \pm 0.1$ &
  $\mathbf{100.0 \pm 0.0}$ \\
$P_{4,2}$ &
  $99.4 \pm 1.0$ &
  $100.0 \pm 0.0$ &
  ${18.9 \pm 1.6}^{\downarrow}$ &
  $90.0 \pm 10.3$ &
  ${58.5 \pm 11.1}^{\downarrow}$ &
  $100.0 \pm 0.0$ &
  $\mathbf{100.0 \pm 0.0}$ \\
$P_{1,4}$ &
  ${61.4 \pm 15.7}$ &
  $100.0 \pm 0.0$ &
  ${30.1 \pm 1.6}^{\downarrow}$ &
  ${40.3 \pm 2.4}^{\downarrow}$ &
  ${79.5 \pm 7.0}^{\downarrow}$ &
  $100.0 \pm 0.0$ &
  $\mathbf{100.0 \pm 0.0}$ \\
$P_{2,4}$ &
  $96.9 \pm 2.5$ &
  $100.0 \pm 0.0$ &
  ${21.5 \pm 9.5}^{\downarrow}$ &
  ${31.0 \pm 8.1}^{\downarrow}$ &
  ${51.6 \pm 7.6}^{\downarrow}$ &
  $100.0 \pm 0.0$ &
  $\mathbf{100.0 \pm 0.0}$ \\
$P_{4,4}$ &
  ${85.0 \pm 15.0}$ &
  ${97.3 \pm 2.2}$ &
  ${69.4 \pm 6.0}^{\downarrow}$ &
  ${21.3 \pm 7.6}^{\downarrow}$ &
  ${58.2 \pm 2.7}^{\downarrow}$ &
  $100.0 \pm 0.0$ &
  $\mathbf{100.0 \pm 0.0}$ \\ \bottomrule
\end{tabular}
}
\end{table*}

%% file: tables/listops_summary.tex

\begin{table*}[ht]
\centering
\small
\caption{ListOps test accuracy (\%). Left block reports mean accuracy averaged across all reported sequence-length bins. Right block reports long-range accuracy on the longest reported bin (60--80 for (3,9), 101--200 for (3,14), (5,9), and (5,14)). Higher is better. Bins are labeled ($x$, $y$) where $x$ is maximum depth and $y$ is maximum arity.}
\label{tab:listops_summary}
\begin{tabular}{lcccccccc}
\toprule
& \multicolumn{4}{c}{\textbf{Mean Accuracy (\%)}} & \multicolumn{4}{c}{\textbf{Longest Bin (\%)}} \\
\cmidrule(lr){2-5}\cmidrule(l){6-9}
\textbf{Architecture}
& \textbf{(3, 9)}
& \textbf{(3, 14)}
& \textbf{(5, 9)}
& \textbf{(5, 14)}
& \textbf{(3, 9)}
& \textbf{(3, 14)}
& \textbf{(5, 9)}
& \textbf{(5, 14)} \\
\midrule
RIR-GRC    & 89.9 & 79.1 & 48.1 & 51.6 & 79.2 & 66.4 & 36.4 & 38.8 \\
BBT-GRC    & 84.1 & 79.1 & 49.6 & 53.1 & 72.5 & 67.7 & 39.2 & 41.4 \\
MLP-LDRU   & 74.7 & 69.7 & 45.9 & 49.0 & 67.8 & 65.2 & 38.8 & 41.5 \\
TF (ALiBi) & 68.2 & 63.3 & 39.0 & 41.7 & 55.1 & 46.4 & 22.8 & 25.8 \\
LSTM       & 65.9 & 62.7 & 44.9 & 48.1 & 50.0 & 51.4 & 36.0 & 38.9 \\
TF (NoPE)  & 37.3 & 37.6 & 29.9 & 32.5 & 32.1 & 31.2 & 22.7 & 24.2 \\
TF (Sin.)  & 35.9 & 25.5 & 21.5 & 22.9 & 9.8 & 8.4  & 8.7  & 8.7 \\
\bottomrule
\end{tabular}
\end{table*}

%% file: appendix.tex
\raggedbottom

\section{Connection to Automata Theory}
\label{app:connection_theory}

This section provides the necessary automata theory to discuss the \emph{monoid} representation of deterministic finite automata, directly motivating our classification of LDRU operators; \citet{sakarovitch2009elements} provides a thorough introduction to this area.

An alphabet $\Sigma$ is a finite set of symbols. A sequence is a finite concatenation of symbols from $\Sigma$, and $\Sigma^\ast$ denotes the set of all sequences. A language is a subset of $\Sigma^\ast$.

A \emph{deterministic finite automaton} (DFA) is defined as $\mathcal{A} = (Q, \Sigma, \delta, q_I, F)$, where $Q$ is a finite set of states, $\delta$ is a transition function $\delta : Q \times \Sigma \rightarrow Q$, $q_I \in Q$ is an initial state, and $F \subseteq Q$ is a set of accepting states. $\delta(q, s) = r$ denotes a transition from $q$ to $r$ labeled with $s$. %
The $(Q, \Sigma, \delta)$ component of a DFA is a \emph{semiautomaton}. A run is a sequence of states $q_0, q_1, \dots, q_n$ where there is a transition between each pair of consecutive states. A sequence is accepted by a DFA if and only if it induces a run from $q_I$ to a state in $F$ using $\delta$; otherwise, the sequence is rejected. The language a DFA recognizes is the sequences it accepts, and DFAs are equivalent when they recognize the same language. A DFA is minimal if no equivalent DFA has fewer states. A language is regular when it can be recognized by a DFA~\citep{sakarovitch2009elements}.

Transduction tasks, like evaluating Modular~Arithmetic expressions, require replacing $F$ in the DFA with an output alphabet $S$ and a function mapping states to output symbols $g:Q \rightarrow S$. This replacement transforms a DFA into a \emph{Moore machine}, i.e., $(Q, \Sigma, S, \delta, q_I, g)$~\citep{sakarovitch2009elements}. The final state of the run induced by processing a sequence is passed to $g$ to produce that sequence's output symbol. For consistency, we treat all tasks as transduction tasks by converting recognition tasks into Moore machines with $S=\{0, 1\}$ and $g(q) = 1$ if $q \in F$ and $g(q) = 0$ otherwise.

A \emph{monoid} is $(M, \odot, \epsilon)$, where $M$ is a set, $\odot:M\times M\rightarrow M$ is an associative binary operator and $\epsilon$ is a neutral element for $\odot$~\citep{sakarovitch2009elements}. 
Let $\mathcal{A}$ be a DFA that recognizes language $L$. Every sequence $w \in \Sigma^\ast$ induces a function $e:Q\rightarrow Q$ in $\mathcal{A}$. The \emph{transition monoid} $(E, \odot, \epsilon)$ of $\mathcal{A}$ is the monoid of $e$ functions induced by $\Sigma^\ast$ in $\mathcal{A}$. The monoid operator works by composing two state mappings into a new state mapping. We refer to $e$ functions as \emph{equivalence classes} (ECs) as there exists a canonical morphism $\varphi: \Sigma^\ast \rightarrow E$ \citep{holzer2004deterministic}. We can process a sequence $w \in \Sigma^\ast$ by computing $\varphi (w) (q_I) \in F$. Transition monoids are equivalent representations of semiautomata. The \emph{syntactic monoid} is the transition monoid of the minimal DFA that recognizes $L$.

It is not practically feasible to process a sequence $w \in \Sigma^\ast$ with $\varphi(w)$ directly as its domain is infinite, but we can derive the sequence's EC compositionally. We compose ECs of subsequences using $\odot$. As $\odot$ is associative, any composition order of ECs will result in a correct evaluation of $\varphi(w)$. %

For example, let $w = w_1\dots w_n$ be a sequence and assume our task is to decide if $w$ is accepted by the regular language $L$ with minimal DFA $(Q, \Sigma, \delta, q_I, F)$ and syntactic monoid $(E, \odot, \epsilon)$. Suppose instead of $\varphi$ we have $\varphi': \Sigma \rightarrow E$. If we apply $\varphi'$ to each $w_i$, then we obtain a sequence of monoid elements $\varphi(w) = \varphi'(w_1) \odot \dots \odot \varphi'(w_n) = e_1 \odot \dots \odot e_n$ with $e_i \in E$. We can efficiently leverage the associativity to process regular languages using the reduction~\citep{data_parallel_algorithms}.

The reduction algorithm takes an associative binary operator $\odot$ to reduce a sequence of elements $e_1, \dots, e_n$ into a single result $e_1 \odot \dots \odot e_n$. This result can be computed efficiently using a balanced binary tree~\citep{BlellochTR90} (see Fig.~\ref{fig:info-flow}). Processing sequences with the reduction directly motivates the LDRU family: parameterized binary operators that learn to approximate a reduction operator $\odot$.

\section{Prefix Languages Definition}
\label{app:prefix_languages_definition}

The prefix language, $P_{p,q}$, with a prefix length of $p$ over $q$ symbols is a Moore machine defined as:
\begin{align*}
    P_{p,q} = (&Q = \left\{0, 1, 2, \ldots, \frac{q^{p+1} - 1}{q - 1} - 1\right\}, \\&\Sigma = \left\{0, \ldots, q - 1\right\}, \\&S = \left\{0, \ldots, q^p\right\}, \\&\delta = \delta_{p, q}, \\&I=0, \\&g = g_{p,q}),
\end{align*} where $g_{p,q}(i) = 0$ when $i\in\left\{0,\ldots, \frac{q^p - 1}{q - 1} - 1\right\}$ and $i - (\frac{q^p - 1}{q - 1} - 1)$ otherwise. The transition function $\delta_{p,q}$ is $\delta(i,j)= iq + 1 + j$ when $i \in \left\{0, \ldots, \frac{q^p - 1}{q-1} - 1\right\}$. Otherwise, $\delta(i,j) = i\ \forall j \in \Sigma$. The output function $g_{p,q}$ ensures that the first $\frac{q^p - 1}{q - 1} - 1$ states output 0, and the remaining states output their state number minus $\frac{q^p - 1}{q - 1} - 1$.

\textbf{Worked Example} To illustrate how to use this definition to construct a prefix language, we consider the case of $P_{4,2}$, i.e., the prefix language with a prefix length of 4 over an alphabet of size 2. For these values, we calculate the associated $p,q$ based terms for $P_{4,2}$: $q-1 = 1$, $q^p = 16$, $\frac{q^{p+1} - 1}{q - 1}-1 = 30$ and $\frac{q^p - 1}{q - 1}-1 = 14$. To provide clarity on these terms:
\begin{enumerate}
\item the first is the maximum integer present in the alphabet, $\Sigma = \{0, 1\}$;
\item the second is the maximum integer present in the output alphabet, $S=\{0,\dots,16\}$ (we do not subtract 1 here because 0 is the invalid prefix symbol and the remainder are the possible different prefixes);
\item the third is the maximum state, $Q = \{0, 1, \ldots, 30\}$ (for a total of 31 states, $\sum_{i=0}^{4} q^i = 31$);
\item the fourth is the term controlling the output function, ensuring that the output is 0 for the first 15 states and the output is the state number minus 14 for the remaining states. It also controls the transition function, ensuring that the first 15 states can transition to the following states while reading the first $p$ symbols, and defines self-transitions when reading any further symbols.
\end{enumerate}
  These terms lead us to the Moore machine for $P_{4,2}$ as follows:
\begin{align*}
    P_{4,2} = (&Q = \{0, \dots, 30\}, \\&\Sigma = \{0, 1\}, \\&S = \{0, \dots, 16\}, \\&\delta = \delta_{4,2}, \\&I=0, \\&g = g_{4,2}),
\end{align*} where $g_{4,2}(i) = 0$ when $i\in\{0,\ldots, 14\}$ and $i - 14$ otherwise (i.e., when $i \in \{15,\dots,30\}$). The transition function $\delta_{4,2}$ is defined as $\delta(i,j)= 2i + 1 + j$ when $i \in \{0, \ldots, 14\}$ and $i$ otherwise. For example, $\delta_{4,2}(0, 0)= 1$ and $\delta_{4,2}(0, 1)= 2$. For state 1, $\delta_{4,2}(1, 0)= 3$ and $\delta_{4,2}(1, 1)= 4$.

\section{MLP-LDRU Details}
\label{app:ldru_operator_details}

\begin{figure}[!htbp]
    \centering
    \begin{subfigure}[b]{0.52\textwidth}
        \centering
        \small
        \begin{tikzpicture}[
            box/.style={rectangle,draw,minimum width=1cm,minimum height=0.8cm, fill=white!20},
            arrow/.style={->,>=stealth}
        ]
        \node[box, align=center, text width=1cm] (x) at (-2.0,1) {$\mathbf{h}_i$};
        \node[box, align=center, text width=1cm] (y) at (2.0,1) {$\mathbf{h}_j$};
        \node[box, align=center, text width=2.4cm] (mlp) at (0,0) {MLP($[\mathbf{h}_i; \mathbf{h}_j]$)};
        \node[box, align=center, text width=1.25cm] (wx) at (-1,-1.33) {$\mathbf{g}_i$};
        \node[box, align=center, text width=1.25cm] (wy) at (1,-1.33) {$\mathbf{g}_j$};
        \node[box, align=center, text width=1.25cm] (wxx) at (-1,-2.67) {$\mathbf{g}_i \circ \mathbf{h}_i$};
        \node[box, align=center, text width=1.25cm] (wyy) at (1,-2.67) {$\mathbf{g}_j \circ \mathbf{h}_j$};
        \node[box, align=center, text width=3.0cm] (linx) at (-2,-4) {$\mathbf{f}_i=\mathbf{V}_i(\mathbf{g}_i \circ \mathbf{h}_i) + \mathbf{b}_i$};
        \node[box, align=center, text width=3.0cm] (liny) at (2,-4) {$\mathbf{f}_j=\mathbf{V}_j(\mathbf{g}_j \circ \mathbf{h}_j) + \mathbf{b}_j$};
        \node[box] (sum) at (0,-5.33) {$\mathbf{W}(\mathbf{f}_i + \mathbf{f}_j) + \mathbf{b}_{\text{out}}$};

        \draw[arrow] (x) -- (0, 1) -- (mlp);
        \draw[arrow] (y) -- (0, 1) -- (mlp);
        \draw[arrow] (mlp) -- (wx);
        \draw[arrow] (mlp) -- (wy);
        \draw[arrow] (x) -- (-2, -2.67) -- (wxx);
        \draw[arrow] (y) -- (2, -2.67) -- (wyy);
        \draw[arrow] (wx) -- (wxx);
        \draw[arrow] (wy) -- (wyy);
        \draw[arrow] (wxx) -- (linx);
        \draw[arrow] (wyy) -- (liny);
        
        \draw[arrow] (linx) -- (sum);
        \draw[arrow] (liny) -- (sum);
        
        \end{tikzpicture}
        \caption{Implementation of the MLP-LDRU operator showing the MLP gating mechanism with element-wise multiplication and separate linear projections.}
        \label{fig:monoid-operator}
    \end{subfigure}
    \hfill
    \begin{subfigure}[b]{0.47\textwidth}
        \centering
        \small
        \begin{tikzpicture}[
            box/.style={rectangle,draw,minimum width=2.5cm,minimum height=0.8cm, fill=white!20},
            arrow/.style={->,>=stealth}
        ]
        \node[box, align=center, text width=2.4cm] (embedding) at (0,0) {Embed sequence};
        \node[box, align=center, text width=2.5cm] (window) at (0,-1.33) {Apply $\odot_\theta$ to non-overlapping consecutive pairs};
        \node[box, align=center, text width=2.4cm] (res) at (0,-2.67) {Residual FFN};
        \node[box] (norm) at (0,-4) {Layer normalization};
        \node[box] (dropout) at (0,-5.33) {Dropout};
        \node[box, align=center, text width=2.4cm] (output) at (0,-6.67) {Linear projection};
        
        \draw[arrow] (embedding) -- (window);
        \draw[arrow] (window) -- (res);
        \draw[arrow] (res) -- (norm);
        \draw[arrow] (norm) -- (dropout);
        \draw[arrow] (dropout) -- (output);
        
        \draw[->] (dropout.east) -- (2,-5.33) -- (2, -3.33) node[right, align=center, text width=2.5cm,] {Repeat for $\lceil \log_2 n \rceil$ steps} -- (2,-1.33) -- (window.east) ;
        \end{tikzpicture}
        \caption{Complete MLP-LDRU model processing of an $n$-length sequence, showing the reduction implementation with residual connections and normalization.}
        \label{fig:pp-reduction}
    \end{subfigure}
    \caption{MLP-LDRU architecture diagrams showing (a) operator implementation, and (b) complete processing pipeline.}
    \label{fig:architecture}
\end{figure}
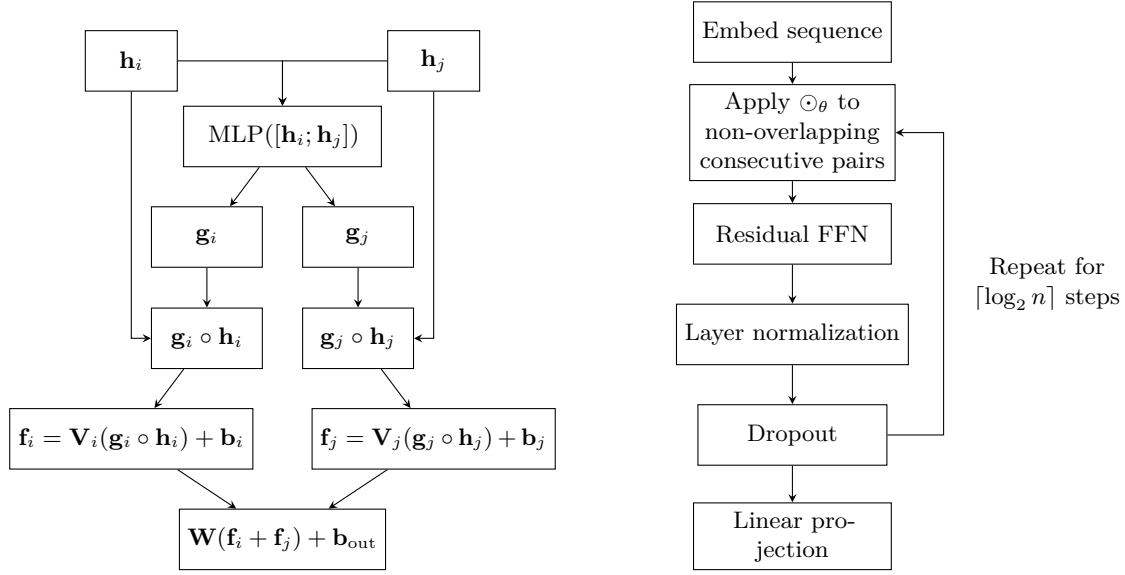

We provide further details of MLP-LDRU, $\odot_\theta$, as described in the main text. The operator is designed to compute a weighted combination of two input embeddings $\mathbf{h}_i, \mathbf{h}_j \in \mathbb{R}^d$ using a three-layer MLP with gating vectors. We provide reference diagrams in Fig.~\ref{fig:architecture}.

\textbf{Residual FFN} Within each reduction step, we apply a two-layer feedforward network to the post-$\odot_\theta$ embeddings and add a residual connection, i.e., $\mathbf{h}\leftarrow \mathbf{h} + \mathrm{FFN}(\mathbf{h})$, followed by layer normalization and dropout. The FFN has expansion $4d$ and output $d$. We use ReLU activation for regular-language experiments and SiLU activation~\citep{elfwing2018sigmoid} for non-regular tasks. This FFN is shared across steps.

The MLP used to produce the gating vectors from the concatenated embeddings $[\mathbf{h}_i; \mathbf{h}_j] \in \mathbb{R}^{2d}$ is structured as follows. The MLP consists of three linear layers with ReLU activations in between, and it outputs two gating vectors $\mathbf{g}_i, \mathbf{g}_j \in \mathbb{R}^d$, one for each input embedding. The hidden dimensions of the layers are ($2d, 4d, 2d$). We expand the MLP function used in Eq.~\ref{eq:mlp} for further clarity. The MLP produces gating vectors given two input embeddings $\mathbf{h}_i, \mathbf{h}_j \in \mathbb{R}^d$ as follows:
\begin{align}
\mathbf{x} &= [\mathbf{h}_i; \mathbf{h}_j] \in \mathbb{R}^{2d} \\
\mathbf{z}_1 &= \mathbf{W}_1 \mathbf{x} + \mathbf{b}_1 \in \mathbb{R}^{2d} \\
\mathbf{z}_2 &= \text{ReLU}(\mathbf{z}_1) \\
\mathbf{z}_3 &= \mathbf{W}_2 \mathbf{z}_2 + \mathbf{b}_2 \in \mathbb{R}^{4d} \\
\mathbf{z}_4 &= \text{ReLU}(\mathbf{z}_3) \\
[\mathbf{g}_i; \mathbf{g}_j] &= \mathbf{W}_3 \mathbf{z}_4 + \mathbf{b}_3 \in \mathbb{R}^{2d}.
\end{align}

We initialize the weights of the MLP and linear projections using standard techniques to ensure effective training. The MLP weights are initialized using Glorot normal initialization~\citep{glorot2010understanding}. The projections $\mathbf{V}_i, \mathbf{V}_j$ and the output projection $\mathbf{W}_{\text{out}}$ are initialized to the identity matrix. All biases are initialized to zero, enabling the model to behave as a gated element-wise sum initially.

\section{Additional Complexity Analysis}
\label{app:computational_complexity}
We provide a detailed computational complexity analysis of MLP-LDRU compared to RNNs and transformers. We analyze complexity in terms of:
\begin{description}
\item[\textbf{Work complexity}]Total number of operations required to process a sequence.
\item[\textbf{Depth complexity}]Longest chain of sequential operations when processing a sequence (determines parallelizability).
\item[\textbf{Memory complexity}]Total memory required to store parameters and intermediate results for processing a sequence.
\end{description}

We consider a sequence of length $n$ and embedding dimension $d$. MLP-LDRU processes the sequence using $n-1$ $\odot_\theta$ operations, each requiring $O(d^2)$ work due to the MLP computation. The total work complexity is therefore $O(n d^2)$. The reduction processes sequences in exactly $\lceil \log_2 n \rceil$ steps, giving it depth $O(\log n)$ because $\lceil \log_2 n \rceil \leq \log_2(2n) = 1 + \log_2(n)$. The parameters of $\odot_\theta$ are $O(d^2)$, leading to a total memory complexity of $O(n d + d^2)$ when including the storing of intermediate embeddings. Note that complexities reflect standard implementations; attention variants beyond dot-product attention can improve transformer scaling.

\begin{table}[!htbp]
\centering
\begin{tabular}{lccc}
\toprule
Architecture & Work & Depth & Memory \\
\midrule
RNN & $O(n d^2)$ & $O(n)$ & $O(d^2)$ \\
LSTM & $O(n d^2)$ & $O(n)$ & $O(d^2)$ \\
Transformer & $O(n^2 d + n d^2)$ & $O(1)$ & $O(n^2 + n d + d^2)$ \\
\textbf{MLP-LDRU} & $\mathbf{O(n d^2)}$ & $\mathbf{O(\log n)}$ & $\mathbf{O(n d + d^2)}$ \\
\bottomrule
\end{tabular}
\caption{Complexity comparison across architectures.}
\end{table}

We give the experimental setup to benchmark the practical scaling of the RNN, MLP-LDRU, and transformer presented in Fig.~\ref{fig:scaling_plots}. We used a batch size of 32, input size of 16, output size of 2 across all models. We set the hyperparameters for each scale to give models similar parameter counts, and we present their configurations with the corresponding number of model parameters in Table~\ref{tab:model_configs_scaling}. We measured the wall-clock time for forward and backward passes across sequence lengths from 4 to 2048 on an NVIDIA RTX 6000 Ada Generation GPU. Runtimes exclude initialization overhead and were measured after warm-up. To obtain stable measurements, we averaged runtimes over 128 passes for each sequence length.

\begin{table*}[bhtp]
\centering
\caption{Model configurations used in the architecture ablation. Hyperparameters were selected to produce approximately matched parameter counts across architectures at each scale.}
\label{tab:model_configs_scaling}
\begin{tabular}{lcc cc cc}
\toprule

& \multicolumn{2}{c}{Small}
& \multicolumn{2}{c}{Medium}
& \multicolumn{2}{c}{Large} \\

\cmidrule(lr){2-3}
\cmidrule(lr){4-5}
\cmidrule(lr){6-7}

Model & Config & Params & Config & Params & Config & Params \\

\midrule

RNN
& $d=400$ & 168k
& $d=1024$ & 1.07M
& $d=1600$ & 1.4M \\

MLP-LDRU
& $d=64$ & 162k
& $d=160$ & 1.01M
& $d=256$ & 1.5M \\

Transformer
& 3 layers, $d=64$ & 150k
& 5 layers, $d=128$ & 991k
& 6 layers, $d=192$ & 1.6M \\

\bottomrule
\end{tabular}
\end{table*}

\begin{figure}[htbp]
    \centering
    \begin{subfigure}[t]{0.47\columnwidth}
        \centering
        \includegraphics[width=\linewidth]{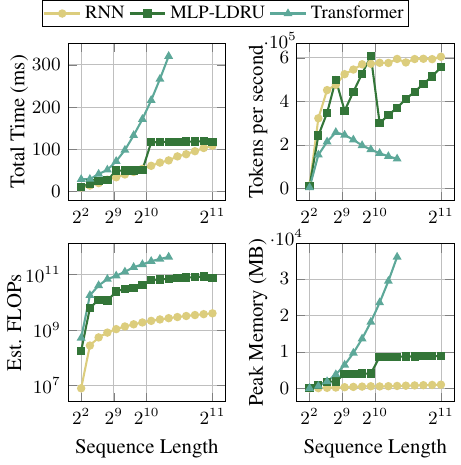}
        \caption{Medium scale}
        \label{fig:medium_scaling_appendix}
    \end{subfigure}
    \hfill
    \begin{subfigure}[t]{0.47\columnwidth}
        \centering
        \includegraphics[width=\linewidth]{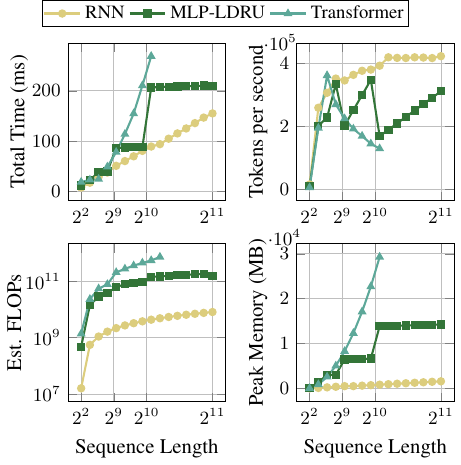}
        \caption{Large scale}
        \label{fig:large_scaling_appendix}
    \end{subfigure}

    \caption{Additional empirical runtime analysis of MLP-LDRU, RNN, and transformer across increased parameter scales.}
    \label{fig:scaling_plots_appendix}
\end{figure}

\section{Regular Task Experimental Details}
\label{app:experimental_details}

This section provides detailed information about the experimental setup, including computing infrastructure, dataset generation procedures, and model hyperparameters to ensure reproducibility. Our codebase is fully implemented in JAX~\citep{jax2018github,deepmind2020jax} in Haiku~\citep{haiku2020github}. The experiments were executed across different clusters consisting of NVIDIA RTX 6000 Ada Generation, NVIDIA L40S, and NVIDIA A100 80GB GPUs.

\textbf{Statistical Analysis} To assess the statistical significance of performance differences between MLP-LDRU and baseline models across tasks, we performed single t-tests on baselines where MLP-LDRU has 100.0\% generalization with zero deviation; otherwise, we conducted a paired t-test comparing the OOD accuracies of both methods.

{\textbf{Task Details} We present the task details and corresponding automata representations in Table~\ref{tab:deletang_tasks} and Table~\ref{tab:bhattamishra_tasks}. We present all tasks as Moore machines for consistency, as described in the main text. The tasks are selected from \citet{deletang2023neural} (the regular tasks) and \citet{Bhattamishra2020OnTA}, which are well-known benchmarks for evaluating length generalization in sequence processing models. The prefix languages evaluated can be constructed using the definition provided in Appendix~\ref{app:prefix_languages_definition}.

\tikzset{
->, 
>=stealth, 
minimum size=0.5cm, 
node distance=0.1cm, 
every state/.style={thick, fill=gray!10}, 
initial text=$ $, 
every axis/.style={-},
}

\begin{table}[t]
\centering
\caption{Descriptions of the Even Pairs, Modular Arithmetic, Parity Check, and Cycle Navigation tasks with diagrams.}
\label{tab:deletang_tasks}
\begin{tabular}{p{2.5cm} p{4cm} p{6cm}}
\toprule
\textbf{Task} & \textbf{Description} & \textbf{Automaton} \\
\midrule

Even Pairs & Determine if input sequence has an even number of $01$ and $10$ pairs. &
\begin{tikzpicture}[scale=0.8, every node/.style={transform shape}, baseline=(current bounding box.center)]
\node[state, initial] (q0) {$0/\epsilon$};
\node[state, above=0.1cm of q0, xshift=1.5cm] (q1) {$1/0$};
\node[state, right=0.5cm of q1] (q2) {$2/1$};
\node[state, below=0.1cm of q0, xshift=1.5cm] (q3) {$3/0$};
\node[state, right=0.5cm of q3] (q4) {$4/1$};
\path[->] 
(q0) edge[] node[above] {0} (q1)
(q0) edge[] node[below] {1} (q3)
(q1) edge[loop above] node {0} (q1)
(q1) edge[bend left] node[above] {1} (q2)
(q2) edge[bend left] node[below] {0} (q1)
(q2) edge[loop above] node {1} (q2)
(q3) edge[loop below] node {1} (q3)
(q3) edge[bend left] node[above] {0} (q4)
(q4) edge[bend left] node[below] {1} (q3)
(q4) edge[loop below] node {0} (q4);
\end{tikzpicture} \\

Modular Arithmetic (mod 5) & Evaluate the input sequence modulo 5. &
Presented separately in Fig.~\ref{fig:mod_arith_automaton} due to its complexity. \\

Parity Check & Determine if input sequence has an even number of 1s. State $0$ represents even parity, state $1$ represents odd parity. & 
\begin{tikzpicture}[scale=0.8, every node/.style={transform shape}, baseline=(current bounding box.center)]
\node[state, initial] (q0) {$0/1$};
\node[state, right=0.5cm of q0] (q1) {$1/0$};
\path[->] 
(q0) edge[bend left] node[above] {1} (q1)
(q1) edge[bend left] node[below] {1} (q0)
(q0) edge[loop above] node {0} (q0)
(q1) edge[loop above] node {0} (q1);
\end{tikzpicture} \\

Cycle Navigation & Navigate a cycle of 5 states based on the input sequence. & %
\begin{tikzpicture}[scale=0.8, every node/.style={transform shape}, baseline=(current bounding box.center)]
\small
\node[state, initial] (q0) {$0/0$};
\node[state, right=0.75cm of q0, yshift=1.75cm] (q1) {$1/1$};
\node[state, right=1.0cm of q1, yshift=-0.6cm] (q2) {$2/2$};

\node[state, right=0.75cm of q0, yshift=-2.0cm] (q4) {$4/4$};
\node[state, right=1.1cm of q4, yshift=0.9cm] (q3) {$3/3$};
\path[->] 
(q0) edge[loop right] node[right] {0} (q0)
(q1) edge[loop above] node[above] {0} (q1)
(q2) edge[loop above] node[above] {0} (q2)
(q3) edge[loop right] node[right] {0} (q3)
(q4) edge[loop above] node[above] {0} (q4)

(q0) edge[bend left=25] node[auto=left] {1} (q1)
(q1) edge[bend left=25] node[auto=left] {1} (q2)
(q2) edge[bend left=25] node[auto=left] {1} (q3)
(q3) edge[bend left=25] node[auto=left] {1} (q4)
(q4) edge[bend left=25] node[auto=left] {1} (q0)
(q0) edge[bend left=25] node[auto=left] {$-1$} (q4)
(q1) edge[bend left=25] node[auto=left] {$-1$} (q0)
(q2) edge[bend left=25] node[auto=left] {$-1$} (q1)
(q3) edge[bend left=25] node[auto=left] {$-1$} (q2)
(q4) edge[bend left=25] node[auto=left] {$-1$} (q3);
\end{tikzpicture} \\

\bottomrule
\end{tabular}
\end{table}

\begin{figure}[!htbp]
\centering
\includegraphics[]{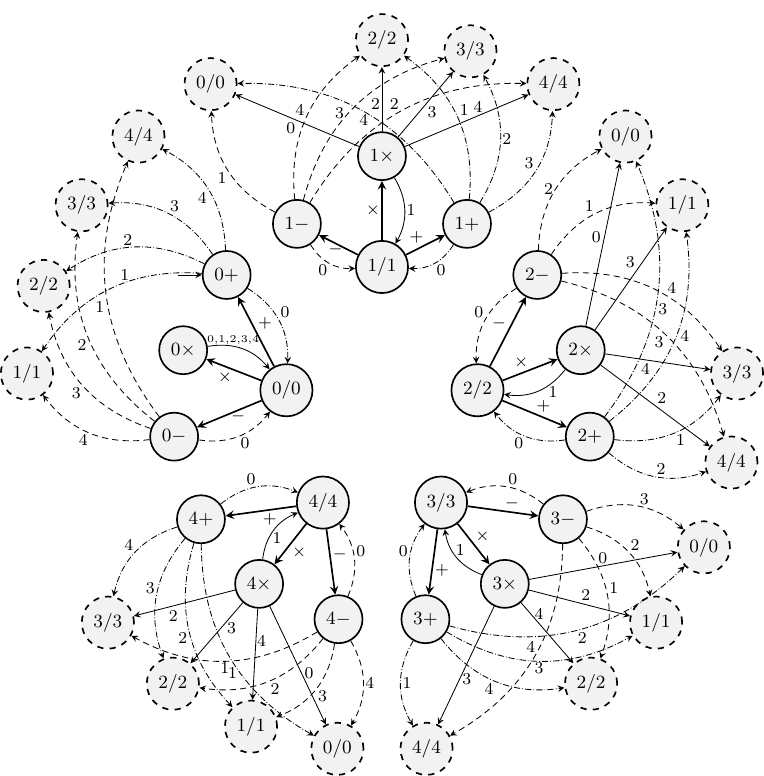}
\caption{Moore machine for evaluating Modular~Arithmetic modulo 5 expressions. The automaton's initial state is $0+$. The sequences are constrained to valid arithmetic expressions, and the automaton handles multiplication, addition, and subtraction. Dashed states indicate symbolically linked states (i.e., the inner states) for presentation clarity.}
\label{fig:mod_arith_automaton}
\end{figure}

\begin{table}[!t]
\centering
\caption{Descriptions of the $D_n$ and Tomita tasks with diagrams. The Tomita languages are originally sourced from \citet{tomita1982dynamic} and their diagrams are adapted from \citet{fdezdelpozoromero2023gradient}. All languages in this table are defined over the binary alphabet $\{0, 1\}$.}
\label{tab:bhattamishra_tasks}
\begin{tabular}{p{2.5cm} p{4cm} p{6cm}}
\toprule
\textbf{Task} & \textbf{Description} & \textbf{Automaton} \\
\midrule
$D_n$ & Recognize sequences that belong to the regular expression $(0D_{n-1}^\ast 1)^\ast$ where $D_1 = (01)^\ast$. The diagram recognizes $D_3$. &
\begin{tikzpicture}[scale=0.8, every node/.style={transform shape}, baseline=(current bounding box.center)]
\small
\node[state, initial] (q0) {$0/1$};
\node[state, right=0.5cm of q0] (q1) {$1/0$};
\node[state, right=0.5cm of q1] (q2) {$2/0$};
\node[state, right=0.5cm of q2] (q3) {$3/0$};
\path[->] 
(q0) edge[bend left] node[auto=left] {0} (q1)
(q1) edge[bend left] node[auto=left] {0} (q2)
(q2) edge[bend left] node[auto=left] {0} (q3)
(q1) edge[bend left] node[auto=left] {1} (q0)
(q2) edge[bend left] node[auto=left] {1} (q1)
(q3) edge[bend left] node[auto=left] {1} (q2);
\end{tikzpicture} \\

Tomita~3 & Recognize sequences where there is no odd-length $0$ consecutive subsequence after an odd-length $1$ consecutive subsequence. & 
\begin{tikzpicture}[scale=0.8, every node/.style={transform shape}, baseline=(current bounding box.center)]
\small
\node[state, initial] (q0) {$0/1$};
\node[state, right=0.5cm of q0] (q1) {$1/1$};
\node[state, below=1.0cm of q1] (q2) {$2/1$};
\node[state, right=0.5cm of q1] (q3) {$3/0$};
\node[state, right=0.5cm of q3] (q4) {$4/0$};
\path[->]
(q0) edge[loop above] node[auto] {0} (q0)
(q0) edge[bend left] node[auto] {1} (q1)
(q1) edge[bend left] node[auto] {1} (q0)
(q1) edge[] node[auto] {0} (q3)
(q2) edge[] node[left] {1} (q1)
(q2) edge[bend left] node[right] {0} (q3)
(q3) edge[bend left] node[left] {0} (q2)
(q3) edge[] node[auto] {1} (q4)
(q4) edge[loop above] node[auto] {0,1} (q4);

\end{tikzpicture}\\

Tomita~4 & Recognize sequences where 000 does not occur & \begin{tikzpicture}[scale=0.8, every node/.style={transform shape}, baseline=(current bounding box.center)]
\small
\node[state, initial] (q0) {$0/1$};
\node[state, right=0.5cm of q0] (q1) {$1/1$};
\node[state, right=0.5cm of q1] (q2) {$2/1$};
\node[state, right=0.5cm of q2] (q3) {$3/0$};
\path[->]
(q0) edge[loop above] node[auto] {1} (q0)
(q0) edge[bend left] node[auto] {0} (q1)
(q1) edge[bend left] node[auto] {1} (q0)
(q1) edge[] node[auto] {0} (q2)
(q2) edge[bend left=60] node[auto=right, yshift=-0.1cm] {1} (q0)
(q2) edge[] node[auto] {0} (q3)
(q3) edge[loop above] node[auto] {0,1} (q3);

\end{tikzpicture}\\

Tomita~5 & Recognize sequences where there are an even number of 0s and 1s. & \begin{tikzpicture}[scale=0.8, every node/.style={transform shape}, baseline=(current bounding box.center)]
\small
\node[state, initial] (q0) {$0/1$};
\node[state, right=1.0cm of q0] (q1) {$1/0$};
\node[state, below=1.0cm of q1] (q2) {$2/0$};
\node[state, below=1.0cm of q0] (q3) {$3/0$};
\path[->]
(q0) edge[bend left=10] node[auto] {1} (q1)
(q1) edge[bend left=10] node[auto] {1} (q0)

(q1) edge[bend left=10] node[auto] {0} (q2)
(q2) edge[bend left=10] node[left] {0} (q1)

(q2) edge[bend left=10] node[auto] {1} (q3)
(q3) edge[bend left=10] node[auto] {1} (q2)

(q3) edge[bend left=10] node[auto] {0} (q0)
(q0) edge[bend left=10] node[auto] {0} (q3);

\end{tikzpicture}\\

Tomita~6 & Recognize sequences where $\text{number of 1s} - \text{number of 0s} \mod 3 = 0$ & \begin{tikzpicture}[scale=0.8, every node/.style={transform shape}, baseline=(current bounding box.center)]
\small
\node[state, initial] (q0) {$0/1$};
\node[state, right=1.5cm of q0, yshift=1cm] (q1) {$1/0$};
\node[state, below=1.5cm of q1, xshift=-0.25cm] (q2) {$2/0$};
\path[->]
(q0) edge[bend left=10] node[auto] {1} (q1)
(q1) edge[bend left=10] node[auto] {0} (q0)

(q1) edge[bend left=10] node[auto] {1} (q2)
(q2) edge[bend left=10] node[left] {0} (q1)

(q2) edge[bend left=10] node[auto] {1} (q0)
(q0) edge[bend left=10] node[auto] {0} (q2);

\end{tikzpicture}\\

Tomita~7 & Recognize sequences that belong to the regular expression $0^*1^*0^*1^*$. & \begin{tikzpicture}[scale=0.8, every node/.style={transform shape}, baseline=(current bounding box.center)]
\small
\node[state, initial] (q0) {$0/1$};
\node[state, right=0.35cm of q0] (q1) {$1/1$};
\node[state, right=0.35cm of q1] (q2) {$2/1$};
\node[state, right=0.35cm of q2] (q3) {$3/1$};
\node[state, right=0.35cm of q3] (q4) {$4/0$};
\path[->]
(q0) edge[loop above] node[auto] {0} (q0)
(q1) edge[loop above] node[auto] {1} (q1)
(q2) edge[loop above] node[auto] {0} (q2)
(q3) edge[loop above] node[auto] {1} (q3)

(q0) edge[] node[auto] {1} (q1)
(q1) edge[] node[auto] {0} (q2)
(q2) edge[] node[auto] {1} (q3)
(q3) edge[] node[auto] {0} (q4)
(q4) edge[loop above] node[auto] {0,1} (q4);

\end{tikzpicture}\\

\bottomrule
\end{tabular}
\end{table}

}

\textbf{Sequence Sampling} Parity~Check sequences are sampled by generating binary sequences uniformly at random of length $n$ and determining the label using the sum of the sequence modulo 2. Likewise, Even~Pairs sequences are sampled in the same manner, but the label is determined by counting the number of $01$ and $10$ pairs in the sequence and checking if their sum is even. Cycle~Navigation sequences are sampled with a uniform distribution over the alphabet $\{-1, 0, 1\}$, and the label is determined by summing the sequence modulo the cycle length (5). Sampling Modular~Arithmetic sequences is different as we always sample valid arithmetic expressions where sequences at even index (so the first token, third token, etc.) are operands ($\{0, 1, 2, 3, 4\}$) and sequences at odd index are operators ($\{+, -, \times\}$). Determining the label is not as simple as evaluating the expression, as we cannot obey the usual orders of operations (a restriction of regular languages), so we evaluate the expression from left to right, ignoring operator precedence. The label is the result of the expression modulo 5.

We sample positive $D_n$ sequences using a modification to the algorithm presented in \citet{arnold1980uniform} that samples balanced parenthesis strings. We constrain the algorithm to obey the fixed depth $n$ and to close brackets at the end of the sequence to return to the initial state, ensuring that it is always a positive sequence. Within the training lengths, we further augment the algorithm with randomness to increase the diversity of the sequences. This randomness is introduced by modifying the probability of closing brackets at each step (when it is possible but not necessary to close a bracket) with noise sampled from $\mathcal{N}(0, 0.15)$ and a depth bias that decreases the probability of closing brackets as the depth increases. This ensures that the sequences are still positive while introducing variability. The depth bias is $0.1 \times \frac{\text{current\_depth}}{\text{max\_depth}}$. These augmentations only occur within the training lengths, so the test sequences are sampled using the constrained algorithm without randomness. Negative sequences are sampled by generating binary sequences uniformly at random and checking if the sequence is accepted. Therefore, the batches are not guaranteed to be balanced, but it is unlikely to be significantly unbalanced.

For the Tomita languages, we explicitly construct their DFAs (Table~\ref{tab:bhattamishra_tasks}) and simulate their behavior to sample sequences. Again, we sample positive and negative sequences separately, but we oversample sequences when we cannot guarantee the acceptance (or rejection) of the sampled sequences. For Tomita~3, we constrained positive sequence sampling by disallowing the transition from state 3 to state 4, but this does not guarantee acceptance, so we oversampled by a factor of 2. When sampling negative sequences, we sampled symbols uniformly at random and checked if the sequence was accepted by the DFA (oversampling factor 2.5). For Tomita~4, we can guarantee the sampling of positive sequences by disallowing the transition from state 2 to state 3. However, we cannot guarantee the rejection of sequences, so we oversample negative sequences by a factor of 3. For Tomita~5, we oversample positive sequences by a factor of 5 and oversample negative sequences by a factor of 2. For Tomita~6, we oversample positive sequences by a factor of 4 and negative sequences by a factor of 2. For Tomita~7, we are able to guarantee the acceptance of positive sequences, but uniform sampling of the transitions would have caused low diversity in the sequences, so we bias the probabilities of taking transitions. We set the probability of a self-transition to $1 - \frac{4}{\max (\text{length}, 16)}$, where $\text{length}$ is the sequence length. The probability to transitioning to the next state to $\frac{4}{\max (\text{length}, 16)}$. This ensures that the sequences are still positive while increasing diversity in the sequences. For negative sequences, we oversample by a factor of 5.

Sequences for $P_{p,q}$ tasks are sampled by generating a sequence of length $n$ with a uniform distribution over the alphabet $\{0,\dots,q - 1\}$. The label is determined by mapping the first $p$ symbols to the corresponding output class.

\textbf{Hyperparameters} We provide general training hyperparameters in Table~\ref{tab:training_hyperparameters} and task-specific training hyperparameters in Table~\ref{tab:task_specific_hyperparameters}. Model hyperparameters are detailed in Table~\ref{tab:model_hyperparameters}. The model hyperparameters for the RNN, LSTM, and transformer are the same as in \citet{deletang2023neural} to take advantage of their extensive experimentation. However, their training hyperparameters differ from ours as we apply linear-warmup on the learning rate, $L_2$ regularization, and the AMSGrad optimizer instead of Adam. All baseline models were retrained under our optimization setup; we have not reused results from prior work to account for this difference. For RegularGPT hyperparameters, we ran a hyperparameter sweep using the Modular~Arithmetic task. We used the range of hyperparameters given in \citet{Chi2023}. The hyperparameter grid is shown in Table~\ref{tab:regulargpt_sweep}.

Tasks that did not converge by 100k steps for any architecture were trained for 1M steps for all architectures. Tasks that were trained for 1M steps were: Modular~Arithmetic, $D_4$, $D_6$, $D_8$, and $D_{12}$. This allocation reflects the greater complexity of processing the underlying languages compared to simpler languages like the Tomita tasks and prefix languages.

For MLP-LDRU, we maintained consistent hyperparameters within task families (e.g., one family is the $D_n$ tasks). We used a lower dropout (0.1) for the \citet{deletang2023neural} family compared to the other tasks (0.25) because we hypothesize that the Modular~Arithmetic task requires more capacity in $\odot_\theta$ than the other tasks. 

Learning rates were assigned based on preliminary experiments and we typically applied a base learning rate of $1\times10^{-3}$ for tasks on 100k steps and $1\times10^{-4}$ for tasks on 1M steps (the only exceptions were: Modular~Arithmetic with a learning rate of $1\times10^{-3}$ and $D_2$ and $D_3$ with a learning rate of $1\times10^{-4}$ to maintain learning rate consistency within task families). This was the case for MLP-LDRU, RNN, LSTM, Gated DeltaNet, transformer, and S5 architectures. However, we applied a learning rate of $1\times10^{-4}$ for all tasks for RegularGPT and SD-SSM because it improved rate of convergence. We used negative eigenvalues for Gated DeltaNet as this empirically and theoretically improves the generalization performance~\citep{grazzi2025unlocking}. During these experiments, we noticed that Gated DeltaNet often converged to strong generalization performance but then diverges after training for further steps. We did not apply early stopping to maintain consistency across architectures, but we note that early stopping would likely have improved the results for Gated DeltaNet, including Parity~Check where our results report almost random accuracy.

\begin{table}[!htbp]
\centering
\caption{RegularGPT hyperparameter sweep on Modular~Arithmetic task. Grid search over optimizer, learning rate, and dropout probability with fixed architectural parameters.}
\label{tab:regulargpt_sweep}
\begin{tabular}{ll}
\toprule
\textbf{Parameter} & \textbf{Values} \\
\midrule
\textbf{Fixed Parameters} & \\
Embedding dimension & 256 \\
Number of heads & 8 \\
Chunk size & 2 \\
Shared weights & True \\
Thickness & 1 \\
\midrule
\textbf{Varied Parameters} & \\
Optimizer & Adam, AMSGrad \\
Learning rate & $1 \times 10^{-4}$, $3 \times 10^{-4}$, $5 \times 10^{-4}$ \\
Dropout probability & 0.0, 0.1 \\
\midrule
\textbf{Total configurations} & \textbf{12} \\
\bottomrule
\end{tabular}
\end{table}

\begin{table}[!htbp]
\centering
\caption{RegularGPT hyperparameter sweep results on Modular~Arithmetic task. OOD accuracy for a single seed per configuration. Models trained for 250k steps on sequences up to length 40, evaluated on lengths 41--500. Note: main results use 1M steps. Best configuration highlighted in bold.}
\label{tab:regulargpt_sweep_results}
\begin{tabular}{llcc}
\toprule
\textbf{Optimizer} & \textbf{Learning Rate} & \textbf{Dropout = 0.0} & \textbf{Dropout = 0.1} \\
\midrule
Adam & $1 \times 10^{-4}$ & 69.8 & 73.2 \\
Adam & $3 \times 10^{-4}$ & 72.2 & 49.2 \\
Adam & $5 \times 10^{-4}$ & 49.0 & 61.1 \\
\midrule
AMSGrad & $1 \times 10^{-4}$ & \textbf{88.7} & 73.2 \\
AMSGrad & $3 \times 10^{-4}$ & 68.4 & 78.1 \\
AMSGrad & $5 \times 10^{-4}$ & 80.8 & 73.4 \\
\bottomrule
\end{tabular}
\end{table}

\begin{table}[!htbp]
\setlength{\tabcolsep}{2pt}
\centering
\caption{General training hyperparameters shared across all experiments.}
\label{tab:training_hyperparameters}
\begin{tabular}{ll}
\toprule
\textbf{Parameter} & \textbf{Value} \\
\midrule
Optimizer & AMSGrad \\
Base learning rate & $1\times 10^{-3}$ / $1\times 10^{-4}$ ($D_n$ tasks; RegularGPT and SD-SSM) \\
Init learning rate & $1\times 10^{-8}$ \\
Learning rate schedule & Linear warmup (20\% of steps)\\
Weight decay & 0.0 \\
$L_2$ regularization & $5\times 10^{-4}$ \\
Gradient clipping & 1.0 (global norm) \\
Centralized gradients & Yes \\
Batch size & 256 \\
Sequence length sampling & Uniform between 1 and max length \\
Max training length & 40 (standard) / 60, 100, 150 (length experiments) \\
Early stopping & None \\
Class balancing & Equal positive/negative examples per batch \\
Precision & Float32 \\
Seeds & 0, 1, 2 \\
\bottomrule
\end{tabular}
\end{table}

\begin{table}[!htbp]
\centering
\caption{Task-specific training hyperparameters showing training steps (all architectures) and dropout (MLP-LDRU only).}
\label{tab:task_specific_hyperparameters}
\begin{tabular}{lrr}
\toprule
\textbf{Task} & \textbf{Training Steps} & \textbf{Dropout} \\
\midrule
\textit{1) \citet{deletang2023neural}} & & \\
Even~Pairs & 100,000 & 0.1 \\
Modular~Arithmetic & 1,000,000 & 0.1 \\
Parity~Check & 100,000 & 0.1 \\
Cycle~Navigation & 100,000 & 0.1 \\
\cmidrule(lr){1-3}
\textit{2) \citet{Bhattamishra2020OnTA}} & & \\
$D_2$ & 100,000 & 0.25 \\
$D_3$ & 100,000 & 0.25 \\
$D_4$ & 1,000,000 & 0.25 \\
$D_6$ & 1,000,000 & 0.25 \\
$D_8$ & 1,000,000 & 0.25 \\
$D_{12}$ & 1,000,000 & 0.25 \\
Tomita~3 & 100,000 & 0.25 \\
Tomita~4 & 100,000 & 0.25 \\
Tomita~5 & 100,000 & 0.25 \\
Tomita~6 & 100,000 & 0.25 \\
Tomita~7 & 100,000 & 0.25 \\
\cmidrule(lr){1-3}
\textit{3) Prefix languages (Ours)} & & \\
$P_{1,2}$ & 100,000 & 0.25 \\
$P_{2,2}$ & 100,000 & 0.25 \\
$P_{4,2}$ & 100,000 & 0.25 \\
$P_{1,4}$ & 100,000 & 0.25 \\
$P_{2,4}$ & 100,000 & 0.25 \\
$P_{4,4}$ & 100,000 & 0.25 \\
\bottomrule
\end{tabular}
\end{table}

\textbf{Model Architecture Hyperparameters} We present the model hyperparameters for all baseline models and MLP-LDRU in Table~\ref{tab:model_hyperparameters}.

\begin{table}[!htbp]
\centering
\caption{Model architecture hyperparameters for all baseline models and MLP-LDRU. Note that while dropout can be applied to RegularGPT, the hyperparameter sweep indicated that zero dropout is better for Modular~Arithmetic, so we did not use it. Total parameters rounded to 2 significant figures.}
\label{tab:model_hyperparameters}
\resizebox{\textwidth}{!}{
\begin{tabular}{@{}llccccccccc@{}}
\toprule
\textbf{Component} &
  \textbf{Parameter} &
  \textbf{RNN} &
  \textbf{LSTM} &
  \textbf{Gated DeltaNet} &
  \textbf{Transformer} &
  \textbf{S5} &
  \textbf{SD-SSM} &
  \textbf{RegularGPT} &
  \textbf{BBT-GRC} &
  \textbf{MLP-LDRU} \\ \midrule
Embedding &
  Embedding dim &
  None &
  None &
  None &
  64 &
  None &
  None &
  256 &
  64 &
  64 \\
 &
  Initialization &
  -- &
  -- &
  -- &
  $\mathcal{N}(0, 0.02)$ &
  -- &
  -- &
  $\mathcal{N}(0, 0.02)$ &
  $\mathcal{N}(0, 1)$ &
  $\mathcal{N}(0, 0.02)$ \\ \midrule
Core Architecture &
  Layers/Blocks &
  1 &
  1 &
  3 &
  5 &
  2 &
  1 &
  1 &
  1 &
  1 \\
 &
  Hidden dim &
  256 &
  256 &
  64 &
  64 &
  256 &
  256 &
  256 &
  96 &
  64 \\ \midrule
Residual Connections &
  Dimension &
  -- &
  -- &
  128 &
  256 &
  -- &
  -- &
  1024 &
  -- &
  256 \\ \midrule
Normalization &
  Layer norm &
  No &
  No &
  Pre-norm &
  Pre-norm \& Post-norm &
  Post-norm &
  Post-norm &
  Pre-norm \& Post-norm &
  Post-norm &
  Post-norm \\ \midrule
MLP-LDRU &
  MLP hidden dims &
  -- &
  -- &
  -- &
  -- &
  -- &
  -- &
  -- &
  128 → 64 → 256 &
  128 → 256 → 128 \\
 &
  Activation &
  -- &
  -- &
  -- &
  -- &
  -- &
  -- &
  -- &
  GELU \& Sigmoid &
  ReLU \\
 &
  MLP initialization &
  -- &
  -- &
  -- &
  -- &
  -- &
  -- &
  -- &
  -- &
  Glorot~\citep{glorot2010understanding} \\
 &
  Projection initialization &
  -- &
  -- &
  -- &
  -- &
  -- &
  -- &
  -- &
  -- &
  Identity \\ \midrule
Gated DeltaNet &
  Negative eigs. &
  -- &
  -- &
  Yes &
  -- &
  -- &
  -- &
  -- &
  -- &
  -- \\ \midrule
S5 &
  Size base &
  -- &
  -- &
  -- &
  -- &
  256 &
  -- &
  -- &
  -- &
  -- \\
 &
  Num blocks &
  -- &
  -- &
  -- &
  -- &
  8 &
  -- &
  -- &
  -- &
  -- \\
 &
  Activation fn &
  -- &
  -- &
  -- &
  -- &
  GELU &
  -- &
  -- &
  -- &
  -- \\
 &
  Mode &
  -- &
  -- &
  -- &
  -- &
  Mean pooling &
  -- &
  -- &
  -- &
  -- \\
 &
  Clipped eigs. &
  -- &
  -- &
  -- &
  -- &
  Yes &
  -- &
  -- &
  -- &
  -- \\
 &
  Discretization &
  -- &
  -- &
  -- &
  -- &
  ZOH &
  -- &
  -- &
  -- &
  -- \\ \midrule
SD-SSM &
  Transition matrices &
  -- &
  -- &
  -- &
  -- &
  -- &
  8 &
  -- &
  -- &
  -- \\
 &
  $L_p$ norm &
  -- &
  -- &
  -- &
  -- &
  -- &
  1.2 &
  -- &
  -- &
  -- \\ \midrule
Transformer &
  Attention heads &
  -- &
  -- &
  4 &
  8 &
  -- &
  -- &
  8 &
  -- &
  -- \\
 &
  Head dimension &
  -- &
  -- &
  12 &
  8 &
  -- &
  -- &
  32 &
  -- &
  -- \\
 &
  Positional encoding &
  -- &
  -- &
  None &
  Varies &
  -- &
  -- &
  None &
  -- &
  -- \\ \midrule
RegularGPT &
  Chunk size &
  -- &
  -- &
  -- &
  -- &
  -- &
  -- &
  2 &
  -- &
  -- \\
 &
  Shared weights &
  -- &
  -- &
  -- &
  -- &
  -- &
  -- &
  Yes &
  -- &
  -- \\
 &
  Dilated attention &
  -- &
  -- &
  -- &
  -- &
  -- &
  -- &
  Yes &
  -- &
  -- \\ \midrule
\textbf{Total Parameters} &
  \textbf{Approx.} &
  \textbf{67k} &
  \textbf{270k} &
  \textbf{230k} &
  \textbf{250k} &
  \textbf{270k} &
  \textbf{530k} &
  \textbf{3200k} &
  \textbf{240k} &
  \textbf{160k} \\ \bottomrule
\end{tabular}
}
\end{table}

\textbf{Different Operator Choices} Here we describe the different reduction operators we evaluate. We have described the MLP-LDRU operator in the main text and Appendix~\ref{app:ldru_operator_details}. Let $\mathbf{h}_i, \mathbf{h}_j$ be the embeddings of the two input tokens. We also evaluate the following operators:
\begin{description}
    \item[Element-wise Sum (Elem. Sum)] The element-wise sum of the two embeddings, $\mathbf{h}_i + \mathbf{h}_j$.
    \item[Linear] A linear projection of the concatenated embeddings, i.e, $\mathbf{W}([\mathbf{h}_i ; \mathbf{h}_j]) + \mathbf{b}$, where $\mathbf{W}$ is a learned weight matrix and $\mathbf{b}$ is a learned bias vector.
    \item[Gated Sum] A gated sum of the two embeddings. We determine the gates using a linear projection of the concatenated embeddings, i.e., $\mathbf{g} = \sigma(\mathbf{W}_g([\mathbf{h}_i ; \mathbf{h}_j]) + \mathbf{b}_g)$, where $\sigma$ is the sigmoid activation function. The output is then $\mathbf{g} \circ \mathbf{h}_i + (1 - \mathbf{g}) \circ \mathbf{h}_j$.
    \item[Gated Recursive Cell~\cite{NEURIPS2019_d8e1344e}] Given in Eq. 12 and 13 of \citet{NEURIPS2019_d8e1344e}, this operator uses gating mechanisms similar to MLP-LDRU to produce an embedding from two adjacent input embeddings. The most significant difference between GRC and MLP-LDRU is that there is non-associativity bias due to the term $\sigma(c_t^i) \odot u_t^i$ in Eq. 13, where $c_t^i$ and $u_t^i$ are both computed by applying a feedforward network on $[\mathbf{h}_i ; \mathbf{h}_j]$.
\end{description}

\subsection{Optimizer Ablation}
\label{app:optimizer_ablation}

\textbf{Experimental Setup} We use the Modular~Arithmetic task to compare the OOD performance of MLP-LDRU when training with the Adam and AMSGrad optimizers with and without dropout. The experiments were conducted with fixed hyperparameters (except for the optimizer algorithm and dropout rate) found in Table~\ref{tab:training_hyperparameters} and Table~\ref{tab:model_hyperparameters}. We train the models for 1M steps on sequences up to length 40 and evaluate them on sequences of lengths 41--500. The results are averaged over 10 seeds.

\begin{figure}[!thbp]
    \centering
    \includegraphics[width=5.5in]{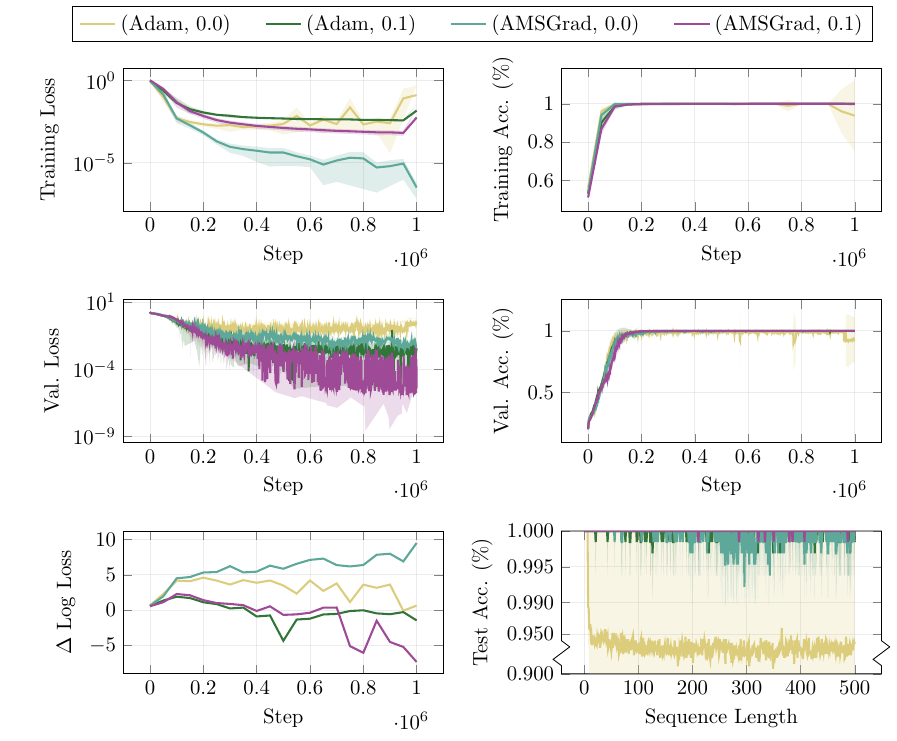}
    \caption{Optimizer ablation study comparing Adam and AMSGrad performance on MLP-LDRU models. Standard deviations are shown as the filled area around the mean values. The experiments show that dropout is necessary to enable generalization outside of the training distribution. It is notable that the AMSGrad optimizer without dropout achieves the lowest cross-entropy training loss but has worse performance outside the training distribution. Note the broken axis on the y-axis for the test accuracy (Test Acc.) plot to better highlight the performance differences between the highly performing models.}
    \label{fig:optimizer_ablation}
\end{figure}

\textbf{Results} We present the results of this ablation in Table~\ref{tab:optimizer_dropout_ablation} and Fig.~\ref{fig:optimizer_ablation}. With both optimizers, dropout improves generalization, as the models trained without dropout (both Adam and AMSGrad) perform worse on the OOD test set. Adam without dropout demonstrated catastrophic forgetting on seed 9, indicating that it is more unstable compared to AMSGrad. AMSGrad with dropout achieves the best performance, indicating robustness when handling the Modular~Arithmetic task.

\begin{table}[!htbp]
\centering
\caption{OOD accuracies when training MLP-LDRU with different optimizers and dropout combinations on the Modular~Arithmetic task. Individual seed results with mean and standard deviation across 10 seeds. The results demonstrate that dropout is necessary for generalization, with AMSGrad showing superior performance when combined with dropout. We note that the AMSGrad optimizer without dropout achieves competitive OOD accuracy compared to the optimizers with dropout.}
\label{tab:optimizer_dropout_ablation}
\begin{tabular}{lcccc}
\toprule
\textbf{Seed} & \textbf{(Adam, 0.0)} & \textbf{(Adam, 0.1)} & \textbf{(AMSGrad, 0.0)} & \textbf{(AMSGrad, 0.1)} \\
\midrule
0 & 99.606 & 99.963 & 99.963 & 100.000 \\
1 & 99.691 & 99.980 & 99.949 & 99.997 \\
2 & 99.888 & 99.986 & 99.888 & 100.000 \\
3 & 99.684 & 99.959 & 99.946 & 100.000 \\
4 & 99.113 & 99.990 & 99.963 & 99.997 \\
5 & 95.143 & 99.993 & 99.898 & 99.997 \\
6 & 99.966 & 99.997 & 99.868 & 99.986 \\
7 & 98.689 & 99.997 & 99.963 & 100.000 \\
8 & 99.925 & 99.868 & 99.983 & 99.993 \\
9 & 40.231 & 99.976 & 99.864 & 100.000 \\
\midrule
\textbf{Mean $\pm$ Std} & \textbf{93.194 $\pm$ 17.707} & \textbf{99.971 $\pm$ 0.037} & \textbf{99.928 $\pm$ 0.042} & \textbf{99.997 $\pm$ 0.004} \\
\bottomrule
\end{tabular}
\end{table}

While the experiment was training, we evaluated a validation batch of 1024 sequences of length 500 every 1000 steps: the performance on this batch is how we determined the validation loss (Val. Loss) and accuracy (Val. Acc.) presented in Fig.~\ref{fig:optimizer_ablation}. This evaluation gave us an idea of how the model would perform on the most OOD sequences during training, and we found it to be a reliable indicator of the model's generalization across all the sequences up to length 500. The test accuracy (Test Acc.) is the performance on the test set of sampled 512 sequences per length, from 1 to 500. The $\Delta \text{Log Loss}$ is the difference between the log of the validation loss and the log of the training loss, which indicates how well the model generalizes to OOD sequences and overfits to the training sequences. We also note that the validation loss is evaluated on the model parameters without dropout, while the training loss includes dropout. The $\Delta \text{Log Loss}$ reveals an interesting difference between AMSGrad with and without dropout: with dropout indicates improving performance on the validation set during training, while without dropout indicates that the model overfits to the training sequences without improving on the validation set.

\section{All Regular Task Results}
\label{app:full_results}

\textbf{All Baselines} We present the OOD accuracies of all baseline models on all tasks in Table~\ref{tab:full_results} under the experimental settings for \textbf{RQ1} (main text). The difference between Table~\ref{tab:full_results} and Table~\ref{tab:central_result} is that the former includes all baselines, while the latter does not include S5 and SD-SSM. We use the same statistical significance notation as in Table~\ref{tab:central_result}.

\input{tables/appendix/full_results_new.tex}

In addition, we present the ID accuracies of all models in Table~\ref{tab:full_results_id}.

\input{tables/appendix/full_results_ID.tex}

\textbf{Individual Seed Results for Main Tasks} We present the OOD accuracies for individual seeds for all 21 regular language tasks in Table~\ref{tab:individual_seeds_main}. The models were trained on sequences up to length 40 and evaluated on sequences of lengths 41--500.

\input{tables/appendix/individual_results_main_new.tex}

\textbf{Extended Sequence Length Results} We present the individual seed results for the sequence length experiments on $D_n$ languages in Table~\ref{tab:individual_seeds_length}. The models were trained on sequences up to length $x \in [40, 60, 100, 150]$ and evaluated on sequences of length from $x+1$ to 500.

\begin{table*}[!htbp]
\centering
\caption{Individual seed results for sequence length experiments on $D_n$ languages.}
\label{tab:individual_seeds_length}
\begin{tabular}{llcccccc}
\toprule
& & \multicolumn{3}{c}{RNN} & \multicolumn{3}{c}{MLP-LDRU} \\
\cmidrule(lr){3-5} \cmidrule(lr){6-8}
Task & Max Length & 0 & 1 & 2 & 0 & 1 & 2 \\
\midrule
$D_4$ & 40 & 85.0 & 100.0 & 100.0 & 100.0 & 100.0 & 100.0 \\
& 60 & 100.0 & 100.0 & 100.0 & 100.0 & 100.0 & 100.0 \\
& 100 & 100.0 & 100.0 & 100.0 & 100.0 & 100.0 & 100.0 \\
& 150 & 100.0 & 100.0 & 100.0 & 100.0 & 100.0 & 100.0 \\
\midrule
$D_6$ & 40 & 91.6 & 99.6 & 100.0 & 99.9 & 95.7 & 99.9 \\
& 60 & 99.9 & 98.8 & 99.9 & 100.0 & 100.0 & 100.0 \\
& 100 & 100.0 & 100.0 & 100.0 & 100.0 & 100.0 & 100.0 \\
& 150 & 100.0 & 100.0 & 100.0 & 100.0 & 100.0 & 100.0 \\
\midrule
$D_8$ & 40 & 100.0 & 97.7 & 99.9 & 99.8 & 98.2 & 96.6 \\
& 60 & 99.9 & 100.0 & 100.0 & 98.8 & 91.5 & 99.9 \\
& 100 & 100.0 & 100.0 & 100.0 & 100.0 & 100.0 & 100.0 \\
& 150 & 100.0 & 100.0 & 100.0 & 100.0 & 100.0 & 100.0 \\
\midrule
$D_{12}$ & 40 & 96.9 & 81.8 & 99.3 & 97.5 & 96.8 & 93.6 \\
& 60 & 81.8 & 82.1 & 98.1 & 96.1 & 98.2 & 96.5 \\
& 100 & 91.3 & 77.8 & 92.6 & 99.9 & 99.9 & 99.9 \\
& 150 & 99.4 & 99.5 & 100.0 & 99.9 & 99.9 & 99.9 \\
\bottomrule
\end{tabular}
\end{table*}

\textbf{Operator Choice Results} We present the individual seed results for the \textbf{RQ3} experiments in Table~\ref{tab:individual_seeds_operator}. The models were trained on sequences up to length 40 and evaluated on sequences of lengths 41--500. The results show that the MLP operator achieves 100.0\% OOD accuracy on all tasks. The other operators show varying performance---but none achieve $100.0\%$ OOD accuracy on Modular~Arithmetic.

\begin{table}[!htbp]
\centering
\caption{Individual seed results for operator study on the regular \citet{deletang2023neural} tasks. All models were trained with a max sequence length of 40.}
\label{tab:individual_seeds_operator}
\resizebox{\textwidth}{!}{\begin{tabular}{lccccccccccccccc}
\toprule
& \multicolumn{3}{c}{MLP} & \multicolumn{3}{c}{GRC} & \multicolumn{3}{c}{Elem. Sum} & \multicolumn{3}{c}{Concat. Proj.} & \multicolumn{3}{c}{Gated Sum} \\
\cmidrule(lr){2-4} \cmidrule(lr){5-7} \cmidrule(lr){8-10} \cmidrule(lr){11-13} \cmidrule(lr){14-16}
Task & 0 & 1 & 2 & 0 & 1 & 2 & 0 & 1 & 2 & 0 & 1 & 2 & 0 & 1 & 2 \\
\midrule
Even~Pairs & 100.0 & 100.0 & 100.0 & 99.9 & 100.0 & 100.0 & 51.9 & 51.8 & 51.9 & 100.0 & 100.0 & 100.0 & 100.0 & 100.0 & 100.0 \\
Modular~Arith. & 100.0 & 100.0 & 100.0 & 99.6 & 76.5 & 70.7 & 32.4 & 32.6 & 32.9 & 59.4 & 60.3 & 62.7 & 68.7 & 68.8 & 63.7 \\
Parity~Check & 100.0 & 100.0 & 100.0 & 100.0 & 100.0 & 100.0 & 100.0 & 100.0 & 100.0 & 100.0 & 100.0 & 100.0 & 100.0 & 100.0 & 100.0 \\
Cycle Nav. & 100.0 & 100.0 & 100.0 & 100.0 & 100.0 & 100.0 & 100.0 & 100.0 & 100.0 & 100.0 & 100.0 & 100.0 & 100.0 & 100.0 & 100.0 \\
\bottomrule
\end{tabular}}
\end{table}

\textbf{MLP-LDRU Size Ablation} To address whether MLP-LDRU's regular-task performance depends on a tuned hidden dimension, we vary the embedding dimension on the four \citet{deletang2023neural} tasks used in the operator study. Table~\ref{tab:ldru_size_ablation} shows that dimensions 64 and 128 solve all four tasks, while dimension 32 nearly solves Modular~Arithmetic and dimension 16 fails only on Modular~Arithmetic.

\begin{table}{!htbp}
\centering
\caption{MLP-LDRU size ablation results on the regular \citet{deletang2023neural} tasks. All models were trained with a max sequence length of 40 and evaluated on sequences of lengths 41--500. The results show that MLP-LDRU can learn to solve all four tasks with embedding dimensions as low as 64, while dimensions of 32 and 16 show some performance degradation on Modular~Arithmetic.}
\label{tab:ldru_size_ablation}
\begin{tabular}{lcccc}
\toprule
Embedding Dim. & Even~Pairs & Modular Arithmetic & Parity Check & Cycle Navigation \\ \midrule
16 & $100.0\pm0.0$ & $50.2\pm5.7$ & $100.0\pm0.0$ & $100.0\pm0.0$ \\
32 & $100.0\pm0.0$ & $96.9\pm4.7$ & $100.0\pm0.0$ & $100.0\pm0.0$ \\
64 & $100.0\pm0.0$ & $100.0\pm0.0$ & $100.0\pm0.0$ & $100.0\pm0.0$ \\
128 & $100.0\pm0.0$ & $100.0\pm0.0$ & $100.0\pm0.0$ & $100.0\pm0.0$ \\ \bottomrule
\end{tabular}
\end{table}

\section{Associativity Regularization}
\label{app:associativity_regularization}

Our results on the regular tasks indicate that MLP-LDRU can learn to approximate an associative operator without explicit regularization. However, we hypothesize that adding an associativity regularization term to the training loss could further improve the model's ability to learn an associative operator, especially in low-data regimes or on non-regular tasks. Associativity encourages the operator to behave as a recurrence, which we believe is beneficial for generalization. To test this hypothesis, we added an associativity regularization term to the training loss during experiments on the ListOps experiments and the natural language tasks described in Appendix~\ref{app:non_regular_experiments}.

We compute this additional loss term by taking every locally valid triple $(h_a, h_b, h_c)$ computed during the forward pass of an LDRU model's reduction and punish deviations from associativity. Let $\odot_\theta$ denote a general LDRU operator. For each step $k$ in the reduction, we partition the sequence of token embeddings into these triples and we evaluate the following expressions:
\begin{equation}
x = \odot_\theta(\odot_\theta(\mathbf{h}_a, \mathbf{h}_b), \mathbf{h}_c); y = \odot_\theta(\mathbf{h}_a, \odot_\theta(\mathbf{h}_b, \mathbf{h}_c)),
\end{equation}
which enables us to compute the associativity loss as:
\begin{equation}
\ell_{assoc}(x,y) = (1 - \frac{x \cdot y}{|x||y| + \epsilon})^2.
\end{equation}
Let $\tau_k$ be the set of valid triples evaluated at step $k$. The associativity loss for step $k$ is:
\begin{equation}
\mathcal{L}_{assoc}^{(k)} = \frac{1}{|\tau_k|} \sum_{(h_a, h_b, h_c) \in \tau_k} \ell_{assoc}(x, y),
\end{equation}and the total associativity loss across all reduction steps is:
\begin{equation}
\mathcal{L}_{assoc} = \frac{1}{K} \sum_{k=1}^{K} \mathcal{L}_{assoc}^{(k)}.
\end{equation}
This loss term is then added to the standard training loss with a weighting factor $\lambda_{assoc}$:
\begin{equation}
\mathcal{L} = \mathcal{L}_{task} + \lambda_{assoc} \mathcal{L}_{assoc}.
\end{equation}

\section{Non-Regular Language Experiments}
\label{app:non_regular_experiments}

We restrict our extended study to sequence classifications to maintain consistency with our regular language experiments. We provide additional details about the ListOps and natural language tasks below.

\subsection{ListOps}
\label{app:listops_details}

We generated a dataset of ListOps sequences following the procedure outlined in \citet{nangia2018listops}. Each sequence is a list beginning with an operation (MAX, MIN, MED, SUM (modulo 10)) followed by either integer or nested list arguments. The sequences are generated with varying lengths, maximum depths, and maximum numbers of arguments per operation. We created training datasets of sizes 100k, 500k, and 1M sequences, containing sequences with lengths ranging from 5 to 40. The test data comprises of multiple sets where each set contains 10k sequences sampled ith different characteristics of length (bucketed), max depth and max number of arguments. This enables a comprehensive evaluation of length, depth and argument generalization of the trained models.

We performed hyperparameter sweeps for each of the baseline models and MLP-LDRU on the ListOps task, except for RIR-GRC and BBT-GRC, where we used the default hyperparameters given in the codebase of \citet{chowdhury2023recursion}. For RIR-GRC, we additionally reduced the model chunk size from 30 to 12 to account for a lower maximum training sequence length compared to the original experiments. This change maintains the ratio of max training length to chunk size as $0.3$ to ensure that RIR-GRC can take advantage of the outer recursion. The hyperparameter sweeps included learning rate, dropout probability, and model hidden dimension, and in the case of MLP-LDRU: the weight of the associativity regularization. We used a single seed (1) for this sweep and we used the Adam~\citep{kingma2015adam} optimizer with a batch size of 128. We continued to use linear warmup for 20\% of the training steps initially set to 1e{-8} but no $L_2$ regularization was used. We trained each configuration for 200k steps on the 500k dataset and evaluated performance on a validation set of 2048 sequences sampled from the same distribution. We selected the best hyperparameters based on the highest average accuracy on the validation data. The final hyperparameters used for each model are presented in Table~\ref{tab:listops_hyperparameters}.

\begin{table}[!htbp]
\centering
\caption{Hyperparameter sweep ranges for LSTM, transformer, and MLP-LDRU for ListOps.}
\label{tab:hparam_sweep}
\resizebox{\textwidth}{!}{%
\begin{tabular}{lcccc}
\toprule
\textbf{Model} & \textbf{Embedding / Hidden Dim} & \textbf{Dropout} & \textbf{Learning Rate} & \textbf{Associativity Regularization}\\
\midrule
LSTM         & \{256, 512, \textbf{1024}\}   & -- & \{1e{-5}, 5e{-5}, \textbf{1e{-4}}\} & -- \\
Transformer  & \{64, \textbf{128}\}    & \{\textbf{0}, 0.1\} & \{1e{-5}, \textbf{5e{-5}}, 1e{-4}\} & -- \\
MLP-LDRU         & \{64, 128, \textbf{256}\}    & \{0, 0.1, 0.2\} & \{1e{-5}, 5e{-5}, 1e{-4}\} & \{0, 0.1, \textbf{1.0}\} \\
MLP-LDRU (additional) & \{\textbf{256}, 512\} & \{0.025, \textbf{0.05}, 0.1\} & \{1e{-4}, \textbf{2.5e{-4}}, 5e{-4}\} & \{\textbf{1.0}\} \\
\bottomrule
\end{tabular}}
\end{table}

\begin{table}[!htbp]
\centering
\caption{Selected model hyperparameters for all baseline models, BBT\textendash GRC, and MLP-LDRU for ListOps.}
\label{tab:listops_hyperparameters}
\resizebox{\textwidth}{!}{%
\small
\begin{tabular}{llccccc}
\toprule
\textbf{Component} & \textbf{Parameter} & \textbf{LSTM} & \textbf{Transformer} & \textbf{MLP-LDRU} & \textbf{BBT-GRC} & \textbf{RIR-GRC} \\
\midrule
Embedding & Embedding dim & None & 128 & 256 & 128 & 128\\
& Initialization & -- & $\mathcal{N}(0,0.02)$ & $\mathcal{N}(0,0.02)$ & -- & -- \\
\midrule
Core Architecture & Layers/Blocks & 1 & 5 & 1 & 1 & 1 \\
& Hidden dim & 1024 & 128 & 256 & 128 & 128 \\
\midrule
Residual Connections & Dimension & -- & 512 & 1024 & -- & -- \\
\midrule
Normalization & Layer norm & No & Pre-norm \& Post-norm & Post-norm & Post-norm & Post-norm \\
\midrule
MLP-LDRU & MLP hidden dims & -- & -- & 512 → 1024 → 512 & -- & -- \\
& Activation & -- & -- & SiLU & -- & -- \\
& MLP initialization & -- & -- & Glorot & -- & -- \\
& Projection initialization & -- & -- & Identity & -- & -- \\
\midrule
Transformer & Attention heads & -- & 8 & -- & -- & -- \\
& Head dimension & -- & 8 & -- & -- & -- \\
& Positional encoding & -- & ALiBi/NoPE/Sinusoidal & -- & -- & -- \\
\midrule
RIR-GRC & Chunk size & -- & -- & -- & -- & 12 \\
\midrule
Regularization & Dropout locations & -- & Attention \& Residual & Post-norm & In \& Out & In \& Out \\
& Associativity & -- & -- & Yes & -- & -- \\
\midrule
\textbf{Total Parameters} & \textbf{Approx.} & \textbf{4.3M} & \textbf{1.0M} & \textbf{2.6M} & \textbf{430k} & \textbf{430k} \\
\bottomrule
\end{tabular}}
\end{table}

\textbf{Results} We report an ablation of associativity strength on the performance of MLP-LDRU in Table~\ref{tab:listops_assoc_summary}. Aside for $\lambda_{assoc}$, we use the same settings as the initial hyperparameter sweep. These results indicate that further biases toward associativity is beneficial for performance on the ListOps task.

\input{tables/appendix/listops_assoc_summary.tex}

We present the full results for the ListOps ablation study across all length buckets, max depth, and max number of arguments in Table~\ref{tab:assoc_str_full_ablation}. The results show that MLP-LDRU with associativity regularization achieves the highest accuracy across most buckets and dataset sizes, particularly in the out-of-distribution buckets. The performance of MLP-LDRU improves with increased training data, and it consistently outperforms the standard baselines at 500k and 1M training samples. These findings suggest that encouraging approximate associativity through regularization is a useful inductive bias for improving generalization on the ListOps task.

\input{tables/appendix/listops_assoc.tex}

The tables presented in Table~\ref{tab:listops_all_results_app} report accuracy and standard deviation for length buckets with four combinations of max depth and max number of arguments under three training-set sizes. BBT-GRC largely attains the highest accuracy across buckets and all dataset sizes. Increasing the training-set size systematically improves MLP-LDRU accuracy: the 500k and 1M models show uniform gains relative to the 100k model across both in-distribution and out-of-distribution buckets. Under the same change in training-set size, transformer with ALiBi exhibits smaller improvements. MLP-LDRU surpasses the standard baselines (LSTM, transformer) at 500k and 1M. A consistent pattern across all sweeps was that MLP-LDRU configurations with non-zero associativity regularization achieved higher validation accuracy than those without it, indicating that encouraging approximate associativity is empirically beneficial and functions as a useful inductive bias rather than a redundant constraint.

\begin{table}[t]
\centering
\caption{ListOps results across different training dataset sizes. Each table presents accuracies (in \%) and standard deviations across 3 seeds for different length buckets under varying max depth and max number of arguments. MLP-LDRU benefits from increased training data, outperforming standard baselines at 500k and 1M training samples.}
\label{tab:listops_all_results_app}
\begin{minipage}[t]{0.32\textwidth}
\centering
\setlength{\tabcolsep}{6.6pt}
\caption*{100k dataset}
\label{tab:listops_100k_app}
\resizebox{\textwidth}{!}{\input{tables/appendix/listops_100k.tex}}
\end{minipage}\hfill
\begin{minipage}[t]{0.32\textwidth}
\centering
\caption*{500k dataset}
\label{tab:listops_500k_app}
\resizebox{\textwidth}{!}{\input{tables/appendix/listops_500k.tex}}
\end{minipage}\hfill
\begin{minipage}[t]{0.32\textwidth}
\centering
\caption*{1M dataset}
\label{tab:listops_1m_app}
\resizebox{\textwidth}{!}{\input{tables/appendix/listops_1m.tex}}
\end{minipage}

\end{table}

\subsection{Natural Language Tasks}
\label{app:nlp_details}

We evaluate MLP-LDRU and transformer baselines on a set of standard sequence-classification datasets. We report results on all GLUE classification tasks except RTE, AX, and WNLI, which we exclude due to their small size, high variance, and limited incremental diagnostic value relative to the larger, more stable benchmarks. We also report results on AG's~News and DBPedia, text classification tasks outside of GLUE. We report performance on the validation data for the GLUE tasks (using the recommended metrics for each individual task) and accuracy on test data for the additional datasets. We use the BERT base (uncased)~\citep{BERT-base-uncased} tokenizer for all tasks. For paired inputs \texttt{u,v}, the concatenation format was the conventional \texttt{[CLS],u,[SEP],v,[SEP]}. Task performance is reported in Table~\ref{tab:nlp_results}. We present the truncation lengths, learning rates, batch sizes, and training steps for each task in Table~\ref{tab:nlp_truncation_lengths}.

\begin{table*}[!bt]
\centering
\caption{Performance across GLUE benchmarks and two additional text classification datasets. Metrics follow GLUE conventions: CoLA (Matthew's Correlation Coefficient), SST-2 / MNLI / QNLI (\%, accuracy), QQP / MRPC (\%, F1/accuracy). Subscript M indicates matched and MM mismatched accuracy for MNLI. AG's~News and DBPedia report classification accuracy on test data. Values are means across 5 seeds.}
\label{tab:nlp_results}
\resizebox{\textwidth}{!}{\begin{tabular}{lccccccccc}
\toprule
 & \multicolumn{7}{c}{\textbf{GLUE Benchmarks}} & \multicolumn{2}{c}{\textbf{Additional Tasks}} \\
\cmidrule(lr){2-8} \cmidrule(lr){9-10}
Architecture & CoLA & SST-2 & MRPC & QQP & MNLI\textsubscript{M} & MNLI\textsubscript{MM} & QNLI & AG's~News & DBPedia \\
\midrule
TF (ALiBi)   & 0.096 & 79.3 & 56.2/62.6 & \underline{74.0/78.3} & \underline{53.9} & \underline{52.8} & \underline{58.1} & \textbf{89.8} & \underline{98.2} \\
TF (Sin.) & \textbf{0.124} & \textbf{81.1} & \textbf{59.1/67.5} & 70.7/75.8 & 49.0 & 50.5 & 56.9 & 89.1 & 97.8 \\
TF (NoPE)    & \underline{0.097} & 79.1 & \underline{59.0/64.0} & 72.8/77.8 & 50.5 & 51.1 & \textbf{58.8} & \underline{89.6} & 97.8 \\
MLP-LDRU         & 0.085 & \underline{80.9} & 57.0/63.2 & \textbf{76.9/79.1} & \textbf{58.3} & \textbf{57.9} & 56.1 & 89.0 & \textbf{98.6} \\
\bottomrule
\end{tabular}}
\end{table*}

\begin{table}[!htbp]
\centering
\caption{Sequence-length truncation thresholds, learning rates, batch sizes, and training steps for all evaluated NLP datasets.}
\label{tab:nlp_truncation_lengths}
\begin{tabular}{lcccc}
\toprule
\textbf{Dataset} & \textbf{Truncation Length} & \textbf{Learning Rate} & \textbf{Batch Size} & \textbf{Steps} \\
\midrule
CoLA & 64 & 1e{-5} & 128 & 4000 \\
SST-2 & 64 & 1e{-4} & 128 & 8000 \\
MRPC & 128 & 1e{-5} & 32 & 4000 \\
QQP & 256 & 1e{-4} & 32 & 50000 \\
MNLI & 384 & 1e{-4} & 32 & 35000 \\
QNLI & 384 & 1e{-4} & 32 & 20000 \\
AG's~News & 512 & 1e{-4} & 32 & 20000 \\
DBPedia & 256 & 1e{-4} & 32 & 30000 \\
\bottomrule
\end{tabular}
\end{table}

We performed a hyperparameter sweep to find the best configurations for transformer baselines (using ALiBi) and MLP-LDRU. We used the AG's~News task to conduct the sweep, varying learning rate, dropout probability, and embedding/hidden dimension. For MLP-LDRU we additionally swept the associativity-regularization weight. We used linear warmup over the first 20\% of updates (initial learning rate $10^{-8}$), and no $L_2$ regularization. We give the sweep parameters and embolden the selected parameters in Table~\ref{tab:nlp_hparams_sweep}. Additional model details for MLP-LDRU and transformer are listed in Table~\ref{tab:nlp_hparams_selected}.

\begin{table}[!htbp]
\centering
\caption{Hyperparameter sweep ranges for all models on natural language tasks. We used ALiBi as the PE for the hyperparameter sweep.}
\label{tab:nlp_hparams_sweep}
\resizebox{\textwidth}{!}{%
\begin{tabular}{lccccc}
\toprule
\textbf{Model} & \textbf{Embedding Dim} & \textbf{Dropout} & \textbf{Learning Rate} & \textbf{Weight Decay} &\textbf{Assoc. Reg.} \\
\midrule
Transformer & \{64, \textbf{128}\} & \{0, \textbf{0.1}\} & \{\textbf{1e{-5}}, 5e{-5}, 1e{-4}\} & \{0.0, 1e{-4}, \textbf{1e{-5}}\} & -- \\
Transformer (Additional) & \{256\} & \{\textbf{0.1}\} & \{\textbf{1e{-5}}\} & \{\textbf{1e{-5}}\} & -- \\
MLP-LDRU & \{64, 128, \textbf{256}\} & \{0, \textbf{0.01}, 0.025, 0.05\} & \{\textbf{1e{-5}}, 5e{-5}, 1e{-4}\} & \{0.0, 1e{-4}, \textbf{1e{-5}}\} & \{0, \textbf{0.1}, 1.0\} \\
MLP-LDRU (Additional) & \{\textbf{256}\} & \{\textbf{0.01}\} & \{\textbf{1e{-5}}, 5e{-5}\} & \{0.0, 1e{-4}, \textbf{1e{-5}}\} & \{0.5\} \\
\bottomrule
\end{tabular}}
\end{table}

\begin{table}[!htbp]
\centering
\caption{Model hyperparameters for all natural language tasks. We note that the hyperparameter values were selected based on a hyperparameter sweep on AG's~News: the selected values are emboldened in Table~\ref{tab:nlp_hparams_sweep}. The sweep included a transformer with approximately 12M parameters (embedding dimension 256), but this configuration was not selected as it did not outperform the smaller transformer presented here.}
\label{tab:nlp_hparams_selected}
\small
\begin{tabular}{llcc}
\toprule
\textbf{Component} & \textbf{Parameter} & \textbf{Transformer} & \textbf{MLP-LDRU} \\
\midrule
Embedding & Embedding dim & 128 & 256 \\
& Initialization & $\mathcal{N}(0,0.02)$ & $\mathcal{N}(0,0.02)$ \\
\midrule
Core & Layers/Blocks & 5 & 1 \\
& Hidden dim & 128 & 256 \\
\midrule
Attention & Heads & 8 & -- \\
& Head dim & 16 & -- \\
& Positional encoding & ALiBi/NoPE/Sinusoidal & -- \\
\midrule
MLP-LDRU & MLP hidden dims & -- & 512 → 1024 → 512 \\
& Activation & -- & SiLU \\
& Assoc. Reg. & -- & Yes \\
\midrule
Regularization & Dropout & 0.1 & 0.01 \\
\midrule
\textbf{Total Parameters} & \textbf{Approx.} & \textbf{4.9M} & \textbf{10M} \\
\bottomrule
\end{tabular}
\end{table}

\textbf{Results} The full results are shown in Table~\ref{tab:full_nlp_results}. Performance differences between MLP-LDRU and transformer baselines are largely moderate across all GLUE tasks, with each architecture exhibiting strengths on different subsets of benchmarks. On QQP and MNLI (matched and mismatched), MLP-LDRU achieves the highest average scores across seeds. On CoLA, SST-2, and MRPC, the transformer (Sinusoidal) obtains the strongest results. Transformer (NoPE) obtains the strongest results on QNLI. On the non-GLUE classification tasks, AG's~News and DBPedia, the MLP-LDRU obtains the lowest accuracy on AG's~News, but is similar to transformer performance in absolute terms. The MLP-LDRU also achieves the highest accuracy on DBPedia (98.6\%), exceeding all transformer variants. Variances across seeds are generally low for all models and tasks. Overall, these results indicate that MLP-LDRU is competitive with standard transformer architectures on natural language classification tasks, despite the fact that natural language is not a regular language.

Across the sweep on AG's~News, MLP-LDRU configurations with non-zero associativity regularization consistently outperformed those with zero weight, and the selected configuration for every dataset used $\lambda_{assoc} = 0.1$. This result indicates that encouraging approximate associativity is beneficial outside of regular-language settings as well.

\begin{table}[hbtp]
\centering
\caption{Performance across GLUE benchmarks, AG's~News, and DBPedia. Metrics follow GLUE conventions: CoLA (Matthew's Correlation Coefficient), SST-2 / MNLI / QNLI (Accuracy), MRPC / QQP (F1/Accuracy). AG's~News and DBPedia report classification accuracy on test data. We give the metrics and their standard deviations over 5 seeds. Subscript M indicates matched and MM mismatched accuracy for MNLI.}
\label{tab:full_nlp_results}
\resizebox{\textwidth}{!}{%
\begin{tabular}{lccccccccc}
\toprule
 & \multicolumn{7}{c}{\textbf{GLUE Benchmarks}} & \multicolumn{2}{c}{\textbf{Additional Tasks}} \\
\cmidrule(lr){2-8} \cmidrule(lr){9-10}
Architecture & CoLA & SST-2 & MRPC & QQP & MNLI\textsubscript{M} & MNLI\textsubscript{MM} & QNLI & AG's~News & DBPedia \\
\midrule
TF (ALiBi)   & $0.096 \pm 0.031$ & $79.3\pm1.0$ & $56.2\pm1.8$ / $62.6\pm2.9$ & \underline{$74.0\pm1.9$ / $78.3\pm0.9$} & \underline{$53.9\pm0.7$} & \underline{$52.8\pm0.7$} & \underline{$58.1\pm0.2$} & $\mathbf{89.8\pm0.4}$ & \underline{$98.2\pm 0.1$} \\
TF (Sin.) & $\mathbf{0.124\pm0.027}$ & $\mathbf{81.1\pm0.9}$ & $\mathbf{59.1\pm1.4}$ / $\mathbf{67.5\pm3.6}$ & $70.7\pm0.7$ / $75.8\pm0.2$ & $49.0\pm0.3$ & $50.5\pm0.5$ & $56.9\pm3.3$ & $89.1\pm0.9$ & $97.8\pm 0.1$ \\
TF (NoPE)    & \underline{$0.097\pm0.028$} & $79.1\pm1.2$ & \underline{$59.0\pm1.6$ / $64.0\pm2.4$} & $72.8\pm1.6$ / $77.8\pm0.8$ & $50.5\pm0.4$ & $51.1\pm0.6$ & $\mathbf{58.8\pm0.9}$ & \underline{$89.6\pm0.3$} & $97.8\pm 0.1$ \\
MLP-LDRU         & $0.085 \pm 0.036$ & \underline{$80.9\pm1.0$} & $57.0\pm3.6$ / $63.2\pm5.7$ & $\mathbf{76.9\pm0.3}$ / $\mathbf{79.1\pm0.3}$ & $\mathbf{58.3\pm0.7}$ & $\mathbf{57.9\pm0.7}$ & $56.1\pm1.2$ & $89.0\pm0.5$ & $\mathbf{98.6\pm 0.0}$ \\
\bottomrule
\end{tabular}}
\end{table}

\section{Interpretability Analysis}
\label{app:interpretability_analysis}

\begin{figure}[ht]
    \centering
    \begin{subfigure}[t]{0.48\textwidth}
        \centering
        \includegraphics[]{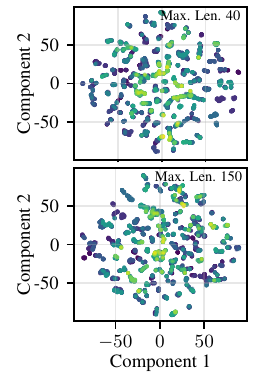}
        \caption{t-SNE plot of $D_6$ embeddings colored by EC, showing distinct clusters learned by MLP-LDRU.}
        \label{fig:tsne_embeddings}
    \end{subfigure}
    \begin{subfigure}[t]{0.48\textwidth}
        \centering
        \includegraphics{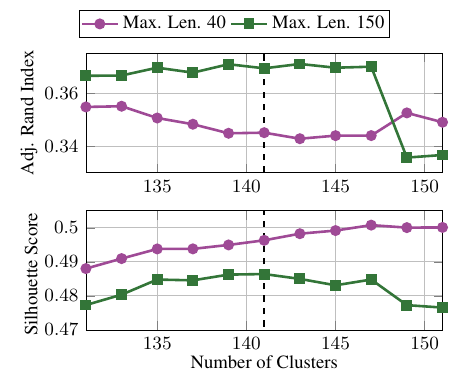}
        \caption{Clustering analysis of $D_6$ embeddings. The model with higher maximum training length displays improved alignment with the monoid structure, shown by higher ARI (top) and Silhouette Score peak (bottom).}
        \label{fig:clustering_plots}
    \end{subfigure}
    \caption{Interpretability analysis of MLP-LDRU embeddings for $D_6$.}
\end{figure}

We analyze the embedding space of $D_6$ to examine if MLP-LDRU's representation is monoid-like. In Fig.~\ref{fig:tsne_embeddings}, we visualize the embeddings of all sequences up to length 12 using t-SNE~\citep{vandermaaten2008visualizing}. The upper and lower plots show embeddings of MLP-LDRU trained up to lengths 40 and 150, respectively. The embeddings display clustering into ECs, supporting our expectation that the MLP-LDRU's induced representation aligns with monoid structures. If MLP-LDRU had learned to behave as a DFA, we would expect fewer clusters (8 states) compared to the syntactic monoid (141 classes).

We further analyze the embeddings using $k$-means clustering, varying $k$ around 141, the number of ECs in the $D_6$ syntactic monoid. We report the Adjusted Rand Index~\citep[ARI;][]{hubert1985comparing} and Silhouette Score~\citep{rousseeuw1987silhouettes} in Fig.~\ref{fig:clustering_plots}. The ARI measures the similarity between fitted clusters and the true ECs, while the Silhouette Score quantifies the separation between clusters. The length 150 model attains a higher ARI than the length 40 model, indicating improved alignment with ECs. The Silhouette Score peaks at $k=141$ for the length 150 model, consistent with clustering at the expected granularity of the monoid. In contrast, the length 40 model has no clear peak, reflecting weaker alignment to the monoid.

\section{Analysis of Monoid Compositions}
\label{app:monoid_compositions}

This section describes how we computed the equivalence class composition patterns shown in Fig.~\ref{fig:monoid_compositions} from the main text and further explains why training sequence length directly impacts generalization performance on $D_n$ languages. Fig.~\ref{fig:monoid_compositions} presents the composition patterns for $D_6$ over varying training and test lengths. To illustrate the explicit structure of monoids for the $D_n$ languages, we present the complete monoid table for $D_2$ in Table~\ref{tab:d2_monoid}, which recognizes the language $(0(01)^* 1)^*$. We present $D_2$ because it has 15 equivalence classes, and $D_6$ is significantly more complex with 141 classes. In general, the monoid of $D_n$ has $1 + \frac{(n+1)(n+2)(2n+3)}{6}$ equivalence classes. The complete $D_2$ automaton has three states: state 0 (initial/accepting), state 1 (after reading `0'), state 2 (the bottom of the fixed stack), and state 3 (rejecting sink state).

However, we focus on the equivalence classes that can be components of positive examples to reduce the complexity compared to modeling all compositions of possible subsequences in $D_n$. This restricts the equivalence classes that we examine to those that contain even-length sequences because of the balanced nature of $D_n$. Despite only considering a subset of equivalence classes (73 for $D_6$), there can still be many classes, so we focus on the $D_6$ language instead of $D_8$ or $D_{12}$ for tractability and simpler visualizations.

We sample sequences of varying lengths in the same fashion as our experimental setup, but restrict the sampling to only positive sequences. We model the equivalence classes produced by a balanced reduction using a monoid composition table. To ensure a balanced distribution of equivalence classes for each length bucket, we target $\approx$1M equivalence classes for each heatmap in Fig.~\ref{fig:monoid_compositions}. We enforce this by tracking the total number of classes observed during reduction and scaling the number of sampled sequences accordingly.

\begin{table}[htbp]
\centering
\caption{Monoid elements for $D_2$. Each equivalence class is characterized by its state mapping function and representative sequences. All other sequences in $\Sigma^\ast$ can be characterized as one of these equivalence classes. A representative sequence is shown in the third column, and the description of each class is provided in the fourth column. The symbol $\epsilon$ denotes the empty sequence.}
\label{tab:d2_monoid}
\begin{tabular}{crrr}
\toprule
\textbf{Element} & \textbf{State Mapping} & \textbf{Representative} & \textbf{Description} \\
\midrule
$e_0$ & $\{0 \mapsto 0, 1 \mapsto 1, 2 \mapsto 2, 3 \mapsto 3\}$ & $\epsilon$ & Identity \\
$e_1$ & $\{0 \mapsto 1, 1 \mapsto 2, 2 \mapsto 3, 3 \mapsto 3\}$ & $0$ & Stack push \\
$e_2$ & $\{0 \mapsto 3, 1 \mapsto 0, 2 \mapsto 1, 3 \mapsto 3\}$ & $1$ & Stack pop \\
\midrule
$e_3$ & $\{0 \mapsto 2, 1 \mapsto 3, 2 \mapsto 3, 3 \mapsto 3\}$ & $00$ & Double stack push \\
$e_4$ & $\{0 \mapsto 0, 1 \mapsto 1, 2 \mapsto 3, 3 \mapsto 3\}$ & $01$ & Partial identity \\
$e_5$ & $\{0 \mapsto 3, 1 \mapsto 1, 2 \mapsto 2, 3 \mapsto 3\}$ & $10$ & Partial identity \\
$e_6$ & $\{0 \mapsto 3, 1 \mapsto 3, 2 \mapsto 0, 3 \mapsto 3\}$ & $11$ & Double stack pop \\
\midrule
$e_7$ & $\{0 \mapsto 3, 1 \mapsto 3, 2 \mapsto 3, 3 \mapsto 3\}$ & $000$ & Annihilation \\
$e_8$ & $\{0 \mapsto 1, 1 \mapsto 3, 2 \mapsto 3, 3 \mapsto 3\}$ & $001$ & Stack push \\
$e_9$ & $\{0 \mapsto 3, 1 \mapsto 0, 2 \mapsto 3, 3 \mapsto 3\}$ & $011$ & Stack pop \\
$e_{10}$ & $\{0 \mapsto 3, 1 \mapsto 2, 2 \mapsto 3, 3 \mapsto 3\}$ & $100$ & Stack push \\
$e_{11}$ & $\{0 \mapsto 3, 1 \mapsto 3, 2 \mapsto 1, 3 \mapsto 3\}$ & $110$ & Stack pop \\
\midrule
$e_{12}$ & $\{0 \mapsto 0, 1 \mapsto 3, 2 \mapsto 3, 3 \mapsto 3\}$ & $0011$ & Partial identity \\
$e_{13}$ & $\{0 \mapsto 3, 1 \mapsto 1, 2 \mapsto 3, 3 \mapsto 3\}$ & $0110$ & Partial identity \\
$e_{14}$ & $\{0 \mapsto 3, 1 \mapsto 3, 2 \mapsto 2, 3 \mapsto 3\}$ & $1100$ & Partial identity \\

\bottomrule
\end{tabular}
\end{table}

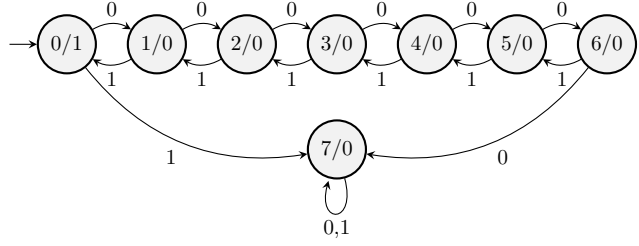
\begin{figure}[!htbp]
    \centering
    \begin{tikzpicture}[scale=0.8, every node/.style={transform shape}]
    \node[state, initial] (q0) {$0/1$};
    \node[state, right=0.5cm of q0] (q1) {$1/0$};
    \node[state, right=0.5cm of q1] (q2) {$2/0$};
    \node[state, right=0.5cm of q2] (q3) {$3/0$};
    \node[state, right=0.5cm of q3] (q4) {$4/0$};
    \node[state, right=0.5cm of q4] (q5) {$5/0$};
    \node[state, right=0.5cm of q5] (q6) {$6/0$};
    \node[state, below=0.75cm of q3] (q7) {$7/0$};

    \path[->] 
    (q0) edge[bend left] node[auto=left] {0} (q1)
    (q1) edge[bend left] node[auto=left] {0} (q2)
    (q2) edge[bend left] node[auto=left] {0} (q3)
    (q1) edge[bend left] node[auto=left] {1} (q0)
    (q2) edge[bend left] node[auto=left] {1} (q1)
    (q3) edge[bend left] node[auto=left] {1} (q2)
    (q3) edge[bend left] node[auto=left] {0} (q4)
    (q4) edge[bend left] node[auto=left] {0} (q5)
    (q5) edge[bend left] node[auto=left] {0} (q6)
    (q4) edge[bend left] node[auto=left] {1} (q3)
    (q5) edge[bend left] node[auto=left] {1} (q4)
    (q6) edge[bend left] node[auto=left] {1} (q5)
    (q6) edge[bend left] node[auto=left] {0} (q7)
    (q7) edge[loop below] node[auto] {0,1} (q7)
    (q0) edge[bend right] node[auto=right] {1} (q7);
    \end{tikzpicture}
    \caption{A complete DFA that recognizes $D_6$. We include the rejecting sink state (a state such that if a sequence reaches it, then it is rejected as it is malformed), omitted from the $D_3$ figure in Table~\ref{tab:bhattamishra_tasks}, to better illustrate the state mappings of the monoid.}
    \label{fig:d6_dfa}
\end{figure}

\textbf{Monoid Computation} For $D_6$, we compute the monoid by first constructing the DFA recognizing the language (see Fig.~\ref{fig:d6_dfa}) and then extracting equivalence classes by generating all sequences up to a fixed length and grouping them by their induced state mappings in the DFA. We enumerated all even-length sequences up to length 12 and found this to be sufficient empirically: longer even-length sequences did not introduce new ECs. We restrict to even lengths because our analysis is over positive sequences (which must be even length). This restriction is natural because the balanced reduction composes even-length subsequences (2, 4, 8, \ldots), so only ECs with even-length representatives arise in this setting. The equivalence classes we do not discover are those that only contain odd-length sequences, meaning that we can characterize any even-length sequence by dividing it into smaller even-length sequences.

\textbf{Reduction Simulation} We simulate the reduction process to estimate the frequency of monoid element compositions that occur during training and testing:

\begin{enumerate}
\item We determine the number of sequences to generate based on fixing the number of equivalence class compositions we want to observe. We fix the number of compositions to 1,000,000 within a range of sequence lengths and vary the number of samples per length within the range to ensure an equal number of equivalence class compositions per length. The ranges of sequence lengths we examined were 10--40 (step 2), 42--60 (step 2), 62--100 (step 4), 102--150 (step 4), and 480--500 (step 2).
\item We determine the equivalence classes of even-length subsequences that reflect increasing depth in the reduction: the first are the 2-length, then 4-length, then 8-length subsequences, and so on until the subsequence is the entire sequence. Each of these subsequences will be composed by the reduction operator as the token embeddings are aggregated.
\item During reduction simulation, we count all pairwise compositions $(e_i, e_j) \rightarrow e_k$ that occur when applying the monoid operator $\odot$. We do not count the compositions that result in the equivalence classes representing the 2-length subsequences because they are only 4 types of compositions (binary alphabet) and would dominate the composition counts.
\item We record composition frequencies to generate probability distributions over monoid element pairs.
\end{enumerate}

The simulation replicates the exact reduction binary tree structure, ensuring that recorded compositions match those encountered during actual model training. The heatmaps in Fig.~\ref{fig:monoid_compositions} reveal insights into why training sequence length influences generalization success:

\textbf{Analysis} Positive sequences of length 10--40 exhibit sparse composition patterns, with most equivalence class pairs showing low log probabilities (lighter regions). The concentration of high-probability compositions in the lower-index region (equivalence classes 0--15) occurs because these represent the most frequently encountered partial sequences in shorter training data. As training sequence length increases to 62--100 and beyond, additional composition patterns emerge. As equivalence classes are discovered dynamically from the generated sequences with increasing length, higher index equivalence classes tend to be rarer and sharper state mappings (i.e., an increasing number of annihilations, see Table~\ref{tab:d2_monoid}).

Complete generalization requires some exposure to all possible monoid element compositions that can occur during testing. The difference in sparsity between training and testing heatmaps directly explains the empirical results: insufficient training sequence length fails to provide sufficient coverage of rare monoid compositions. Moreover, some languages are likely to be impossible to learn (i.e., generalize to) without a sufficient maximum training sequence length, as they require compositions between equivalence classes that are never encountered in shorter sequences.

%% file: tables/appendix/full_results_new.tex
\begin{table*}[!htbp]
\caption{OOD accuracy results across 21 regular tasks. All models are trained on sequences with lengths 1--40 and evaluated on lengths 41--500. Results show mean $\pm$ standard deviation across seeds. A $^{\downarrow}$ indicates MLP-LDRU performance is statistically significantly better ($p < 0.05$) than the baseline, while a $^{\uparrow}$ indicates the baseline is statistically significantly better than MLP-LDRU.}
\label{tab:full_results}
\centering
\resizebox{\textwidth}{!}{\begin{tabular}{@{}lccccccccccc@{}}
\toprule
Task &
  RNN &
  LSTM &
  Gated DeltaNet &
  TF (NoPE) &
  TF (ALiBi) &
  TF ($\sim$RoPE) &
  S5 &
  SD-SSM &
  RegularGPT &
  BBT-GRC &
  MLP-LDRU \\ \midrule
\textit{1)~\citet{deletang2023neural}} &
   &
   &
   &
   &
   &
   &
   &
   &
   &
   &
   \\
Even~Pairs &
  ${77.4 \pm 12.2}$ &
  $100.0 \pm 0.0$ &
  ${52.2 \pm 0.8}^{\downarrow}$ &
  ${50.3 \pm 0.0}^{\downarrow}$ &
  $61.1 \pm 5.1^{\downarrow}$ &
  $91.5 \pm 7.9$ &
  ${53.9 \pm 0.8}^{\downarrow}$ &
  ${65.7 \pm 0.8}^{\downarrow}$ &
  ${91.9 \pm 0.7}^{\downarrow}$ &
  $100.0 \pm 0.0$ &
  $\mathbf{100.0 \pm 0.0}$ \\
Modular~Arithmetic &
  $100.0 \pm 0.0$ &
  $100.0 \pm 0.0$ &
  ${76.5 \pm 3.1}^{\downarrow}$ &
  ${76.0 \pm 0.0}^{\downarrow}$ &
  $54.4 \pm 4.5^{\downarrow}$ &
  $31.4 \pm 5.5^{\downarrow}$ &
  ${47.6 \pm 6.7}^{\downarrow}$ &
  ${99.9 \pm 0.0}^{\downarrow}$ &
  $99.1 \pm 1.3$ &
  $94.0 \pm 3.2$ &
  $\mathbf{100.0 \pm 0.0}$ \\
Parity~Check &
  $100.0 \pm 0.0$ &
  $100.0 \pm 0.0$ &
  ${53.4 \pm 0.3}^{\downarrow}$ &
  ${50.1 \pm 0.0}^{\downarrow}$ &
  $49.9 \pm 0.0^{\downarrow}$ &
  $49.9 \pm 0.0^{\downarrow}$ &
  ${50.1 \pm 0.1}^{\downarrow}$ &
  ${71.4 \pm 1.4}^{\downarrow}$ &
  $99.8 \pm 0.7$ &
  $100.0 \pm 0.0$ &
  $\mathbf{100.0 \pm 0.0}$ \\
Cycle~Navigation &
  $100.0 \pm 0.0$ &
  ${60.3 \pm 2.4}^{\downarrow}$ &
  ${25.8 \pm 3.8}^{\downarrow}$ &
  ${20.0 \pm 0.0}^{\downarrow}$ &
  $21.6 \pm 0.7^{\downarrow}$ &
  $23.3 \pm 1.5^{\downarrow}$ &
  ${22.9 \pm 0.7}^{\downarrow}$ &
  ${44.2 \pm 0.8}^{\downarrow}$ &
  $99.9 \pm 0.2$ &
  $100.0 \pm 0.1$ &
  $\mathbf{100.0 \pm 0.0}$ \\ \midrule
\textit{2)~\citet{Bhattamishra2020OnTA}} &
   &
   &
   &
   &
   &
   &
   &
   &
   &
   &
   \\
$D_2$ &
  $100.0 \pm 0.0$ &
  $100.0 \pm 0.0$ &
  $100.0 \pm 0.0$ &
  ${100.0 \pm 0.0}^{\downarrow}$ &
  $100.0 \pm 0.0$ &
  $98.2 \pm 1.7$ &
  $88.4 \pm 10.3$ &
  $100.0 \pm 0.0$ &
  $93.7 \pm 5.5$ &
  $100.0 \pm 0.0$ &
  $\mathbf{100.0 \pm 0.0}$ \\
$D_3$ &
  $100.0 \pm 0.0$ &
  $97.7 \pm 4.0$ &
  $91.6 \pm 14.3$ &
  ${99.9 \pm 0.0}^{\downarrow}$ &
  $98.3 \pm 2.6$ &
  $92.5 \pm 10.2$ &
  $83.6 \pm 9.3$ &
  $96.3 \pm 3.9$ &
  ${87.6 \pm 4.6}^{\downarrow}$ &
  $98.5 \pm 2.3$ &
  $\mathbf{100.0 \pm 0.0}$ \\
$D_4$ &
  $95.0 \pm 8.7$ &
  $100.0 \pm 0.0$ &
  $93.1 \pm 6.0$ &
  ${99.5 \pm 0.0}^{\downarrow}$ &
  $96.0 \pm 4.7$ &
  $97.6 \pm 0.4^{\downarrow}$ &
  $81.7 \pm 10.3$ &
  $97.3 \pm 2.3$ &
  $94.1 \pm 4.7$ &
  ${84.6 \pm 4.6}^{\downarrow}$ &
  $\mathbf{100.0 \pm 0.0}$ \\
$D_6$ &
  $97.1 \pm 4.7$ &
  $98.4 \pm 1.6$ &
  ${75.6 \pm 0.6}^{\downarrow}$ &
  ${96.5 \pm 0.0}$ &
  $92.3 \pm 6.8$ &
  $98.4 \pm 1.0$ &
  ${75.0 \pm 1.5}^{\downarrow}$ &
  $92.2 \pm 4.3$ &
  ${87.9 \pm 3.7}^{\downarrow}$ &
  ${84.7 \pm 6.0}^{\downarrow}$ &
  $\mathbf{98.5 \pm 2.4}$ \\
$D_8$ &
  $\mathbf{99.2 \pm 1.3}$ &
  ${89.0 \pm 4.3}$ &
  ${76.0 \pm 1.3}^{\downarrow}$ &
  ${89.0 \pm 0.2}^{\downarrow}$ &
  $89.7 \pm 7.7$ &
  $98.7 \pm 0.8$ &
  ${73.7 \pm 0.5}^{\downarrow}$ &
  $87.2 \pm 2.9$ &
  ${91.0 \pm 2.8}^{\downarrow}$ &
  $84.3 \pm 8.0$ &
  $98.1 \pm 1.6$ \\
$D_{12}$ &
  $92.7 \pm 9.5$ &
  ${82.1 \pm 1.7}^{\downarrow}$ &
  ${74.6 \pm 0.5}^{\downarrow}$ &
  ${81.3 \pm 0.3}^{\downarrow}$ &
  $84.4 \pm 9.6$ &
  $\mathbf{98.7 \pm 1.0}^{\uparrow}$ &
  ${73.5 \pm 0.3}^{\downarrow}$ &
  ${72.9 \pm 4.3}^{\downarrow}$ &
  ${89.8 \pm 4.1}^{\downarrow}$ &
  $90.5 \pm 0.7$ &
  $96.0 \pm 2.1$ \\
Tomita~3 &
  $100.0 \pm 0.0$ &
  $100.0 \pm 0.0$ &
  $99.9 \pm 0.1$ &
  ${98.7 \pm 0.0}^{\downarrow}$ &
  $100.0 \pm 0.0^{\downarrow}$ &
  $70.7 \pm 8.9^{\downarrow}$ &
  ${72.8 \pm 1.6}^{\downarrow}$ &
  $100.0 \pm 0.0$ &
  $93.7 \pm 3.3$ &
  $99.4 \pm 0.7$ &
  $\mathbf{100.0 \pm 0.0}$ \\
Tomita~4 &
  $100.0 \pm 0.0$ &
  $100.0 \pm 0.0$ &
  ${99.1 \pm 0.2}^{\downarrow}$ &
  ${96.1 \pm 0.0}^{\downarrow}$ &
  $100.0 \pm 0.0$ &
  $85.5 \pm 2.1^{\downarrow}$ &
  $72.3 \pm 12.8$ &
  $100.0 \pm 0.0$ &
  $98.4 \pm 1.5$ &
  $99.4 \pm 1.1$ &
  $\mathbf{100.0 \pm 0.0}$ \\
Tomita~5 &
  $100.0 \pm 0.0$ &
  $100.0 \pm 0.0$ &
  ${75.1 \pm 0.4}^{\downarrow}$ &
  ${74.3 \pm 0.0}^{\downarrow}$ &
  $74.3 \pm 0.0^{\downarrow}$ &
  $74.3 \pm 0.0^{\downarrow}$ &
  ${65.8 \pm 3.2}^{\downarrow}$ &
  ${81.1 \pm 1.8}^{\downarrow}$ &
  $97.7 \pm 2.2$ &
  $100.0 \pm 0.0$ &
  $\mathbf{100.0 \pm 0.0}$ \\
Tomita~6 &
  $100.0 \pm 0.0$ &
  $100.0 \pm 0.0$ &
  ${54.7 \pm 0.5}^{\downarrow}$ &
  ${50.2 \pm 0.0}^{\downarrow}$ &
  $50.0 \pm 0.0^{\downarrow}$ &
  $50.0 \pm 0.0^{\downarrow}$ &
  ${49.9 \pm 0.1}^{\downarrow}$ &
  ${85.6 \pm 3.7}^{\downarrow}$ &
  $92.2 \pm 4.7$ &
  $100.0 \pm 0.0$ &
  $\mathbf{100.0 \pm 0.0}$ \\
Tomita~7 &
  $100.0 \pm 0.0$ &
  $100.0 \pm 0.0$ &
  $100.0 \pm 0.0$ &
  ${100.0 \pm 0.0}$ &
  $100.0 \pm 0.0$ &
  $100.0 \pm 0.0$ &
  $98.7 \pm 2.2$ &
  $100.0 \pm 0.0$ &
  $100.0 \pm 0.0$ &
  $100.0 \pm 0.0$ &
  $\mathbf{100.0 \pm 0.0}$ \\ \midrule
\textit{3)}~Prefix languages (Ours) &
   &
   &
   &
   &
   &
   &
   &
   &
   &
   &
   \\
$P_{1,2}$ &
  ${82.2 \pm 3.3}^{\downarrow}$ &
  $100.0 \pm 0.0$ &
  ${52.4 \pm 1.7}^{\downarrow}$ &
  ${50.3 \pm 0.0}^{\downarrow}$ &
  $56.0 \pm 2.3^{\downarrow}$ &
  $89.2 \pm 12.7$ &
  ${52.9 \pm 2.8}^{\downarrow}$ &
  ${61.8 \pm 3.5}^{\downarrow}$ &
  $100.0 \pm 0.0$ &
  $100.0 \pm 0.0$ &
  $\mathbf{100.0 \pm 0.0}$ \\
$P_{2,2}$ &
  ${65.4 \pm 24.0}$ &
  $100.0 \pm 0.0$ &
  ${22.0 \pm 11.9}^{\downarrow}$ &
  ${25.2 \pm 0.0}^{\downarrow}$ &
  $93.0 \pm 5.6$ &
  $86.4 \pm 5.0^{\downarrow}$ &
  ${29.3 \pm 1.8}^{\downarrow}$ &
  ${34.0 \pm 1.6}^{\downarrow}$ &
  $99.6 \pm 0.3$ &
  $99.9 \pm 0.1$ &
  $\mathbf{100.0 \pm 0.0}$ \\
$P_{4,2}$ &
  $99.4 \pm 1.0$ &
  $100.0 \pm 0.0$ &
  ${18.9 \pm 1.6}^{\downarrow}$ &
  ${6.3 \pm 0.0}^{\downarrow}$ &
  $90.0 \pm 10.3$ &
  $58.5 \pm 11.1^{\downarrow}$ &
  ${12.1 \pm 2.1}^{\downarrow}$ &
  $99.8 \pm 0.2$ &
  ${91.5 \pm 2.1}^{\downarrow}$ &
  $100.0 \pm 0.0$ &
  $\mathbf{100.0 \pm 0.0}$ \\
$P_{1,4}$ &
  ${61.4 \pm 15.7}$ &
  $100.0 \pm 0.0$ &
  ${30.1 \pm 1.6}^{\downarrow}$ &
  ${25.2 \pm 0.0}^{\downarrow}$ &
  $40.3 \pm 2.4^{\downarrow}$ &
  $79.5 \pm 7.0^{\downarrow}$ &
  ${31.8 \pm 3.6}^{\downarrow}$ &
  ${38.7 \pm 0.5}^{\downarrow}$ &
  $100.0 \pm 0.0$ &
  $100.0 \pm 0.0$ &
  $\mathbf{100.0 \pm 0.0}$ \\
$P_{2,4}$ &
  $96.9 \pm 2.5$ &
  $100.0 \pm 0.0$ &
  ${21.5 \pm 9.5}^{\downarrow}$ &
  ${6.3 \pm 0.0}^{\downarrow}$ &
  $31.0 \pm 8.1^{\downarrow}$ &
  $51.6 \pm 7.6^{\downarrow}$ &
  ${14.4 \pm 1.3}^{\downarrow}$ &
  ${30.2 \pm 4.9}^{\downarrow}$ &
  $99.7 \pm 0.2$ &
  $100.0 \pm 0.0$ &
  $\mathbf{100.0 \pm 0.0}$ \\
$P_{4,4}$ &
  ${85.0 \pm 15.0}$ &
  ${97.3 \pm 2.2}$ &
  ${69.4 \pm 6.0}^{\downarrow}$ &
  ${0.4 \pm 0.0}^{\downarrow}$ &
  $21.3 \pm 7.6^{\downarrow}$ &
  $58.2 \pm 2.7^{\downarrow}$ &
  ${8.0 \pm 3.4}^{\downarrow}$ &
  ${42.9 \pm 2.5}^{\downarrow}$ &
  ${95.3 \pm 0.6}^{\downarrow}$ &
  $100.0 \pm 0.0$ &
  $\mathbf{100.0 \pm 0.0}$ \\ \bottomrule
\end{tabular}
}
\end{table*}

%% file: tables/appendix/full_results_ID.tex
\begin{table*}[!htbp]
\caption{ID accuracies across 21 regular tasks. All models are trained on sequences with lengths 1--40 and evaluated on lengths 41--500. Results show mean $\pm$ standard deviation across seeds.}
\label{tab:full_results_id}
\centering
\resizebox{\textwidth}{!}{\begin{tabular}{@{}lccccccccccc@{}}
\toprule
Task &
  RNN &
  LSTM &
  Gated DeltaNet &
  TF (NoPE) &
  TF (ALiBi) &
  TF ($\sim$RoPE) &
  S5 &
  SD-SSM &
  RegularGPT &
  \multicolumn{1}{l}{BBT-GRC} &
  MLP-LDRU \\ \midrule
\textit{1)~\citet{deletang2023neural}} &
   &
   &
   &
   &
   &
   &
   &
   &
   &
  \multicolumn{1}{l}{} &
   \\
Even$\sim$Pairs &
  $100.0 \pm 0.0$ &
  $100.0 \pm 0.0$ &
  $100.0 \pm 0.0$ &
  $90.1 \pm 0.0$ &
  $100.0 \pm 0.0$ &
  $99.4 \pm 0.9$ &
  $100.0 \pm 0.0$ &
  $100.0 \pm 0.0$ &
  $100.0 \pm 0.0$ &
  $100.0 \pm 0.0$ &
  $100.0 \pm 0.0$ \\
Modular$\sim$Arithmetic &
  $100.0 \pm 0.0$ &
  $100.0 \pm 0.0$ &
  $85.4 \pm 2.6$ &
  $94.9 \pm 0.0$ &
  $86.3 \pm 1.0$ &
  $75.7 \pm 5.3$ &
  $67.6 \pm 0.7$ &
  $100.0 \pm 0.0$ &
  $100.0 \pm 0.0$ &
  $99.9 \pm 0.1$ &
  $100.0 \pm 0.0$ \\
Parity$\sim$Check &
  $100.0 \pm 0.0$ &
  $100.0 \pm 0.0$ &
  $100.0 \pm 0.0$ &
  $90.1 \pm 0.0$ &
  $51.9 \pm 0.3$ &
  $56.6 \pm 1.2$ &
  $99.1 \pm 0.1$ &
  $100.0 \pm 0.0$ &
  $100.0 \pm 0.0$ &
  $100.0 \pm 0.0$ &
  $100.0 \pm 0.0$ \\
Cycle$\sim$Navigation &
  $100.0 \pm 0.0$ &
  $100.0 \pm 0.0$ &
  $99.3 \pm 0.4$ &
  $84.8 \pm 0.1$ &
  $84.7 \pm 7.1$ &
  $89.7 \pm 5.2$ &
  $100.0 \pm 0.0$ &
  $100.0 \pm 0.0$ &
  $100.0 \pm 0.0$ &
  $100.0 \pm 0.0$ &
  $100.0 \pm 0.0$ \\ \midrule
\textit{2)~\citet{Bhattamishra2020OnTA}} &
   &
   &
   &
   &
   &
   &
   &
   &
   &
  \multicolumn{1}{l}{} &
   \\
$D_2$ &
  $100.0 \pm 0.0$ &
  $100.0 \pm 0.0$ &
  $100.0 \pm 0.0$ &
  $100.0 \pm 0.0$ &
  $98.7 \pm 1.9$ &
  $98.7 \pm 0.4$ &
  $100.0 \pm 0.0$ &
  $100.0 \pm 0.0$ &
  $100.0 \pm 0.0$ &
  $100.0 \pm 0.0$ &
  $100.0 \pm 0.0$ \\
$D_3$ &
  $100.0 \pm 0.0$ &
  $100.0 \pm 0.0$ &
  $93.8 \pm 10.6$ &
  $100.0 \pm 0.0$ &
  $99.2 \pm 1.0$ &
  $99.1 \pm 0.6$ &
  $100.0 \pm 0.0$ &
  $100.0 \pm 0.0$ &
  $100.0 \pm 0.0$ &
  $100.0 \pm 0.0$ &
  $100.0 \pm 0.0$ \\
$D_4$ &
  $100.0 \pm 0.0$ &
  $100.0 \pm 0.0$ &
  $98.4 \pm 2.6$ &
  $99.9 \pm 0.0$ &
  $99.8 \pm 0.0$ &
  $99.6 \pm 0.0$ &
  $100.0 \pm 0.0$ &
  $100.0 \pm 0.0$ &
  $100.0 \pm 0.0$ &
  $99.9 \pm 0.0$ &
  $100.0 \pm 0.0$ \\
$D_6$ &
  $100.0 \pm 0.0$ &
  $100.0 \pm 0.0$ &
  $98.7 \pm 0.5$ &
  $99.3 \pm 0.0$ &
  $99.8 \pm 0.1$ &
  $99.8 \pm 0.1$ &
  $100.0 \pm 0.0$ &
  $100.0 \pm 0.0$ &
  $100.0 \pm 0.0$ &
  $99.9 \pm 0.0$ &
  $100.0 \pm 0.0$ \\
$D_8$ &
  $100.0 \pm 0.0$ &
  $100.0 \pm 0.0$ &
  $99.4 \pm 0.8$ &
  $98.9 \pm 0.0$ &
  $99.8 \pm 0.0$ &
  $99.8 \pm 0.1$ &
  $100.0 \pm 0.0$ &
  $100.0 \pm 0.0$ &
  $100.0 \pm 0.0$ &
  $99.9 \pm 0.0$ &
  $100.0 \pm 0.0$ \\
$D_{12}$ &
  $100.0 \pm 0.0$ &
  $100.0 \pm 0.0$ &
  $96.7 \pm 4.8$ &
  $98.7 \pm 0.0$ &
  $99.6 \pm 0.3$ &
  $99.9 \pm 0.0$ &
  $100.0 \pm 0.0$ &
  $100.0 \pm 0.0$ &
  $100.0 \pm 0.0$ &
  $99.9 \pm 0.0$ &
  $100.0 \pm 0.0$ \\
Tomita$\sim$3 &
  $100.0 \pm 0.0$ &
  $100.0 \pm 0.0$ &
  $99.2 \pm 1.4$ &
  $99.8 \pm 0.0$ &
  $99.1 \pm 0.0$ &
  $96.8 \pm 0.6$ &
  $100.0 \pm 0.0$ &
  $100.0 \pm 0.0$ &
  $100.0 \pm 0.0$ &
  $100.0 \pm 0.0$ &
  $100.0 \pm 0.0$ \\
Tomita$\sim$4 &
  $100.0 \pm 0.0$ &
  $100.0 \pm 0.0$ &
  $100.0 \pm 0.0$ &
  $99.7 \pm 0.0$ &
  $98.9 \pm 0.1$ &
  $98.7 \pm 0.1$ &
  $100.0 \pm 0.0$ &
  $100.0 \pm 0.0$ &
  $100.0 \pm 0.0$ &
  $100.0 \pm 0.0$ &
  $100.0 \pm 0.0$ \\
Tomita$\sim$5 &
  $100.0 \pm 0.0$ &
  $100.0 \pm 0.0$ &
  $100.0 \pm 0.0$ &
  $93.4 \pm 0.0$ &
  $74.2 \pm 0.0$ &
  $74.2 \pm 0.0$ &
  $97.9 \pm 2.0$ &
  $100.0 \pm 0.0$ &
  $100.0 \pm 0.0$ &
  $100.0 \pm 0.0$ &
  $100.0 \pm 0.0$ \\
Tomita$\sim$6 &
  $100.0 \pm 0.0$ &
  $100.0 \pm 0.0$ &
  $99.9 \pm 0.1$ &
  $89.7 \pm 0.1$ &
  $51.3 \pm 0.0$ &
  $51.3 \pm 0.0$ &
  $93.5 \pm 10.5$ &
  $100.0 \pm 0.0$ &
  $100.0 \pm 0.0$ &
  $100.0 \pm 0.0$ &
  $100.0 \pm 0.0$ \\
Tomita$\sim$7 &
  $100.0 \pm 0.0$ &
  $100.0 \pm 0.0$ &
  $100.0 \pm 0.0$ &
  $100.0 \pm 0.0$ &
  $99.0 \pm 0.3$ &
  $99.9 \pm 0.0$ &
  $100.0 \pm 0.0$ &
  $100.0 \pm 0.0$ &
  $100.0 \pm 0.0$ &
  $100.0 \pm 0.0$ &
  $100.0 \pm 0.0$ \\ \midrule
\textit{3)}~Prefix languages (Ours) &
   &
   &
   &
   &
   &
   &
   &
   &
   &
   &
   \\
$P_{1,2}$ &
  $100.0 \pm 0.0$ &
  $100.0 \pm 0.0$ &
  $98.8 \pm 1.3$ &
  $90.1 \pm 0.0$ &
  $99.9 \pm 0.1$ &
  $98.8 \pm 2.1$ &
  $100.0 \pm 0.0$ &
  $100.0 \pm 0.0$ &
  $100.0 \pm 0.0$ &
  $100.0 \pm 0.0$ &
  $100.0 \pm 0.0$ \\
$P_{2,2}$ &
  $100.0 \pm 0.0$ &
  $100.0 \pm 0.0$ &
  $100.0 \pm 0.0$ &
  $85.0 \pm 0.0$ &
  $97.4 \pm 0.1$ &
  $96.2 \pm 1.1$ &
  $100.0 \pm 0.0$ &
  $100.0 \pm 0.0$ &
  $100.0 \pm 0.0$ &
  $100.0 \pm 0.0$ &
  $100.0 \pm 0.0$ \\
$P_{4,2}$ &
  $100.0 \pm 0.0$ &
  $100.0 \pm 0.0$ &
  $100.0 \pm 0.0$ &
  $81.3 \pm 0.0$ &
  $99.3 \pm 0.1$ &
  $97.5 \pm 0.9$ &
  $100.0 \pm 0.0$ &
  $100.0 \pm 0.0$ &
  $100.0 \pm 0.0$ &
  $100.0 \pm 0.0$ &
  $100.0 \pm 0.0$ \\
$P_{1,4}$ &
  $100.0 \pm 0.0$ &
  $100.0 \pm 0.0$ &
  $100.0 \pm 0.0$ &
  $85.0 \pm 0.0$ &
  $99.8 \pm 0.1$ &
  $99.9 \pm 0.1$ &
  $100.0 \pm 0.0$ &
  $100.0 \pm 0.0$ &
  $100.0 \pm 0.0$ &
  $100.0 \pm 0.0$ &
  $100.0 \pm 0.0$ \\
$P_{2,4}$ &
  $100.0 \pm 0.0$ &
  $100.0 \pm 0.0$ &
  $100.0 \pm 0.0$ &
  $81.2 \pm 0.0$ &
  $98.3 \pm 0.3$ &
  $97.4 \pm 0.1$ &
  $100.0 \pm 0.0$ &
  $100.0 \pm 0.0$ &
  $100.0 \pm 0.0$ &
  $100.0 \pm 0.0$ &
  $100.0 \pm 0.0$ \\
$P_{4,2}$ &
  $100.0 \pm 0.0$ &
  $100.0 \pm 0.0$ &
  $100.0 \pm 0.0$ &
  $81.3 \pm 0.0$ &
  $99.3 \pm 0.1$ &
  $97.5 \pm 0.9$ &
  $100.0 \pm 0.0$ &
  $100.0 \pm 0.0$ &
  $100.0 \pm 0.0$ &
  $100.0 \pm 0.0$ &
  $100.0 \pm 0.0$ \\
$P_{4,4}$ &
  $100.0 \pm 0.0$ &
  $100.0 \pm 0.0$ &
  $100.0 \pm 0.0$ &
  $78.4 \pm 0.1$ &
  $99.0 \pm 0.3$ &
  $98.8 \pm 0.3$ &
  $100.0 \pm 0.0$ &
  $100.0 \pm 0.0$ &
  $100.0 \pm 0.0$ &
  $100.0 \pm 0.0$ &
  $100.0 \pm 0.0$ \\ \bottomrule
\end{tabular}
}
\end{table*}

%% file: tables/appendix/individual_results_main_new.tex
\begin{table*}[!htbp]
\centering
\caption{Individual seed results for all 21 regular language tasks. Models trained on sequences up to length 40, evaluated on lengths 41--500.}
\label{tab:individual_seeds_main}
\resizebox{\textwidth}{!}{\begin{tabular}{@{}lccccccccccccccccccccccccccccccccc@{}}
\toprule
 &
  \multicolumn{3}{c}{RNN} &
  \multicolumn{3}{c}{LSTM} &
  \multicolumn{3}{c}{Gated DeltaNet} &
  \multicolumn{3}{c}{TF (NoPE)} &
  \multicolumn{3}{c}{TF (ALiBi)} &
  \multicolumn{3}{c}{TF ($\sim$RoPE)} &
  \multicolumn{3}{c}{S5} &
  \multicolumn{3}{c}{SD-SSM} &
  \multicolumn{3}{c}{RegularGPT} &
  \multicolumn{3}{c}{BBT-GRC} &
  \multicolumn{3}{c}{MLP-LDRU} \\ \cmidrule(l){2-34} 
Task &
  0 &
  1 &
  2 &
  0 &
  1 &
  2 &
  0 &
  1 &
  2 &
  0 &
  1 &
  2 &
  0 &
  1 &
  2 &
  0 &
  1 &
  2 &
  0 &
  1 &
  2 &
  0 &
  1 &
  2 &
  0 &
  1 &
  2 &
  0 &
  1 &
  2 &
  0 &
  1 &
  2 \\ \midrule
\textit{1) \citet{deletang2023neural}} &
   &
   &
   &
   &
   &
   &
   &
   &
   &
   &
   &
   &
   &
   &
   &
   &
   &
   &
   &
   &
   &
   &
   &
   &
   &
   &
   &
   &
   &
   &
   &
   &
   \\
Even~Pairs &
  82.4 &
  86.3 &
  63.4 &
  100.0 &
  100.0 &
  100.0 &
  51.7 &
  53.1 &
  51.8 &
  50.3 &
  50.3 &
  50.3 &
  66.0 &
  55.9 &
  61.5 &
  95.0 &
  82.4 &
  97.0 &
  53.9 &
  53.2 &
  54.7 &
  66.5 &
  65.7 &
  64.8 &
  92.2 &
  91.1 &
  92.2 &
  100.0 &
  100.0 &
  100.0 &
  100.0 &
  100.0 &
  100.0 \\
Modular~Arithmetic &
  100.0 &
  100.0 &
  100.0 &
  100.0 &
  100.0 &
  100.0 &
  78.9 &
  73.0 &
  77.6 &
  76.0 &
  76.0 &
  76.0 &
  55.4 &
  58.3 &
  49.5 &
  33.3 &
  35.8 &
  25.2 &
  51.5 &
  51.4 &
  39.9 &
  99.9 &
  99.9 &
  99.9 &
  97.5 &
  99.9 &
  99.8 &
  97.2 &
  90.9 &
  94.0 &
  100.0 &
  100.0 &
  100.0 \\
Parity~Check &
  100.0 &
  100.0 &
  100.0 &
  100.0 &
  100.0 &
  100.0 &
  53.5 &
  53.1 &
  53.8 &
  50.0 &
  50.1 &
  50.1 &
  49.9 &
  49.9 &
  49.9 &
  49.9 &
  49.9 &
  49.9 &
  50.0 &
  50.2 &
  50.1 &
  69.8 &
  72.3 &
  71.9 &
  99.4 &
  100.0 &
  100.0 &
  100.0 &
  100.0 &
  100.0 &
  100.0 &
  100.0 &
  100.0 \\
Cycle~Navigation &
  100.0 &
  100.0 &
  100.0 &
  62.1 &
  57.5 &
  61.2 &
  29.8 &
  25.6 &
  22.1 &
  20.0 &
  20.0 &
  20.0 &
  20.8 &
  22.0 &
  21.9 &
  21.6 &
  24.0 &
  24.3 &
  22.1 &
  23.6 &
  23.0 &
  45.0 &
  44.3 &
  43.4 &
  99.9 &
  100.0 &
  99.7 &
  100.0 &
  99.9 &
  100.0 &
  100.0 &
  100.0 &
  100.0 \\ \midrule
\textit{2) \citet{Bhattamishra2020OnTA}} &
   &
   &
   &
   &
   &
   &
   &
   &
   &
   &
   &
   &
   &
   &
   &
   &
   &
   &
   &
   &
   &
   &
   &
   &
   &
   &
   &
   &
   &
   &
   &
   &
   \\
$D_2$ &
  100.0 &
  100.0 &
  100.0 &
  100.0 &
  100.0 &
  100.0 &
  100.0 &
  100.0 &
  100.0 &
  100.0 &
  100.0 &
  100.0 &
  100.0 &
  100.0 &
  100.0 &
  99.5 &
  96.3 &
  98.8 &
  85.0 &
  100.0 &
  80.2 &
  100.0 &
  100.0 &
  100.0 &
  100.0 &
  89.7 &
  91.5 &
  100.0 &
  100.0 &
  100.0 &
  100.0 &
  100.0 &
  100.0 \\
$D_3$ &
  100.0 &
  100.0 &
  100.0 &
  100.0 &
  93.2 &
  100.0 &
  99.8 &
  99.9 &
  75.1 &
  99.9 &
  99.9 &
  99.9 &
  95.3 &
  100.0 &
  99.5 &
  98.3 &
  98.5 &
  80.8 &
  79.4 &
  94.3 &
  77.2 &
  100.0 &
  96.8 &
  92.1 &
  92.7 &
  83.7 &
  86.4 &
  99.8 &
  95.9 &
  99.9 &
  100.0 &
  100.0 &
  100.0 \\
$D_4$ &
  85.0 &
  100.0 &
  100.0 &
  100.0 &
  100.0 &
  100.0 &
  87.9 &
  91.9 &
  99.6 &
  99.5 &
  99.5 &
  99.5 &
  100.0 &
  90.7 &
  97.2 &
  97.4 &
  97.4 &
  98.1 &
  75.5 &
  93.6 &
  76.1 &
  100.0 &
  96.3 &
  95.7 &
  89.1 &
  94.6 &
  98.5 &
  87.1 &
  87.4 &
  79.2 &
  100.0 &
  100.0 &
  100.0 \\
$D_6$ &
  91.6 &
  99.6 &
  100.0 &
  100.0 &
  98.5 &
  96.8 &
  75.2 &
  76.3 &
  75.3 &
  96.6 &
  96.6 &
  96.5 &
  93.1 &
  85.1 &
  98.7 &
  98.7 &
  97.2 &
  99.1 &
  73.7 &
  76.6 &
  74.7 &
  97.1 &
  89.4 &
  90.2 &
  90.7 &
  83.7 &
  89.3 &
  91.5 &
  80.5 &
  82.0 &
  99.9 &
  95.7 &
  99.9 \\
$D_8$ &
  100.0 &
  97.7 &
  99.9 &
  85.4 &
  93.8 &
  87.8 &
  77.4 &
  75.4 &
  75.0 &
  89.0 &
  89.3 &
  88.8 &
  97.6 &
  89.4 &
  82.1 &
  99.3 &
  97.7 &
  99.0 &
  74.2 &
  73.2 &
  73.6 &
  84.5 &
  86.6 &
  90.4 &
  93.6 &
  88.1 &
  91.4 &
  93.5 &
  79.5 &
  79.9 &
  99.8 &
  98.2 &
  96.6 \\
$D_{12}$ &
  96.9 &
  81.8 &
  99.3 &
  80.2 &
  82.6 &
  83.4 &
  75.0 &
  74.0 &
  74.7 &
  81.1 &
  81.3 &
  81.6 &
  95.3 &
  77.6 &
  80.2 &
  99.3 &
  99.3 &
  97.6 &
  73.9 &
  73.4 &
  73.3 &
  77.7 &
  69.4 &
  71.6 &
  93.5 &
  90.5 &
  85.3 &
  90.3 &
  90.0 &
  91.2 &
  97.5 &
  96.8 &
  93.6 \\
Tomita~3 &
  100.0 &
  100.0 &
  100.0 &
  100.0 &
  100.0 &
  100.0 &
  100.0 &
  99.8 &
  99.9 &
  98.7 &
  98.7 &
  98.7 &
  100.0 &
  100.0 &
  100.0 &
  68.6 &
  80.5 &
  63.1 &
  74.3 &
  71.0 &
  73.1 &
  100.0 &
  100.0 &
  100.0 &
  96.8 &
  94.2 &
  90.2 &
  99.6 &
  98.7 &
  100.0 &
  100.0 &
  100.0 &
  100.0 \\
Tomita~4 &
  100.0 &
  100.0 &
  100.0 &
  100.0 &
  100.0 &
  100.0 &
  99.2 &
  99.3 &
  98.8 &
  96.1 &
  96.1 &
  96.1 &
  100.0 &
  100.0 &
  100.0 &
  83.1 &
  86.5 &
  86.8 &
  58.5 &
  83.7 &
  74.8 &
  100.0 &
  100.0 &
  100.0 &
  96.7 &
  98.9 &
  99.6 &
  98.1 &
  100.0 &
  100.0 &
  100.0 &
  100.0 &
  100.0 \\
Tomita~5 &
  100.0 &
  100.0 &
  100.0 &
  100.0 &
  100.0 &
  100.0 &
  75.6 &
  74.9 &
  74.8 &
  74.3 &
  74.3 &
  74.3 &
  74.3 &
  74.3 &
  74.3 &
  74.3 &
  74.3 &
  74.3 &
  63.0 &
  65.3 &
  69.2 &
  82.7 &
  81.4 &
  79.2 &
  99.5 &
  98.2 &
  95.3 &
  100.0 &
  100.0 &
  100.0 &
  100.0 &
  100.0 &
  100.0 \\
Tomita~6 &
  100.0 &
  100.0 &
  100.0 &
  100.0 &
  100.0 &
  100.0 &
  55.0 &
  54.1 &
  54.9 &
  50.2 &
  50.2 &
  50.2 &
  50.0 &
  50.0 &
  50.0 &
  50.0 &
  50.0 &
  50.0 &
  50.0 &
  50.0 &
  49.7 &
  82.6 &
  84.6 &
  89.7 &
  93.2 &
  87.1 &
  96.4 &
  100.0 &
  100.0 &
  100.0 &
  100.0 &
  100.0 &
  100.0 \\
Tomita~7 &
  100.0 &
  100.0 &
  100.0 &
  100.0 &
  100.0 &
  100.0 &
  100.0 &
  100.0 &
  100.0 &
  100.0 &
  100.0 &
  100.0 &
  100.0 &
  100.0 &
  100.0 &
  100.0 &
  100.0 &
  100.0 &
  96.2 &
  100.0 &
  100.0 &
  100.0 &
  100.0 &
  100.0 &
  100.0 &
  100.0 &
  100.0 &
  100.0 &
  100.0 &
  100.0 &
  100.0 &
  100.0 &
  100.0 \\ \midrule
\textit{3) Prefix languages (Ours)} &
   &
   &
   &
   &
   &
   &
   &
   &
   &
   &
   &
   &
   &
   &
   &
   &
   &
   &
   &
   &
   &
   &
   &
   &
   &
   &
   &
   &
   &
   &
   &
   &
   \\
$P_{1,2}$ &
  78.4 &
  84.3 &
  84.0 &
  100.0 &
  100.0 &
  100.0 &
  54.4 &
  51.7 &
  51.2 &
  50.3 &
  50.3 &
  50.3 &
  58.7 &
  54.2 &
  55.2 &
  97.8 &
  95.1 &
  74.6 &
  56.1 &
  50.6 &
  52.1 &
  65.8 &
  60.2 &
  59.4 &
  100.0 &
  100.0 &
  100.0 &
  100.0 &
  100.0 &
  100.0 &
  100.0 &
  100.0 &
  100.0 \\
$P_{2,2}$ &
  43.4 &
  61.8 &
  91.0 &
  100.0 &
  100.0 &
  100.0 &
  28.1 &
  29.6 &
  8.3 &
  25.2 &
  25.2 &
  25.2 &
  86.5 &
  97.0 &
  95.5 &
  87.1 &
  91.0 &
  81.2 &
  30.6 &
  27.2 &
  30.0 &
  32.2 &
  34.7 &
  35.2 &
  99.3 &
  99.8 &
  99.9 &
  99.8 &
  100.0 &
  100.0 &
  100.0 &
  100.0 &
  100.0 \\
$P_{4,2}$ &
  100.0 &
  100.0 &
  98.2 &
  100.0 &
  100.0 &
  100.0 &
  18.0 &
  18.1 &
  20.8 &
  6.3 &
  6.3 &
  6.3 &
  78.0 &
  96.2 &
  95.6 &
  62.0 &
  67.5 &
  46.0 &
  14.5 &
  11.2 &
  10.6 &
  99.9 &
  99.5 &
  100.0 &
  93.0 &
  89.1 &
  92.4 &
  100.0 &
  100.0 &
  100.0 &
  100.0 &
  100.0 &
  100.0 \\
$P_{1,4}$ &
  78.4 &
  57.0 &
  47.8 &
  100.0 &
  100.0 &
  100.0 &
  28.9 &
  32.0 &
  29.5 &
  25.2 &
  25.2 &
  25.2 &
  38.8 &
  39.0 &
  43.1 &
  86.8 &
  72.9 &
  78.9 &
  28.6 &
  35.7 &
  31.1 &
  38.3 &
  39.3 &
  38.5 &
  100.0 &
  100.0 &
  100.0 &
  100.0 &
  100.0 &
  100.0 &
  100.0 &
  100.0 &
  100.0 \\
$P_{2,4}$ &
  94.1 &
  97.7 &
  98.9 &
  100.0 &
  100.0 &
  100.0 &
  12.2 &
  31.2 &
  21.2 &
  6.3 &
  6.3 &
  6.3 &
  27.9 &
  40.2 &
  24.8 &
  59.9 &
  49.7 &
  45.1 &
  15.5 &
  13.0 &
  14.7 &
  27.6 &
  27.0 &
  35.9 &
  99.5 &
  99.7 &
  99.8 &
  100.0 &
  100.0 &
  100.0 &
  100.0 &
  100.0 &
  100.0 \\
$P_{4,4}$ &
  68.0 &
  90.0 &
  96.8 &
  98.4 &
  94.7 &
  98.8 &
  66.6 &
  65.3 &
  76.2 &
  0.4 &
  0.4 &
  0.4 &
  30.1 &
  17.4 &
  16.5 &
  60.0 &
  59.4 &
  55.1 &
  5.8 &
  11.9 &
  6.2 &
  43.3 &
  40.2 &
  45.2 &
  95.1 &
  95.9 &
  94.8 &
  100.0 &
  100.0 &
  100.0 &
  100.0 &
  100.0 &
  100.0 \\ \bottomrule
\end{tabular}}
\end{table*}

%% file: tables/appendix/listops_assoc_summary.tex

\begin{table*}[t]
\centering
\small
\caption{Summary of ListOps ablation on associativity regularization strength. Left block reports validation accuracy; middle block reports mean accuracy averaged across all reported sequence-length bins, and right block reports long-range accuracy on the longest reported bin (60--80 for (3,9), 101--200 for [3,14], (5,9), and (5,14)). Higher is better. Bins are labeled ($x$, $y$) where $x$ is maximum depth and $y$ is maximum arity. $^\ast$ indicates that the strength was not included in the original hyperparameter sweep and was added to provide a more complete picture of the relationship between associativity strength and performance.}
\label{tab:listops_assoc_summary}
\begin{tabular}{lccccccccc}
\toprule
& \multicolumn{1}{c}{\textbf{Val. Acc.}} & \multicolumn{4}{c}{\textbf{Mean Accuracy}} & \multicolumn{4}{c}{\textbf{Longest Bin}} \\
\cmidrule(l){2-2}\cmidrule(lr){3-6}\cmidrule(l){7-10}
\textbf{Assoc. Strength}
& \textbf{(3, 9)}
& \textbf{(3, 9)}
& \textbf{(3, 14)}
& \textbf{(5, 9)}
& \textbf{(5, 14)}
& \textbf{(3, 9)}
& \textbf{(3, 14)}
& \textbf{(5, 9)}
& \textbf{(5, 14)} \\
\midrule
0.00    & 84.0 & 65.5 & 63.8 & 44.3 & 47.7 & 56.8 & 58.8 & 39.0 & 41.7 \\
0.10    & 83.7 & 65.3 & 63.8 & 44.1 & 47.2 & 57.8 & 59.6 & 39.0 & 42.1 \\
0.25$^\ast$    & 89.0 & 75.7 & 70.3 & 46.8 & 49.6 & 68.1 & 65.0 & 38.6 & 41.0 \\
0.50$^\ast$    & 84.6 & 66.4 & 64.8 & 44.5 & 47.5 & 59.1 & 59.6 & 39.3 & 41.1 \\
1.00    & 88.7 & 76.6 & 71.0 & 47.1 & 50.4 & 69.1 & 65.3 & 38.6 & 41.2 \\
2.00$^\ast$    & 84.2 & 67.1 & 64.9 & 44.6 & 47.9 & 58.5 & 59.1 & 38.6 & 41.6 \\
5.00$^\ast$    & 89.3 & 76.7 & 70.8 & 47.1 & 50.2 & 69.3 & 65.3 & 38.7 & 41.4 \\
\bottomrule
\end{tabular}
\end{table*}

%% file: tables/appendix/listops_assoc.tex
\begin{table*}[t]
\centering
\small
\caption{Full results of ListOps ablation on associativity regularization strength. Each row represents a different associativity regularization strength and entries indicate the corresponding accuracy. Higher is better. $^\ast$ indicates that the strength was not included in the original hyperparameter sweep and was added to provide a more complete picture of the relationship between associativity strength and performance.}
\label{tab:assoc_str_full_ablation}

\begin{tabular}{@{} lcccccc @{}}\toprule
\textbf{Assoc. Str.} &
\multicolumn{6}{c}{\textbf{Sequence length}} \\
\cmidrule(lr){2-7}
& \textbf{5--20} & \textbf{21--40} & \textbf{41--60} & \textbf{61--80} & \textbf{81--100} & \textbf{101--200} \\
\midrule

\multicolumn{7}{@{}l}{\textbf{\emph{Max Depth: 3; Max Arguments: 9}}} \\
\midrule
0.0 & 72.3 & 72.6 & 60.2 & 56.8 & -- & -- \\
0.1 & 71.0 & 71.7 & 60.7 & 57.8  & -- & -- \\
0.25$^\ast$ & 88.3 & 79.6 & 66.8 & 68.1 & -- & -- \\
0.5$^\ast$ & 73.1 & 72.9 & 60.6 & 59.1 & -- & -- \\
1.0 & 88.6 & 81.1 & 67.5 & 69.1 & -- & -- \\
2.0$^\ast$ & 76.6 & 73.0 & 60.4 & 58.5 & -- & -- \\
5.0$^\ast$ & 89.3 & 80.7 & 67.6 & 69.3 & -- & -- \\

\midrule
\multicolumn{7}{@{}l}{\textbf{\emph{Max Depth: 3; Max Arguments: 14}}} \\
\midrule
0.0 & 72.0 & 73.1 & 62.4 & 61.1 & 55.6 & 58.8 \\
0.1 & 71.0 & 71.5 & 62.7 & 61.1 & 56.7 & 59.6 \\
0.25$^\ast$ & 87.1 & 77.7 & 66.7 & 66.8 & 58.6 & 65.0 \\
0.5$^\ast$ & 73.0 & 72.4 & 62.9 & 62.8 & 57.9 & 59.6 \\
1.0 & 88.0 & 78.4 & 66.2 & 68.1 & 59.7 & 65.3 \\
2.0$^\ast$ & 75.6 & 73.2 & 62.5 & 61.6 & 57.3 & 59.1 \\
5.0$^\ast$ & 88.9 & 78.2 & 66.4 & 67.5 & 58.5 & 65.3 \\

\midrule
\multicolumn{7}{@{}l}{\textbf{\emph{Max Depth: 5; Max Arguments: 9}}} \\
\midrule
0.0 & 72.5 & 42.4 & 38.4 & 37.5 & 36.0 & 39.0 \\
0.1 & 71.5 & 42.2 & 38.5 & 37.7 & 35.8 & 39.0 \\
0.25$^\ast$ & 88.5 & 42.3 & 38.2 & 37.1 & 35.9 & 38.6 \\
0.5$^\ast$ & 73.7 & 42.2 & 37.7 & 38.0 & 36.2 & 39.3 \\
1.0 & 89.3 & 42.7 & 37.9 & 37.6 & 36.4 & 38.6 \\
2.0$^\ast$ & 75.4 & 42.4 & 37.9 & 37.4 & 36.0 & 38.6 \\
5.0$^\ast$ & 90.0 & 42.5 & 38.3 & 37.2 & 35.7 & 38.7 \\

\midrule
\multicolumn{7}{@{}l}{\textbf{\emph{Max Depth: 5; Max Arguments: 14}}} \\
\midrule
0.0 & 74.9 & 49.1 & 41.1 & 40.2 & 38.9 & 41.7 \\
0.1 & 73.9 & 48.3 & 40.3 & 39.8 & 38.5 & 42.1 \\
0.25$^\ast$ & 90.3 & 48.6 & 40.2 & 39.6 & 38.1 & 41.0 \\
0.5$^\ast$ & 75.5 & 48.7 & 40.2 & 40.5 & 38.8 & 41.1 \\
1.0 & 91.7 & 49.5 & 40.2 & 40.9 & 38.8 & 41.2 \\
2.0$^\ast$ & 78.1 & 48.4 & 40.0 & 40.7 & 38.5 & 41.6 \\
5.0$^\ast$ & 92.2 & 48.5 & 40.5 & 40.1 & 38.4 & 41.4 \\

\bottomrule
\end{tabular}
\end{table*}

%% file: tables/appendix/listops_100k.tex
\begin{tabular}{@{} lcccccc @{}}\toprule
\textbf{Architecture} &
\multicolumn{6}{c}{\textbf{Sequence length}} \\
\cmidrule(lr){2-7}
& \textbf{5--20} & \textbf{21--40} & \textbf{41--60} & \textbf{61--80} & \textbf{81--100} & \textbf{101--200} \\
\midrule
\multicolumn{7}{@{}l}{\textbf{Max Depth: 3; Max Arguments: 9}} \\
\midrule
RIR-GRC    & 89.2$\pm$0.6 & 65.1$\pm$1.0 & 60.0$\pm$0.3 & 56.8$\pm$0.5 & -- & -- \\
BBT-GRC    & 68.5$\pm$0.4 & 58.7$\pm$0.4 & 55.5$\pm$0.3 & 52.4$\pm$0.6 & -- & -- \\
MLP-LDRU       & 63.6$\pm$1.5 & 56.7$\pm$0.4 & 53.0$\pm$0.5 & 51.1$\pm$0.4 & -- & -- \\
TF (ALiBi) & 65.3$\pm$0.8 & 58.6$\pm$0.7 & 55.1$\pm$0.6 & 51.3$\pm$0.5 & -- & -- \\
LSTM       & 57.8$\pm$1.2 & 55.2$\pm$1.2 & 46.9$\pm$1.0 & 42.4$\pm$0.6 & -- & -- \\
TF (NoPE)  & 40.9$\pm$0.3 & 23.3$\pm$0.3 & 19.1$\pm$0.7 & 18.2$\pm$0.9 & -- & -- \\
TF (Sin.)  & 30.6$\pm$4.0 & 57.4$\pm$0.7 & 15.3$\pm$1.1 & 12.8$\pm$0.9 & -- & -- \\
\midrule
\multicolumn{7}{@{}l}{\textbf{Max Depth: 3; Max Arguments: 14}} \\
\midrule
RIR-GRC    & 86.5$\pm$0.7 & 65.3$\pm$0.5 & 61.3$\pm$0.4 & 60.2$\pm$0.2 & 58.2$\pm$0.3 & 57.5$\pm$0.3 \\
BBT-GRC    & 69.6$\pm$0.5 & 61.7$\pm$0.3 & 59.7$\pm$0.7 & 58.3$\pm$0.3 & 57.5$\pm$0.1 & 56.0$\pm$0.5 \\
MLP-LDRU       & 64.1$\pm$1.5 & 59.9$\pm$0.2 & 57.1$\pm$0.3 & 55.8$\pm$0.6 & 54.2$\pm$0.6 & 54.6$\pm$0.5 \\
TF (ALiBi) & 66.0$\pm$0.7 & 60.3$\pm$0.7 & 57.9$\pm$0.8 & 56.6$\pm$0.8 & 55.0$\pm$1.2 & 51.9$\pm$1.9 \\
LSTM       & 57.7$\pm$1.4 & 57.2$\pm$1.0 & 50.5$\pm$1.0 & 46.8$\pm$1.0 & 43.8$\pm$1.5 & 39.1$\pm$4.1 \\
TF (NoPE)  & 43.1$\pm$0.2 & 26.9$\pm$0.3 & 22.5$\pm$0.3 & 20.7$\pm$0.6 & 19.5$\pm$0.4 & 17.4$\pm$1.9 \\
TF (Sin.)  & 30.8$\pm$4.7 & 58.8$\pm$0.5 & 16.1$\pm$1.4 & 13.5$\pm$1.3 & 11.9$\pm$0.6 & 8.8$\pm$2.7 \\
\midrule
\multicolumn{7}{@{}l}{\textbf{Max Depth: 5; Max Arguments: 9}} \\
\midrule
RIR-GRC    & 90.5$\pm$0.2 & 45.2$\pm$0.6 & 40.2$\pm$0.6 & 39.2$\pm$0.2 & 37.8$\pm$0.4 & 39.4$\pm$0.3 \\
BBT-GRC    & 73.4$\pm$0.4 & 45.7$\pm$0.4 & 41.3$\pm$0.4 & 40.7$\pm$0.4 & 39.3$\pm$0.2 & 40.6$\pm$0.2 \\
MLP-LDRU       & 68.2$\pm$1.9 & 41.4$\pm$0.6 & 38.2$\pm$0.3 & 37.6$\pm$0.3 & 36.5$\pm$0.3 & 38.9$\pm$0.3 \\
TF (ALiBi) & 71.1$\pm$1.0 & 39.2$\pm$1.1 & 34.8$\pm$1.1 & 34.3$\pm$1.1 & 33.1$\pm$1.2 & 33.5$\pm$0.6 \\
LSTM       & 63.0$\pm$1.2 & 40.0$\pm$0.7 & 34.6$\pm$0.6 & 33.8$\pm$0.9 & 31.2$\pm$1.2 & 27.2$\pm$2.2 \\
TF (NoPE)  & 51.9$\pm$0.6 & 19.0$\pm$0.6 & 15.8$\pm$0.8 & 15.0$\pm$1.4 & 14.1$\pm$0.9 & 12.7$\pm$1.0 \\
TF (Sin.)  & 35.4$\pm$5.8 & 43.4$\pm$0.5 & 13.3$\pm$0.7 & 11.7$\pm$0.8 & 11.0$\pm$0.5 & 8.8$\pm$0.9 \\
\midrule
\multicolumn{7}{@{}l}{\textbf{Max Depth: 5; Max Arguments: 14}} \\
\midrule
RIR-GRC    & 93.1$\pm$0.3 & 49.7$\pm$0.4 & 42.1$\pm$0.5 & 41.9$\pm$0.4 & 40.9$\pm$0.4 & 41.9$\pm$0.5 \\
BBT-GRC    & 75.9$\pm$0.3 & 50.4$\pm$0.3 & 44.0$\pm$0.3 & 43.9$\pm$0.3 & 42.5$\pm$0.5 & 43.8$\pm$0.3 \\
MLP-LDRU       & 71.1$\pm$2.4 & 46.3$\pm$0.5 & 40.1$\pm$0.4 & 40.1$\pm$0.3 & 39.3$\pm$0.3 & 41.8$\pm$0.3 \\
TF (ALiBi) & 74.0$\pm$1.2 & 44.4$\pm$0.4 & 36.9$\pm$1.0 & 36.5$\pm$1.7 & 35.5$\pm$1.5 & 36.0$\pm$1.1 \\
LSTM       & 65.3$\pm$1.6 & 43.6$\pm$0.7 & 36.4$\pm$0.6 & 35.9$\pm$0.9 & 34.4$\pm$1.5 & 30.4$\pm$2.2 \\
TF (NoPE)  & 56.8$\pm$0.5 & 21.1$\pm$0.5 & 16.8$\pm$0.5 & 16.3$\pm$0.8 & 15.6$\pm$0.6 & 14.1$\pm$0.6 \\
TF (Sin.)  & 36.4$\pm$7.1 & 46.8$\pm$0.3 & 13.5$\pm$0.8 & 12.2$\pm$0.7 & 11.3$\pm$0.6 & 9.2$\pm$1.8 \\
\bottomrule
\end{tabular}%

%% file: tables/appendix/listops_500k.tex
\begin{tabular}{@{} lcccccc @{}}\toprule
\textbf{Architecture} &
\multicolumn{6}{c}{\textbf{Sequence length}} \\
\cmidrule(lr){2-7}
& \textbf{5--20} & \textbf{21--40} & \textbf{41--60} & \textbf{61--80} & \textbf{81--100} & \textbf{101--200} \\
\midrule

\multicolumn{7}{@{}l}{\textbf{Max Depth: 3; Max Arguments: 9}} \\
\midrule
RIR-GRC    & 98.7$\pm$1.2 & 92.3$\pm$3.3 & 83.6$\pm$2.5 & 74.6$\pm$2.2 & -- & -- \\
BBT-GRC    & 97.2$\pm$0.4 & 89.4$\pm$1.0 & 80.6$\pm$1.2 & 72.7$\pm$0.8 & -- & -- \\
MLP-LDRU       & 80.0$\pm$3.6 & 75.1$\pm$2.3 & 62.5$\pm$2.2 & 61.1$\pm$2.6 & -- & -- \\
TF (ALiBi) & 70.1$\pm$0.9 & 84.5$\pm$3.4 & 61.7$\pm$1.3 & 54.6$\pm$1.3 & -- & -- \\
LSTM       & 71.6$\pm$1.8 & 76.6$\pm$2.7 & 55.9$\pm$0.8 & 48.2$\pm$0.8 & -- & -- \\
TF (NoPE)  & 48.3$\pm$0.2 & 35.3$\pm$0.2 & 31.5$\pm$0.2 & 31.9$\pm$0.3 & -- & -- \\
TF (Sin.)  & 34.1$\pm$8.2 & 90.0$\pm$2.9 & 11.1$\pm$0.6 & 9.7$\pm$0.9 & -- & -- \\

\midrule
\multicolumn{7}{@{}l}{\textbf{Max Depth: 3; Max Arguments: 14}} \\
\midrule
RIR-GRC    & 96.7$\pm$1.3 & 86.4$\pm$2.7 & 77.1$\pm$2.0 & 69.9$\pm$1.6 & 65.1$\pm$2.8 & 62.1$\pm$3.1 \\
BBT-GRC    & 97.2$\pm$0.5 & 88.7$\pm$1.0 & 80.9$\pm$1.6 & 74.8$\pm$1.3 & 71.3$\pm$1.4 & 67.9$\pm$0.8 \\
MLP-LDRU       & 79.6$\pm$4.0 & 74.6$\pm$2.8 & 64.4$\pm$1.6 & 63.4$\pm$1.4 & 57.3$\pm$1.0 & 61.3$\pm$2.3 \\
TF (ALiBi) & 70.1$\pm$0.6 & 80.8$\pm$4.6 & 64.1$\pm$1.1 & 58.8$\pm$1.9 & 55.5$\pm$2.6 & 44.6$\pm$9.6 \\
LSTM       & 71.1$\pm$2.0 & 75.7$\pm$1.8 & 59.3$\pm$0.9 & 53.6$\pm$1.5 & 51.3$\pm$1.9 & 49.2$\pm$2.6 \\
TF (NoPE)  & 50.6$\pm$0.2 & 38.7$\pm$0.2 & 35.1$\pm$0.2 & 34.5$\pm$0.5 & 33.4$\pm$0.8 & 32.1$\pm$1.3 \\
TF (Sin.)  & 35.8$\pm$8.0 & 78.7$\pm$3.1 & 10.9$\pm$1.2 & 9.7$\pm$1.3 & 9.2$\pm$1.7 & 8.2$\pm$1.5 \\

\midrule
\multicolumn{7}{@{}l}{\textbf{Max Depth: 5; Max Arguments: 9}} \\
\midrule
RIR-GRC    & 94.6$\pm$0.9 & 46.8$\pm$1.0 & 39.5$\pm$0.4 & 38.1$\pm$1.0 & 36.3$\pm$1.2 & 36.0$\pm$2.4 \\
BBT-GRC    & 93.7$\pm$0.2 & 46.7$\pm$0.3 & 40.9$\pm$0.3 & 40.2$\pm$0.2 & 38.8$\pm$0.2 & 39.9$\pm$0.5 \\
MLP-LDRU       & 80.9$\pm$4.3 & 42.5$\pm$0.4 & 38.3$\pm$0.5 & 37.6$\pm$0.6 & 35.9$\pm$0.5 & 38.6$\pm$0.4 \\
TF (ALiBi) & 71.5$\pm$0.5 & 41.2$\pm$2.0 & 33.7$\pm$1.0 & 31.4$\pm$1.7 & 29.4$\pm$2.6 & 21.7$\pm$6.5 \\
LSTM       & 75.9$\pm$2.6 & 42.6$\pm$0.4 & 37.2$\pm$0.2 & 37.0$\pm$0.4 & 35.7$\pm$0.5 & 35.3$\pm$1.1 \\
TF (NoPE)  & 59.1$\pm$0.1 & 26.1$\pm$0.2 & 23.4$\pm$0.2 & 23.5$\pm$0.1 & 22.2$\pm$0.1 & 22.0$\pm$0.3 \\
TF (Sin.)  & 46.4$\pm$11.4 & 45.3$\pm$0.8 & 10.7$\pm$0.9 & 10.1$\pm$0.9 & 9.6$\pm$0.8 & 8.6$\pm$0.8 \\

\midrule
\multicolumn{7}{@{}l}{\textbf{Max Depth: 5; Max Arguments: 14}} \\
\midrule
RIR-GRC    & 96.9$\pm$1.0 & 54.3$\pm$1.5 & 41.4$\pm$0.5 & 40.4$\pm$0.7 & 39.1$\pm$1.4 & 36.8$\pm$2.9 \\
BBT-GRC    & 96.9$\pm$0.3 & 54.1$\pm$0.7 & 43.5$\pm$0.4 & 42.5$\pm$0.2 & 41.8$\pm$0.3 & 42.6$\pm$1.1 \\
MLP-LDRU       & 83.6$\pm$4.7 & 48.9$\pm$0.6 & 40.4$\pm$0.5 & 40.3$\pm$0.5 & 38.7$\pm$0.7 & 41.6$\pm$0.3 \\
TF (ALiBi) & 72.9$\pm$0.8 & 48.8$\pm$2.3 & 34.0$\pm$1.8 & 32.4$\pm$2.8 & 31.1$\pm$3.3 & 23.6$\pm$6.5 \\
LSTM       & 78.0$\pm$2.8 & 49.2$\pm$0.7 & 38.9$\pm$0.4 & 38.8$\pm$0.3 & 39.0$\pm$0.4 & 38.3$\pm$1.3 \\
TF (NoPE)  & 64.2$\pm$0.3 & 30.3$\pm$0.3 & 24.3$\pm$0.2 & 24.6$\pm$0.2 & 24.2$\pm$0.4 & 23.9$\pm$0.6 \\
TF (Sin.)  & 49.1$\pm$12.9 & 51.8$\pm$1.7 & 10.3$\pm$0.6 & 9.8$\pm$1.3 & 9.8$\pm$1.3 & 8.5$\pm$0.7 \\

\bottomrule
\end{tabular}

%% file: tables/appendix/listops_1m.tex
\begin{tabular}{@{} lcccccc @{}}\toprule
\textbf{Architecture} &
\multicolumn{6}{c}{\textbf{Sequence length}} \\
\cmidrule(lr){2-7}
& \textbf{5--20} & \textbf{21--40} & \textbf{41--60} & \textbf{61--80} & \textbf{81--100} & \textbf{101--200} \\
\midrule
\multicolumn{7}{@{}l}{\textbf{Max Depth: 3; Max Arguments: 9}} \\
\midrule
RIR-GRC    & 99.2$\pm$0.2 & 94.3$\pm$0.8 & 86.9$\pm$1.4 & 79.2$\pm$3.0 & -- & -- \\
BBT-GRC    & 96.0$\pm$2.0 & 88.0$\pm$3.5 & 80.0$\pm$3.1 & 72.5$\pm$2.5 & -- & -- \\
MLP-LDRU       & 82.9$\pm$9.6 & 81.9$\pm$2.0 & 66.1$\pm$1.5 & 67.8$\pm$3.6 & -- & -- \\
TF (ALiBi) & 70.5$\pm$1.1 & 85.1$\pm$4.8 & 62.2$\pm$0.6 & 55.1$\pm$0.8 & -- & -- \\
LSTM       & 74.4$\pm$2.4 & 79.3$\pm$2.9 & 59.8$\pm$0.8 & 50.0$\pm$1.1 & -- & -- \\
TF (NoPE)  & 49.5$\pm$0.3 & 35.6$\pm$0.3 & 32.1$\pm$0.4 & 32.1$\pm$0.4 & -- & -- \\
TF (Sin.)  & 33.3$\pm$7.7 & 89.6$\pm$5.3 & 10.8$\pm$0.5 & 9.8$\pm$0.6 & -- & -- \\
\midrule
\multicolumn{7}{@{}l}{\textbf{Max Depth: 3; Max Arguments: 14}} \\
\midrule
RIR-GRC    & 97.5$\pm$0.8 & 88.5$\pm$1.8 & 79.6$\pm$2.1 & 73.7$\pm$1.8 & 69.1$\pm$2.0 & 66.4$\pm$2.3 \\
BBT-GRC    & 96.0$\pm$1.8 & 86.9$\pm$2.1 & 79.2$\pm$1.4 & 74.0$\pm$1.1 & 70.6$\pm$0.6 & 67.7$\pm$0.9 \\
MLP-LDRU       & 82.2$\pm$9.2 & 79.4$\pm$1.2 & 66.1$\pm$0.7 & 66.8$\pm$1.5 & 58.6$\pm$0.5 & 65.2$\pm$1.4 \\
TF (ALiBi) & 70.2$\pm$0.9 & 81.4$\pm$4.9 & 65.1$\pm$0.8 & 59.6$\pm$2.0 & 56.8$\pm$1.5 & 46.4$\pm$5.1 \\
LSTM       & 74.4$\pm$3.1 & 78.7$\pm$1.5 & 62.7$\pm$1.7 & 56.5$\pm$5.0 & 52.6$\pm$4.1 & 51.4$\pm$6.2 \\
TF (NoPE)  & 51.6$\pm$0.3 & 39.3$\pm$0.3 & 35.5$\pm$0.3 & 34.4$\pm$0.3 & 33.3$\pm$0.4 & 31.2$\pm$0.9 \\
TF (Sin.)  & 34.6$\pm$7.7 & 80.0$\pm$2.8 & 10.8$\pm$1.0 & 9.8$\pm$1.8 & 9.4$\pm$1.6 & 8.4$\pm$1.6 \\
\midrule
\multicolumn{7}{@{}l}{\textbf{Max Depth: 5; Max Arguments: 9}} \\
\midrule
RIR-GRC    & 94.8$\pm$0.2 & 45.5$\pm$2.3 & 38.6$\pm$1.7 & 37.3$\pm$1.9 & 35.8$\pm$2.0 & 36.4$\pm$2.8 \\
BBT-GRC    & 93.1$\pm$0.8 & 46.5$\pm$0.7 & 40.8$\pm$0.5 & 39.7$\pm$0.3 & 38.5$\pm$0.2 & 39.2$\pm$0.6 \\
MLP-LDRU       & 82.6$\pm$10.2 & 42.8$\pm$0.2 & 38.0$\pm$0.5 & 37.3$\pm$0.4 & 35.9$\pm$0.3 & 38.8$\pm$0.3 \\
TF (ALiBi) & 71.4$\pm$0.9 & 41.4$\pm$2.0 & 34.3$\pm$1.5 & 33.0$\pm$1.4 & 31.2$\pm$1.9 & 22.8$\pm$5.8 \\
LSTM       & 78.5$\pm$3.2 & 43.8$\pm$0.4 & 38.1$\pm$0.4 & 37.2$\pm$0.6 & 35.8$\pm$0.7 & 36.0$\pm$0.5 \\
TF (NoPE)  & 60.4$\pm$0.3 & 26.3$\pm$0.1 & 23.7$\pm$0.1 & 23.6$\pm$0.1 & 22.7$\pm$0.1 & 22.7$\pm$0.4 \\
TF (Sin.)  & 44.9$\pm$9.7 & 45.2$\pm$1.0 & 11.1$\pm$0.6 & 9.7$\pm$1.0 & 9.5$\pm$0.9 & 8.7$\pm$0.6 \\
\midrule
\multicolumn{7}{@{}l}{\textbf{Max Depth: 5; Max Arguments: 14}} \\
\midrule
RIR-GRC    & 97.3$\pm$0.2 & 53.1$\pm$1.8 & 40.8$\pm$1.4 & 40.0$\pm$1.7 & 39.4$\pm$1.6 & 38.8$\pm$2.3 \\
BBT-GRC    & 96.5$\pm$0.6 & 53.7$\pm$0.4 & 43.1$\pm$0.3 & 42.4$\pm$0.3 & 41.2$\pm$0.4 & 41.4$\pm$1.2 \\
MLP-LDRU       & 84.8$\pm$10.5 & 49.3$\pm$0.3 & 40.0$\pm$0.4 & 40.1$\pm$0.6 & 38.4$\pm$0.5 & 41.5$\pm$0.6 \\
TF (ALiBi) & 73.1$\pm$1.1 & 49.3$\pm$2.3 & 35.2$\pm$1.7 & 33.9$\pm$2.2 & 32.7$\pm$1.9 & 25.8$\pm$6.4 \\
LSTM       & 81.1$\pm$3.5 & 50.8$\pm$0.3 & 39.9$\pm$0.5 & 39.0$\pm$0.6 & 38.7$\pm$0.7 & 38.9$\pm$0.6 \\
TF (NoPE)  & 65.5$\pm$0.6 & 30.8$\pm$0.1 & 24.8$\pm$0.1 & 24.9$\pm$0.1 & 24.7$\pm$0.3 & 24.2$\pm$0.5 \\
TF (Sin.)  & 46.9$\pm$10.7 & 51.8$\pm$2.2 & 10.5$\pm$0.6 & 10.0$\pm$1.2 & 9.8$\pm$1.3 & 8.7$\pm$0.9 \\
\bottomrule
\end{tabular}

%% file: references.bib
@inproceedings{reddi2018on,
title={{On the Convergence of Adam and Beyond}},
author={Sashank J. Reddi and Satyen Kale and Sanjiv Kumar},
booktitle={International Conference on Learning Representations},
year={2018},
url={https://openreview.net/forum?id=ryQu7f-RZ},
}

@inproceedings{kingma2015adam,
  author       = {Diederik P. Kingma and
                  Jimmy Ba},
  title        = {{Adam: {A} Method for Stochastic Optimization}},
  booktitle    = {3rd International Conference on Learning Representations, {ICLR} 2015,
                  San Diego, CA, USA, May 7--9, 2015, Conference Track Proceedings},
  year         = {2015},
}

@inproceedings{butoi2025training,
title={{Training Neural Networks as Recognizers of Formal Languages}},
author={Alexandra Butoi and Ghazal Khalighinejad and Anej Svete and Josef Valvoda and Ryan Cotterell and Brian DuSell},
booktitle={The Thirteenth International Conference on Learning Representations},
year={2025},
url={https://openreview.net/forum?id=aWLQTbfFgV}
}

@inproceedings{ong2022learnable,
title={{Learnable Commutative Monoids for Graph Neural Networks}},
author={Euan Ong and Petar Veli{\v{c}}kovi{\'c}},
booktitle={The First Learning on Graphs Conference},
year={2022},
url={https://openreview.net/forum?id=WtFobB28VDey}
}

@inproceedings{Chi2023,
 author = {Chi, Ta-Chung and Fan, Ting-Han and Rudnicky, Alexander and Ramadge, Peter},
 title = {{Transformer Working Memory Enables Regular Language Reasoning and Natural Language Length Extrapolation}},
 booktitle = {Findings of the Association for Computational Linguistics: EMNLP 2023},
 year = {2023},
 publisher = {Association for Computational Linguistics}
}

@inproceedings{soulos2024recurrent,
title={{Recurrent Transformers Trade-off Parallelism for Length Generalization on Regular Languages}},
author={Paul Soulos and Aleksandar Terzic and Michael Hersche and Abbas Rahimi},
booktitle={The First Workshop on System-2 Reasoning at Scale, NeurIPS'24},
year={2024},
url={https://openreview.net/forum?id=6PjZA4Jvge}
}

@inproceedings{chung2014empirical,
title = {{Empirical Evaluation of Gated Recurrent Neural Networks on Sequence Modeling}},
author = "Junyoung Chung and Caglar Gulcehre and Kyunghyun Cho and Yoshua Bengio",
year = "2014",
booktitle = " NIPS Deep Learning and Representation Learning Workshop",
}

@inproceedings{yang2025gated,
title={Gated Delta Networks: Improving Mamba2 with Delta Rule},
author={Songlin Yang and Jan Kautz and Ali Hatamizadeh},
booktitle={The Thirteenth International Conference on Learning Representations},
year={2025},
url={https://openreview.net/forum?id=r8H7xhYPwz}
}

@InProceedings{jozefowicz2015an,
  title = 	 {{An Empirical Exploration of Recurrent Network Architectures}},
  author = 	 {Jozefowicz, Rafal and Zaremba, Wojciech and Sutskever, Ilya},
  booktitle = 	 {Proceedings of the 32nd International Conference on Machine Learning},
  year = 	 {2015},
  publisher =    {PMLR},
  url = 	 {https://proceedings.mlr.press/v37/jozefowicz15.html},
}

@inproceedings{terzic2025sdssm,
      title={{On the Expressiveness and Length Generalization of Selective State-Space Models on Regular Languages}}, 
      author={Aleksandar Terzić and Michael Hersche and Giacomo Camposampiero and Thomas Hofmann and Abu Sebastian and Abbas Rahimi},
      year={2025},
      booktitle = {Proceedings of the AAAI Conference on Artificial Intelligence}
}

@inproceedings{terzic_2025_pdssm,
 	author = {Terzić, Aleksandar and Menet, Nicolas and Hersche, Michael and Hofmann, Thomas and Rahimi, Abbas},
 	booktitle = {Advances in Neural Information Processing Systems (NeurIPS)},
 	title = {{Structured Sparse Transition Matrices to Enable State Tracking in State-Space Models}},
 	year = {2025}
}

@inproceedings{walker2025structured,
title={{Structured Linear {CDE}s: Maximally Expressive and Parallel-in-Time Sequence Models}},
author={Benjamin Walker and Lingyi Yang and Nicola Muca Cirone and Cristopher Salvi and Terry Lyons},
booktitle={The Thirty-ninth Annual Conference on Neural Information Processing Systems},
year={2025},
url={https://openreview.net/forum?id=HKDyRDzy1E}
}

@inproceedings{
grazzi2025unlocking,
title={{Unlocking State-Tracking in Linear {RNN}s Through Negative Eigenvalues}},
author={Riccardo Grazzi and Julien Siems and J{\"o}rg K.H. Franke and Arber Zela and Frank Hutter and Massimiliano Pontil},
booktitle={The Thirteenth International Conference on Learning Representations},
year={2025},
url={https://openreview.net/forum?id=UvTo3tVBk2}
}

@inproceedings{nangia2018listops,
  title={{ListOps: A Diagnostic Dataset for Latent Tree Learning}},
  author={Nangia, Nikita and Bowman, Samuel},
  booktitle={Proceedings of the 2018 Conference of the North American Chapter of the Association for Computational Linguistics: Student Research Workshop},
  pages={92--99},
  year={2018}
}

@inproceedings{munkhdalai-yu-2017-neural,
    title = {{Neural Tree Indexers for Text Understanding}},
    author = {Munkhdalai, Tsendsuren  and
      Yu, Hong},
    booktitle = "Proceedings of the 15th Conference of the {E}uropean Chapter of the Association for Computational Linguistics: Volume 1, Long Papers",
    year = "2017",
    publisher = "Association for Computational Linguistics",
    url = "https://aclanthology.org/E17-1002/",
    pages = "11--21",
}

@inproceedings{yu-liu-2018-sliced,
    title = {{Sliced Recurrent Neural Networks}},
    author = {Yu, Zeping  and
      Liu, Gongshen},
    booktitle = "Proceedings of the 27th International Conference on Computational Linguistics",
    year = "2018",
    publisher = "Association for Computational Linguistics",
    url = "https://aclanthology.org/C18-1250/",
    pages = "2953--2964",
}

@inproceedings{shi-etal-2018-tree,
    title = {{On Tree-Based Neural Sentence Modeling}},
    author = {Shi, Haoyue  and
      Zhou, Hao  and
      Chen, Jiaze  and
      Li, Lei},
    booktitle = "Proceedings of the 2018 Conference on Empirical Methods in Natural Language Processing",
    year = "2018",
    publisher = "Association for Computational Linguistics",
    url = "https://aclanthology.org/D18-1492/",
    doi = "10.18653/v1/D18-1492",
    pages = "4631--4641",
}

@inproceedings{
chowdhury2023recursion,
title={{Recursion in Recursion: Two-Level Nested Recursion for Length Generalization with Scalability}},
author={Jishnu Ray Chowdhury and Cornelia Caragea},
booktitle={Thirty-seventh Conference on Neural Information Processing Systems},
year={2023},
url={https://openreview.net/forum?id=o6yTKfdnbA}
}

@inproceedings{socher-etal-2013-recursive,
    title = {{Recursive Deep Models for Semantic Compositionality Over a Sentiment Treebank}},
    author = {Socher, Richard  and
      Perelygin, Alex  and
      Wu, Jean  and
      Chuang, Jason  and
      Manning, Christopher D.  and
      Ng, Andrew  and
      Potts, Christopher},
    booktitle = "Proceedings of the 2013 Conference on Empirical Methods in Natural Language Processing",
    year = "2013",
    publisher = "Association for Computational Linguistics",
    url = "https://aclanthology.org/D13-1170/",
    pages = "1631--1642"
}

@inproceedings{tai-etal-2015-improved,
    title = {{Improved Semantic Representations From Tree-Structured Long Short-Term Memory Networks}},
    author = {Tai, Kai Sheng  and
      Socher, Richard  and
      Manning, Christopher D.},
    booktitle = "Proceedings of the 53rd Annual Meeting of the Association for Computational Linguistics and the 7th International Joint Conference on Natural Language Processing (Volume 1: Long Papers)",
    year = "2015",
    publisher = "Association for Computational Linguistics",
    url = "https://aclanthology.org/P15-1150/",
    doi = "10.3115/v1/P15-1150",
    pages = "1556--1566"
}

@inproceedings{NEURIPS2019_d8e1344e,
  title = {{Ordered Memory}},
  author = {Shen, Yikang and Tan, Shawn and Hosseini, Arian and Lin, Zhouhan and Sordoni, Alessandro and Courville, Aaron C},
  booktitle = {Advances in Neural Information Processing Systems 32},
  pages = {5038--5049},
  year = {2019},
  url = {http://papers.nips.cc/paper/8748-ordered-memory.pdf}
}

@inproceedings{liu2023transformers,
title={{Transformers Learn Shortcuts to Automata}},
author={Bingbin Liu and Jordan T. Ash and Surbhi Goel and Akshay Krishnamurthy and Cyril Zhang},
booktitle={The Eleventh International Conference on Learning Representations },
year={2023},
url={https://openreview.net/forum?id=De4FYqjFueZ}
}

@inproceedings{fan2025looped,
title={{Looped Transformers for Length Generalization}},
author={Ying Fan and Yilun Du and Kannan Ramchandran and Kangwook Lee},
booktitle={The Thirteenth International Conference on Learning Representations},
year={2025},
url={https://openreview.net/forum?id=2edigk8yoU}
}

@article{ladner1980parallel,
author = {Ladner, Richard E. and Fischer, Michael J.},
title = {{Parallel Prefix Computation}},
year = {1980},
publisher = {Association for Computing Machinery},
volume = {27},
number = {4},
doi = {10.1145/322217.322232},
journal = {J. ACM},
pages = {831--838},
}

@article{Hahn2020,
 author = {Hahn, Michael},
 title = {{Theoretical Limitations of Self-Attention in Neural Sequence Models}},
 journal = {Transactions of the Association for Computational Linguistics},
 volume = {8},
 pages = {156--171},
 year = {2020}
}

@inproceedings{huang2025a,
title={{A Formal Framework for Understanding Length Generalization in Transformers}},
author={Xinting Huang and Andy Yang and Satwik Bhattamishra and Yash Sarrof and Andreas Krebs and Hattie Zhou and Preetum Nakkiran and Michael Hahn},
booktitle={The Thirteenth International Conference on Learning Representations},
year={2025},
url={https://openreview.net/forum?id=U49N5V51rU}
}

@inproceedings{huang2025how,
title={{How Transformers Learn Regular Language Recognition: A Theoretical Study on Training Dynamics and Implicit Bias}},
author={Ruiquan Huang and Yingbin Liang and Jing Yang},
booktitle={Forty-second International Conference on Machine Learning},
year={2025},
url={https://openreview.net/forum?id=yTAR011mOF}
}

@inproceedings{Merrill2020,
 author = {Merrill, William and Weiss, Gail and Goldberg, Yoav and Schwartz, Roy and Smith, Noah A. and Yahav, Eran},
 title = {{A Formal Hierarchy of RNN Architectures}},
 booktitle = {Proceedings of the 58th Annual Meeting of the Association for Computational Linguistics},
 year = {2020},
 pages = {443--459},
 publisher = {Association for Computational Linguistics}
}

@software{jax2018github,
  author = {James Bradbury and Roy Frostig and Peter Hawkins and Matthew James Johnson and Chris Leary and Dougal Maclaurin and George Necula and Adam Paszke and Jake Vander{P}las and Skye Wanderman-{M}ilne and Qiao Zhang},
  title = {{JAX}: composable transformations of {P}ython+{N}um{P}y programs},
  url = {http://github.com/jax-ml/jax},
  version = {0.4.38},
  year = {2018},
}

@software{haiku2020github,
  author = {Tom Hennigan and Trevor Cai and Tamara Norman and Lena Martens and Igor Babuschkin},
  title = {{H}aiku: {S}onnet for {JAX}},
  url = {http://github.com/deepmind/dm-haiku},
  version = {0.0.13},
  year = {2020},
}

@software{deepmind2020jax,
  title = {The {D}eep{M}ind {JAX} {E}cosystem},
  author = {DeepMind and Babuschkin, Igor and Baumli, Kate and Bell, Alison and Bhupatiraju, Surya and Bruce, Jake and Buchlovsky, Peter and Budden, David and Cai, Trevor and Clark, Aidan and Danihelka, Ivo and Dedieu, Antoine and Fantacci, Claudio and Godwin, Jonathan and Jones, Chris and Hemsley, Ross and Hennigan, Tom and Hessel, Matteo and Hou, Shaobo and Kapturowski, Steven and Keck, Thomas and Kemaev, Iurii and King, Michael and Kunesch, Markus and Martens, Lena and Merzic, Hamza and Mikulik, Vladimir and Norman, Tamara and Papamakarios, George and Quan, John and Ring, Roman and Ruiz, Francisco and Sanchez, Alvaro and Sartran, Laurent and Schneider, Rosalia and Sezener, Eren and Spencer, Stephen and Srinivasan, Srivatsan and Stanojevi\'{c}, Milo\v{s} and Stokowiec, Wojciech and Wang, Luyu and Zhou, Guangyao and Viola, Fabio},
  url = {http://github.com/google-deepmind},
  year = {2020},
}

@InProceedings{glorot2010understanding,
  title = 	 {{Understanding the difficulty of training deep feedforward neural networks}},
  author = 	 {Glorot, Xavier and Bengio, Yoshua},
  booktitle = 	 {Proceedings of the Thirteenth International Conference on Artificial Intelligence and Statistics},
  pages = 	 {249--256},
  year = 	 {2010},
  volume = 	 {9},
  series = 	 {Proceedings of Machine Learning Research},
  publisher =    {PMLR},
  url = 	 {https://proceedings.mlr.press/v9/glorot10a.html},
}

@article{Strobl2024,
 author = {Strobl, Lena and Merrill, William and Weiss, Gail and Chiang, David and Angluin, Dana},
 title = {{What Formal Languages Can Transformers Express? A Survey}},
 journal = {Transactions of the Association for Computational Linguistics},
 volume = {12},
 pages = {543--561},
 year = {2024}
}

@inproceedings{chiang2022overcoming,
    title = {{Overcoming a Theoretical Limitation of Self-Attention}},
    author = {Chiang, David  and
      Cholak, Peter},
    booktitle = {Proceedings of the 60th Annual Meeting of the Association for Computational Linguistics (Volume 1: Long Papers)},
    year = {2022},
    publisher = {Association for Computational Linguistics},
    url = {https://aclanthology.org/2022.acl-long.527/},
    doi = {10.18653/v1/2022.acl-long.527},
}

@article{holzer2004deterministic,
title = {{On Deterministic Finite Automata and Syntactic Monoid Size}},
journal = {Theoretical Computer Science},
year = {2004},
doi = {https://doi.org/10.1016/j.tcs.2004.04.010},
author = {Markus Holzer and Barbara König},
}

@book{sakarovitch2009elements,
  title={{Elements of Automata Theory}},
  author={Sakarovitch, Jacques},
  year={2009},
  publisher={Cambridge University Press}
}

@inproceedings{vaswani2017attention,
 author = {Vaswani, Ashish and Shazeer, Noam and Parmar, Niki and Uszkoreit, Jakob and Jones, Llion and Gomez, Aidan N and Kaiser, {\L}ukasz and Polosukhin, Illia},
 booktitle = {Advances in Neural Information Processing Systems},
 title = {{Attention is All you Need}},
 year = {2017}
}

@inproceedings{Bhattamishra2020OnTA,
  title={{On the Ability and Limitations of Transformers to Recognize Formal Languages}},
  author={S. Bhattamishra and Kabir Ahuja and Navin Goyal},
  booktitle={Conference on Empirical Methods in Natural Language Processing},
  year={2020},
  url={https://api.semanticscholar.org/CorpusID:222225236}
}

@article{arnold1980uniform,
author = {Arnold, D. B. and Sleep, M. R.},
title = {{Uniform Random Generation of Balanced Parenthesis Strings}},
year = {1980},
publisher = {Association for Computing Machinery},
volume = {2},
issn = {0164-0925},
url = {https://doi.org/10.1145/357084.357091},
doi = {10.1145/357084.357091},
journal = {ACM Trans. Program. Lang. Syst.},
pages = {122--128},
numpages = {7}
}

@inproceedings{deletang2023neural,
  author       = {Gr{\'{e}}goire Del{\'{e}}tang and
                  Anian Ruoss and
                  Jordi Grau{-}Moya and
                  Tim Genewein and
                  Li Kevin Wenliang and
                  Elliot Catt and
                  Chris Cundy and
                  Marcus Hutter and
                  Shane Legg and
                  Joel Veness and
                  Pedro A. Ortega},
  title        = {{Neural Networks and the Chomsky Hierarchy}},
  booktitle    = {11th International Conference on Learning Representations},
  year         = {2023},
}

@ARTICLE{bengio1994learning,
  author={Bengio, Y. and Simard, P. and Frasconi, P.},
  journal={IEEE Transactions on Neural Networks}, 
  title={{Learning Long-Term Dependencies with Gradient Descent is Difficult}}, 
  year={1994},
  volume={5},
  doi={10.1109/72.279181}
}

@INPROCEEDINGS{bengio1993the,
  author={Bengio, Y. and Frasconi, P. and Simard, P.},
  booktitle={IEEE International Conference on Neural Networks}, 
  title={{The Problem of Learning Long-Term Dependencies in Recurrent Networks}}, 
  year={1993},
  doi={10.1109/ICNN.1993.298725}
}

@article{elfwing2018sigmoid,
  title={{Sigmoid-Weighted Linear Units for Neural Network Function Approximation in Reinforcement Learning}},
  author={Elfwing, Stefan and Uchibe, Eiji and Doya, Kenji},
  journal={Neural Networks},
  volume={107},
  pages={3--11},
  year={2018}
}

@article{vandermaaten2008visualizing,
  author  = {Laurens van der Maaten and Geoffrey Hinton},
  title   = {{Visualizing Data using t-SNE}},
  journal = {Journal of Machine Learning Research},
  year    = {2008},
  volume  = {9},
  number  = {86},
  pages   = {2579--2605}
}

@article{hubert1985comparing,
	title = {{Comparing Partitions}},
	volume = {2},
	issn = {1432-1343},
	doi = {10.1007/BF01908075},
	pages = {193--218},
	journal = {Journal of Classification},
	author = {Hubert, Lawrence and Arabie, Phipps},
  year={1985}
}

@article{gu2023mamba,
  title={{Mamba: Linear-Time Sequence Modeling with Selective State Spaces}},
  author={Gu, Albert and Dao, Tri},
  journal={arXiv preprint arXiv:2312.00752},
  year={2023}
}

@inproceedings{gu2022efficiently,
title={{Efficiently Modeling Long Sequences with Structured State Spaces}},
author={Albert Gu and Karan Goel and Christopher Re},
booktitle={International Conference on Learning Representations},
year={2022},
url={https://openreview.net/forum?id=uYLFoz1vlAC}
}

@article{rousseeuw1987silhouettes,
	title = {{Silhouettes: a graphical aid to the interpretation and validation of cluster analysis}},
	volume = {20},
	issn = {0377-0427},
	doi = {https://doi.org/10.1016/0377-0427(87)90125-7},
	pages = {53--65},
	journal = {Journal of Computational and Applied Mathematics},
	author = {Rousseeuw, Peter J.},
	year={1987},
}

@inproceedings{fan2024advancing,
    title = {{Advancing Regular Language Reasoning in Linear Recurrent Neural Networks}},
    author = "Fan, Ting-Han  and
      Chi, Ta-Chung  and
      Rudnicky, Alexander",
    booktitle = "Proceedings of the 2024 Conference of the North American Chapter of the Association for Computational Linguistics: Human Language Technologies (Volume 2: Short Papers)",
    year = "2024",
    publisher = "Association for Computational Linguistics",
    url = "https://aclanthology.org/2024.naacl-short.4/",
    doi = "10.18653/v1/2024.naacl-short.4",
    pages = "45--53",
}

@inproceedings{smith2023simplified,
title={{Simplified State Space Layers for Sequence Modeling}},
author={Jimmy T.H. Smith and Andrew Warrington and Scott Linderman},
booktitle={The Eleventh International Conference on Learning Representations },
year={2023},
url={https://openreview.net/forum?id=Ai8Hw3AXqks}
}

@inproceedings{lu2023structured,
title={{Structured State Space Models for In-Context Reinforcement Learning}},
author={Chris Lu and Yannick Schroecker and Albert Gu and Emilio Parisotto and Jakob Nicolaus Foerster and Satinder Singh and Feryal Behbahani},
booktitle={Thirty-seventh Conference on Neural Information Processing Systems},
year={2023},
url={https://openreview.net/forum?id=4W9FVg1j6I}
}

@inproceedings{Martin2018,
  title={{Parallelizing Linear Recurrent Neural Nets Over Sequence Length}},
  author={Martin, Eric and Cundy, Chris},
  booktitle={International Conference on Learning Representations},
  year={2018}
}

@inproceedings{Wang2020,
  title={{BPPSA: Scaling Back-propagation by Parallel Scan Algorithm}},
  author={Wang, Shang and Bai, Yifan and Pekhimenko, Gennady},
  booktitle={Proceedings of Machine Learning and Systems},
  volume={2},
  pages={451--469},
  year={2020}
}

@techreport{BlellochTR90,
	author = {Guy~E. Blelloch},
	title = {Prefix Sums and Their Applications},
	institution = {School of Computer Science, Carnegie Mellon University},
	year = {1990} 
}

@article{data_parallel_algorithms,
author = {Hillis, W. Daniel and Steele, Guy L.},
title = {Data Parallel Algorithms},
year = {1986},
url = {https://doi.org/10.1145/7902.7903},
journal = {Commun. ACM},
}

@article{elman1990finding,
author = {Elman, Jeffrey L.},
title = {{Finding Structure in Time}},
journal = {Cognitive Science},
volume = {14},
number = {2},
pages = {179--211},
year = {1990}
}

@article{hochreiter1997long,
author = {Hochreiter, Sepp and Schmidhuber, J\"{u}rgen},
title = {{Long Short-Term Memory}},
year = {1997},
volume = {9},
number = {8},
issn = {0899-7667},
url = {https://doi.org/10.1162/neco.1997.9.8.1735},
doi = {10.1162/neco.1997.9.8.1735},
journal = {Neural Comput.},
}

@inproceedings{sarrof2024the,
title={{The Expressive Capacity of State Space Models: A Formal Language Perspective}},
author={Yash Sarrof and Yana Veitsman and Michael Hahn},
booktitle={The Thirty-eighth Annual Conference on Neural Information Processing Systems},
year={2024},
url={https://openreview.net/forum?id=eV5YIrJPdy}
}

@inproceedings{hahn2024sensitive,
  title={{Why are Sensitive Functions Hard for Transformers?}}, 
  author={Michael Hahn and Mark Rofin},
  year={2024},
	booktitle={Proceedings of the 2024 Annual Conference of the Association for Computational Linguistics (ACL 2024)},
	url={https://arxiv.org/abs/2402.09963}
}

@inproceedings{zhou2024what,
title={{What Algorithms can Transformers Learn? A Study in Length Generalization}},
author={Hattie Zhou and Arwen Bradley and Etai Littwin and Noam Razin and Omid Saremi and Joshua M. Susskind and Samy Bengio and Preetum Nakkiran},
booktitle={The Twelfth International Conference on Learning Representations},
year={2024},
url={https://openreview.net/forum?id=AssIuHnmHX}
}

@inproceedings{ruoss2023randomized,
  author       = {Anian Ruoss and
                  Gr{\'{e}}goire Del{\'{e}}tang and
                  Tim Genewein and
                  Jordi Grau{-}Moya and
                  R{\'{o}}bert Csord{\'{a}}s and
                  Mehdi Bennani and
                  Shane Legg and
                  Joel Veness},
  title        = {{Randomized Positional Encodings Boost Length Generalization of Transformers}},
  booktitle    = {61st Annual Meeting of the Association for Computational Linguistics},
  year         = {2023},
}

@inproceedings{kazemnejad2023the,
title={{The Impact of Positional Encoding on Length Generalization in Transformers}},
author={Amirhossein Kazemnejad and Inkit Padhi and Karthikeyan Natesan and Payel Das and Siva Reddy},
booktitle={Thirty-seventh Conference on Neural Information Processing Systems},
year={2023},
url={https://openreview.net/forum?id=Drrl2gcjzl}
}

@inproceedings{zhou2024transformers,
title={{Transformers Can Achieve Length Generalization But Not Robustly}},
author={Yongchao Zhou and Uri Alon and Xinyun Chen and Xuezhi Wang and Rishabh Agarwal and Denny Zhou},
booktitle={ICLR 2024 Workshop on Mathematical and Empirical Understanding of Foundation Models},
year={2024},
url={https://openreview.net/forum?id=DWkWIh3vFJ}
}

@inproceedings{
cho2025arithmetic,
title={{Arithmetic Transformers Can Length-Generalize in Both Operand Length and Count}},
author={Hanseul Cho and Jaeyoung Cha and Srinadh Bhojanapalli and Chulhee Yun},
booktitle={The Thirteenth International Conference on Learning Representations},
year={2025},
url={https://openreview.net/forum?id=eIgGesYKLG}
}

@inproceedings{
cho2024position,
title={{Position Coupling: Improving Length Generalization of Arithmetic Transformers Using Task Structure}},
author={Hanseul Cho and Jaeyoung Cha and Pranjal Awasthi and Srinadh Bhojanapalli and Anupam Gupta and Chulhee Yun},
booktitle={The Thirty-eighth Annual Conference on Neural Information Processing Systems},
year={2024},
url={https://openreview.net/forum?id=5cIRdGM1uG}
}

@inproceedings{
mcleish2024transformers,
title={{Transformers Can Do Arithmetic with the Right Embeddings}},
author={Sean Michael McLeish and Arpit Bansal and Alex Stein and Neel Jain and John Kirchenbauer and Brian R. Bartoldson and Bhavya Kailkhura and Abhinav Bhatele and Jonas Geiping and Avi Schwarzschild and Tom Goldstein},
booktitle={The 4th Workshop on Mathematical Reasoning and AI at NeurIPS'24},
year={2024},
url={https://openreview.net/forum?id=cBFsFt1nDW}
}

@inproceedings{anil2022exploring,
author = {Anil, Cem and Wu, Yuhuai and Andreassen, Anders and Lewkowycz, Aitor and Misra, Vedant and Ramasesh, Vinay and Slone, Ambrose and Gur-Ari, Guy and Dyer, Ethan and Neyshabur, Behnam},
title = {{Exploring Length Generalization in Large Language Models}},
year = {2022},
booktitle = {Proceedings of the 36th International Conference on Neural Information Processing Systems},
series = {NIPS '22}
}

@inproceedings{press2022train,
title={{Train Short, Test Long: Attention with Linear Biases Enables Input Length Extrapolation}},
author={Ofir Press and Noah Smith and Mike Lewis},
booktitle={International Conference on Learning Representations},
year={2022},
url={https://openreview.net/forum?id=R8sQPpGCv0}
}

@inproceedings{
wang2018glue,
title={{{GLUE}: A Multi-Task Benchmark and Analysis Platform for Natural Language Understanding}},
author={Alex Wang and Amanpreet Singh and Julian Michael and Felix Hill and Omer Levy and Samuel R. Bowman},
booktitle={International Conference on Learning Representations},
year={2019},
url={https://openreview.net/forum?id=rJ4km2R5t7},
}

@inproceedings{Zhang2015CharacterlevelCN,
 author = {Zhang, Xiang and Zhao, Junbo and LeCun, Yann},
 booktitle = {Advances in Neural Information Processing Systems},
 pages = {},
 title = {{Character-level Convolutional Networks for Text Classification}},
 volume = {28},
 year = {2015}
}

@inproceedings{nye2022show,
title={{Show Your Work: Scratchpads for Intermediate Computation with Language Models}},
author={Maxwell Nye and Anders Johan Andreassen and Guy Gur-Ari and Henryk Michalewski and Jacob Austin and David Bieber and David Dohan and Aitor Lewkowycz and Maarten Bosma and David Luan and Charles Sutton and Augustus Odena},
booktitle={Deep Learning for Code Workshop},
year={2022},
url={https://openreview.net/forum?id=HBlx2idbkbq}
}

@inproceedings{
yang2024parallelizing,
title={{Parallelizing Linear Transformers with the Delta Rule over Sequence Length}},
author={Songlin Yang and Bailin Wang and Yu Zhang and Yikang Shen and Yoon Kim},
booktitle={The Thirty-eighth Annual Conference on Neural Information Processing Systems},
year={2024},
url={https://openreview.net/forum?id=y8Rm4VNRPH}
}

@article{guo2025log,
  title={{Log-Linear Attention}},
  author={Guo, Han and Yang, Songlin and Goel, Tarushii and Xing, Eric P and Dao, Tri and Kim, Yoon},
  journal={arXiv preprint arXiv:2506.04761},
  year={2025}
}

@article{BERT-base-uncased,
  author    = {Jacob Devlin and
               Ming{-}Wei Chang and
               Kenton Lee and
               Kristina Toutanova},
  title     = {{{BERT:} Pre-training of Deep Bidirectional Transformers for Language
               Understanding}},
  year      = {2018},
  url       = {http://arxiv.org/abs/1810.04805},
  archivePrefix = {arXiv},
  eprint    = {1810.04805},
}

@article{yau2025sequential,
  title={{Sequential-Parallel Duality in Prefix Scannable Models}},
  author={Yau, Morris and Gupta, Sharut and Engelmayer, Valerie and Irie, Kazuki and Jegelka, Stefanie and Andreas, Jacob},
  journal={arXiv preprint arXiv:2506.10918},
  year={2025}
}

@inproceedings{tomita1982dynamic,
  author = {Tomita, Masaru},
  booktitle = {{P}roceedings of the Fourth Annual Conference of the Cognitive Science Society},
  title = {{Dynamic Construction of Finite Automata From Examples Using Hill-Climbing}},
  year = {1982}
}

@InProceedings{fdezdelpozoromero2023gradient,
author="Fdez. del Pozo Romero, Juan
and Lago-Fern{\'a}ndez, Luis F.",
title={{Gradient-Based Learning of Finite Automata}},
booktitle="Artificial Neural Networks and Machine Learning -- ICANN 2023",
year="2023",
pages="294--305",
isbn="978-3-031-44198-1"
}
